\pdfoutput=1

\documentclass[11pt]{article}

\usepackage{acl}

\usepackage{times}
\usepackage{latexsym}

\usepackage[T1]{fontenc}
\usepackage[utf8]{inputenc}

\usepackage{microtype}

\usepackage{graphicx}
\usepackage{caption}
\usepackage{subcaption}
\usepackage{booktabs}
\usepackage{colortbl}%
\usepackage{multirow}
\usepackage{bm}
\usepackage[export]{adjustbox} %
\usepackage{amsmath}
\usepackage[normalem]{ulem}
\usepackage{array}

\usepackage{cleveref}
\crefname{section}{\S}{\S\S}
\Crefname{section}{\S}{\S\S}
\crefformat{section}{\S#2#1#3}
\crefname{figure}{Figure}{Figures}
\crefname{alg}{Alg.}{Algs.}
\crefname{thm}{Theorem}{Theorems}
\crefname{line}{line}{lines}
\crefname{appendix}{Appendix}{}
\crefname{equation}{Eq.}{Eqs.}
\crefname{defin}{Def.}{Defs.}
\crefname{tab}{Table}{Tables}
\crefname{prop}{Proposition}{Propositions}
\crefname{cor}{Corollary}{}
\crefname{observation}{Observation}{}
\crefname{assumption}{Assumption}{}
\crefname{hypothesis}{Hyp.}{Hypotheses}

\usepackage{todonotes}
\makeatletter
\newcommand*\iftodonotes{\if@todonotes@disabled\expandafter\@secondoftwo\else\expandafter\@firstoftwo\fi}  %
\makeatother

\definecolor{tableau-blue}{RGB}{31, 119, 180} 
\definecolor{tableau-orange}{RGB}{255, 127, 14} 
\definecolor{tableau-green}{RGB}{44, 160, 44} 

\title{Few-shot Fine-tuning vs. In-context Learning:\\ A Fair Comparison and Evaluation}

\author{Marius Mosbach\textsuperscript{1} \,
 Tiago Pimentel\textsuperscript{2} \,
 Shauli Ravfogel\textsuperscript{3} \, Dietrich Klakow\textsuperscript{1} \, Yanai Elazar\textsuperscript{4,5} \ \\
\textsuperscript{1}Saarland University, Saarland Informatics Campus, 
\textsuperscript{2}University of Cambridge, \\
\textsuperscript{3}Bar-Ilan University,
\textsuperscript{4}Allen Institute for Artificial Intelligence, 
\textsuperscript{5}University of Washington\\
{\tt  mmosbach@lsv.uni-saarland.de}
  }

\begin{document}
\maketitle

\begin{abstract}

Few-shot fine-tuning and in-context learning are two alternative strategies for task adaptation of pre-trained language models. 
Recently, in-context learning has gained popularity over fine-tuning due to its simplicity and improved out-of-domain generalization, and because extensive evidence shows that fine-tuned models pick up on spurious correlations.
Unfortunately, previous comparisons of the two approaches were done using models of different sizes. This raises the question of whether the observed weaker out-of-domain generalization of fine-tuned models is an inherent property of fine-tuning or a limitation of the experimental setup.
In this paper, we compare the generalization of few-shot fine-tuning and in-context learning to challenge datasets, while controlling for the models used, the number of examples, and the number of parameters, ranging from 125M to 30B.
Our results show that fine-tuned language models \emph{can} in fact generalize well out-of-domain.
We find that both approaches generalize similarly; they exhibit large variation and depend on properties such as model size and the number of examples, highlighting that robust task adaptation remains a challenge.
\footnote{Code available at: \href{https://github.com/uds-lsv/llmft}{https://github.com/uds-lsv/llmft}.\looseness=-1
}

\end{abstract}

\section{Introduction}
\label{sec:introduction}
\begin{figure}[t] 
    \centering
\resizebox{0.99\columnwidth}{!}{%
  \begin{tabular}{cc}
    \small{ICL -- 16 samples} & \small{FT -- 16 samples} \\
      \includegraphics[valign=m,width=1.5in]{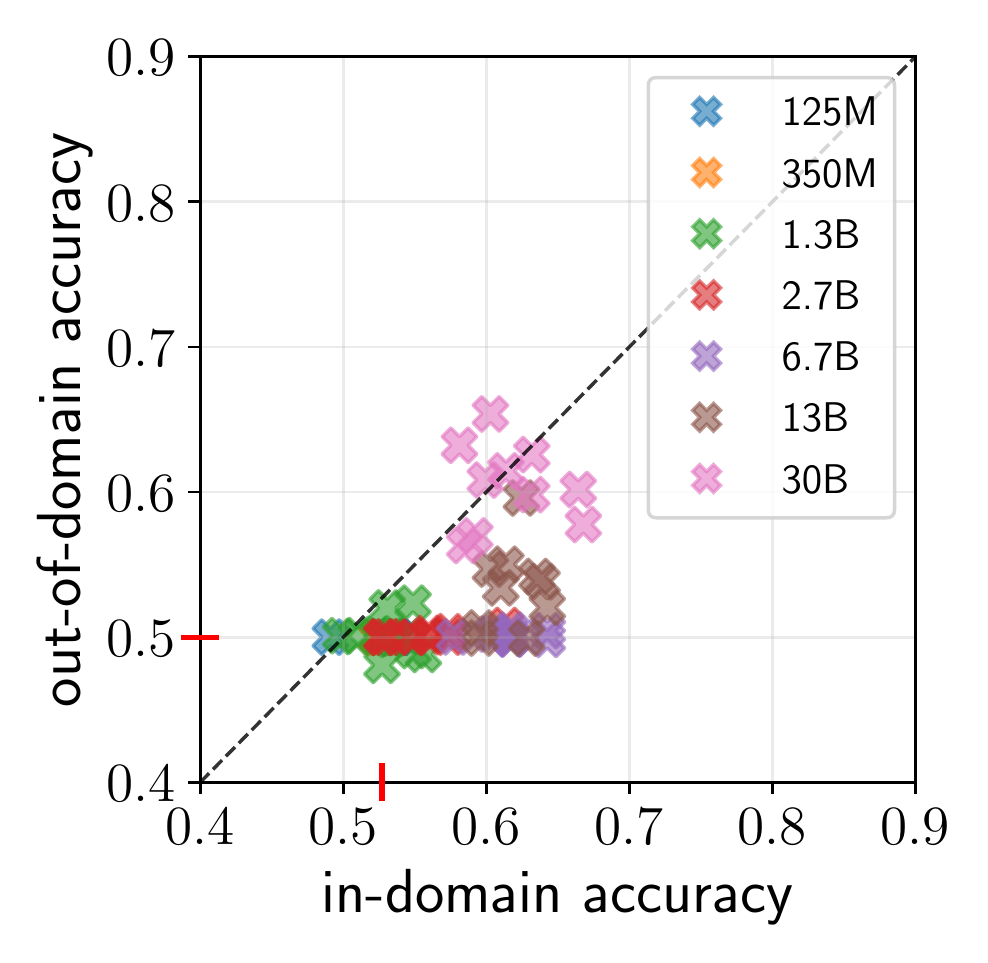} &
      \includegraphics[valign=m,width=1.5in]{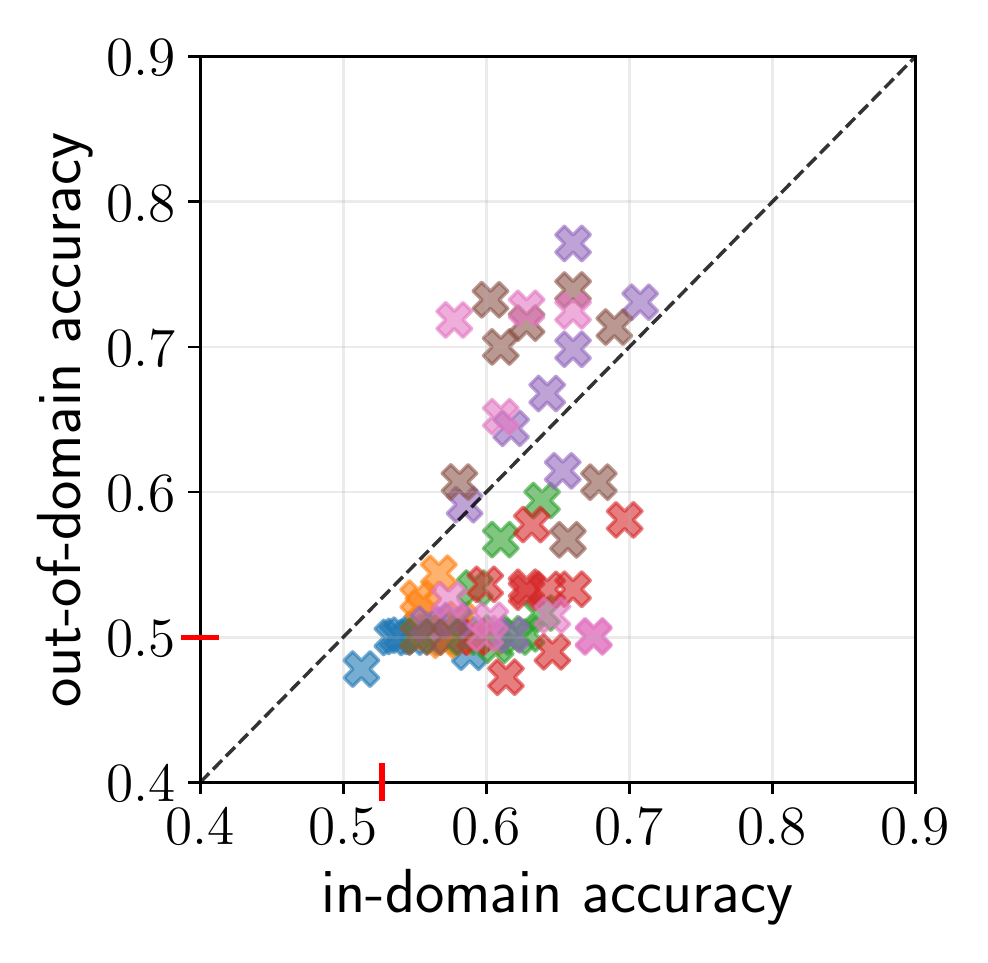} \\
  \end{tabular}
  }%
  \caption{In-domain (RTE) and out-of-domain performance (HANS) for in-context learning (ICL) and fine-tuning (FT) with OPT models of various sizes. We fine-tune models using pattern-based fine-tuning. We report results using 10 different data seeds. When using 16 samples, ICL's performance with a 30B model is comparable to that of FT with smaller models (6.7B) and for most model sizes, FT outperforms ICL (see \Cref{tab:appendix-statistical-tests-in-domain-ood-rte} for significance tests).
    \textcolor{red}{$\boldsymbol{-}$} in the x- and y-axes indicates majority class accuracy. 
    } %
    \vspace{-7pt}
    \label{fig:teaser}
 \end{figure}

Adapting a pre-trained language model to a target task is of high practical importance to the natural language processing (NLP) community \citep[as seen in][\emph{inter alia}]{peters-etal-2018-deep,howard-ruder-2018-universal, devlin-etal-2019-bert, brown-etal-2020-language}.
Among the commonly used \textit{task adaptation} strategies, two stand out: \textit{fine-tuning} (FT) and \textit{in-context learning} (ICL).\footnote{We describe both these strategies in further detail in Section \cref{sec:background}. In short, fine-tuning a model involves a supervised learning setup on a target dataset; while ICL involves prompting a model with a series of input--label pairs, without updating the model's parameters.}\looseness-1

Both approaches come with pros and cons: 
ICL reuses a single pre-trained model for various downstream tasks, allows specifying the desired behavior via natural language, and has recently shown impressive results on challenging reasoning tasks \citep{brown-etal-2020-language,wei2022chain,press2022measuring}. 
However, the model's context size limits the number of demonstrations that can be used. For instance, using 32 randomly selected examples from the RTE dataset \citep{dagan-etal-2006-rte} already exceeds the context size of OPT models \citep{zhang2022opt}.\footnote{While GPT-3 and OPT both have a context size of 2048 tokens, more recent models such as GPT-4 \citep{openai2023gpt4} -- which has been developed concurrently to this work -- support larger contexts of up to 8192 tokens.}
In addition, ICL is highly sensitive to the format and order of its inputs \citep{lu-etal-2022-fantastically, min-etal-2022-rethinking}.
FT, on the other hand, typically results in a single specialized model per task,\footnote{Parameter-efficient FT methods (e.g. \citet{ben-zaken-etal-2022-bitfit, hu2022lora}) address this issue and allow to re-use most of the pre-trained weights across tasks.} and can be applied to training sets of arbitrary size. 
However, such models are sensitive to initialization \cite{dodge-etal-2020-fine} and can suffer from instability during training \citep{mosbach-etal-2021-on}.

For text classification tasks, where both strategies often lead to similar performance on in-domain data (when using the same amount of data), recent works have argued that ICL leads to better out-of-domain (OOD) generalization \citep{si2023prompting, awadalla-etal-2022-exploring}. 
However, these comparisons of generalization abilities were not conducted under equal conditions.
Most studies compare the ICL abilities of large models \citep[e.g. GPT-3, 175B;][]{brown-etal-2020-language} to the FT abilities of much smaller models  \citep[e.g. RoBERTa-large, 350M; ][]{liu-etal-2019-roberta}.
These comparisons raise the question of whether FT indeed leads to weaker OOD generalization than ICL, or whether this is just a byproduct of the experimental setup.
In \cref{fig:teaser}, we show this is indeed the case: when given only 16 examples, fine-tuning a 6.7B parameters model already achieves similar results to ICL with a 30B model, and FT performance keeps improving with larger models.\footnote{\Cref{tab:appendix-statistical-tests-in-domain-ood-rte} presents significance tests for these results.} Moreover, we show in Section \ref{sec:closer-look-ft} that fine-tuning performance improves even further when training on more data.

In this paper, we compare ICL and FT on an \textbf{equal footing} (\cref{sec:approach}). We compare both strategies using the same model \citep[OPT;][]{zhang2022opt}, the same number of parameters (from 125M to 30B), and the same number of examples.
Our results and analyses (\cref{sec:results}) show that both approaches often achieve comparable results. Both methods are unstable and can perform badly on in-domain and OOD data due to training instability, or prompt choice. We also find that both approaches improve as we increase model size, and that, for the models and datasets we consider, FT often generalizes even better than ICL. 
Notably, this is in contrast to prior work (\cref{sec:related-work}), highlighting the need for fair comparisons of task adaptation strategies.
Based on our findings, we discuss the strengths and limitations of FT and ICL (\cref{sec:ft-icl-comparison}), which can inform when to use and how to get the most out of each method.\looseness=-1

\section{Background} 
\label{sec:background}
\subsection{Fine-tuning}

\textit{Pattern-based fine-tuning} (PBFT) is a recently proposed FT approach that uses the pre-trained language modeling head\footnote{In the case of encoder-only masked language models, such as BERT, this is usually an MLP layer. In the case of decoder-only models, such as OPT, this is a linear projection.} 
instead of a randomly initialized classifier (as used in standard fine-tuning; \citealt{howard-ruder-2018-universal, devlin-etal-2019-bert}), to obtain predictions \citep[\emph{inter alia}]{schick-schutze-2021-exploiting, gao-etal-2021-making}.
Compared to vanilla FT, we have to specify an \textit{input pattern} (to cast the task as a language modeling problem) and define a \textit{verbalizer} \citep[which maps tokens in the pre-trained model's vocabulary to labels;][]{schick-etal-2020-automatically}.
For example, a NLI pattern might look as follows: \texttt{\{premise\} Question: \{hypothesis\} Yes or No?}, and the verbalizer will use \texttt{Yes} and \texttt{No} as tokens.
Given these inputs and targets, model parameters are fine-tuned as usual.
This method has been shown to be efficient for few-shot learning despite having no advantage over vanilla FT when the number of examples is large \cite{tam-etal-2021-improving,logan-iv-etal-2022-cutting}.

\subsection{In-context learning}

\textit{In-context learning} (ICL) is a task adaptation strategy that does not update the weights of the pre-trained model \cite{brown-etal-2020-language}; instead, ICL adapts a model to a task by conditioning it on a sequence of \textit{demonstrations}.
A demonstration typically refers to an input $x$ accompanied by its ground-truth label $y$, both of which have been converted to a specific format using a \textit{pattern} and a \textit{verbalizer} (similar to PBFT). ICL thus feeds the model a sequence of such demonstrations, followed by the test input (modified by applying the pattern transformation). The language model is then expected to predict the label of this final data point.\footnote{The evaluation only considers the probabilities assigned to the verbalizer tokens, ignoring any probability mass assigned to other tokens. See \cref{sec:in-context-setup} for details.}
Recent work has argued that ICL leads to better out-of-domain performance, when compared to FT \citep{si2023prompting, awadalla-etal-2022-exploring}. 
We show that this often does not hold.

\vspace{-3pt}
\section{A fair comparison of FT and ICL} 
\label{sec:approach}

We perform a fair comparison of task adaptation via FT and ICL, focusing on in-domain and OOD generalization. We compare them in the few-shot setting using the same models. In the following paragraphs, we provide details about our setup.

\paragraph{In-domain generalization}

We measure in-domain generalization by measuring accuracy on the validation set of each dataset. This is a common practice in analysis works, and used in previous work \cite{utama-etal-2021-avoiding, bandel-etal-2022-lexical}. 

\paragraph{Out-of-domain generalization}

We consider OOD generalization under \textit{covariate shift} \citep{hupkes-etal-2022-state-of}. Specifically, we focus on generalization to \emph{challenge datasets}, designed to test whether models adopt a particular heuristic, or make predictions based on spurious correlations during inference \citep{mccoy-etal-2019-right,elazar-etal-2021-back}.

\paragraph{Models} 

We run all our experiments using 7 different OPT models \citep{zhang2022opt} ranging from 125 million to 30 billion parameters, all of which have been trained on the same data. This allows us to study the effect of model size on performance without the confound of using different training data.\footnote{OPT 30B is the largest model we were able to fit given our resources.}\looseness-1
\paragraph{Tasks and datasets}

We focus on two classification tasks in English: natural language inference (NLI) and paraphrase identification. For NLI, we use MNLI \citep{williams-etal-2018-broad} and RTE \citep{dagan-etal-2006-rte} as in-domain datasets, and evaluate OOD generalization on the lexical overlap subset of HANS \citep{mccoy-etal-2019-right}.\footnote{Due to similar trends on different HANS subsets in preliminary experiments, we focus on the lexical overlap subset.\looseness=-1} We binarize MNLI by removing the neutral examples\footnote{We compare this to merging the neutral and contradiction classes in \cref{appendix:additional-results-ft}, and obtain very similar results.} which allows us to better compare MNLI with RTE (which only has two labels).
For paraphrase identification, we train on QQP \citep{qqp} and evaluate OOD generalization on PAWS-QQP \citep{zhang-etal-2019-paws}. Given the large size of the QQP validation set (more than 300k examples), we randomly select 1000 validation examples.

\begin{figure*}[ht] 
    \centering
  \begin{tabular}{lccc}
        & \hspace{4mm} \small{MNLI} & \hspace{4mm} \small{RTE} & \hspace{4mm} \small{QQP} \\
    
      \small{ICL} & 
      \includegraphics[valign=m,width=1.5in]{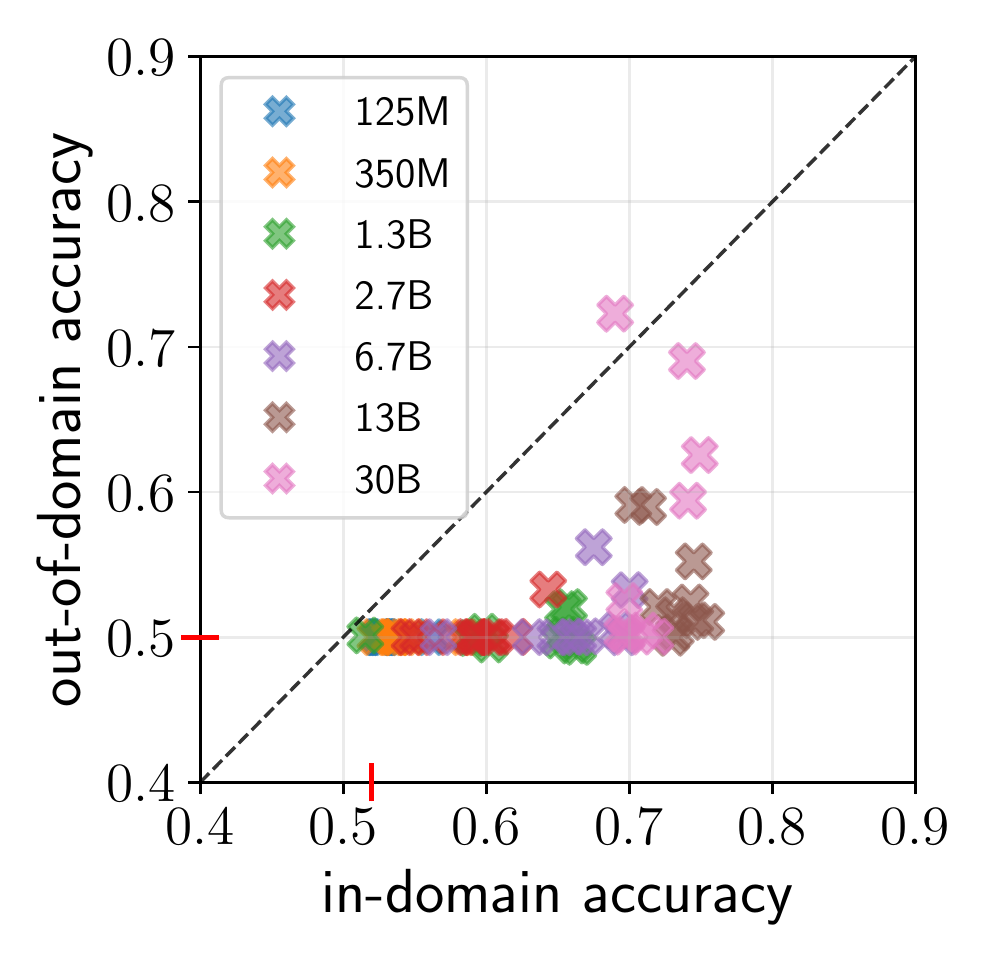} &
      \includegraphics[valign=m,width=1.5in]{figures/in-context/in-context_all-models_16-shots_rte_gpt-3} &
      \includegraphics[valign=m,width=1.5in]{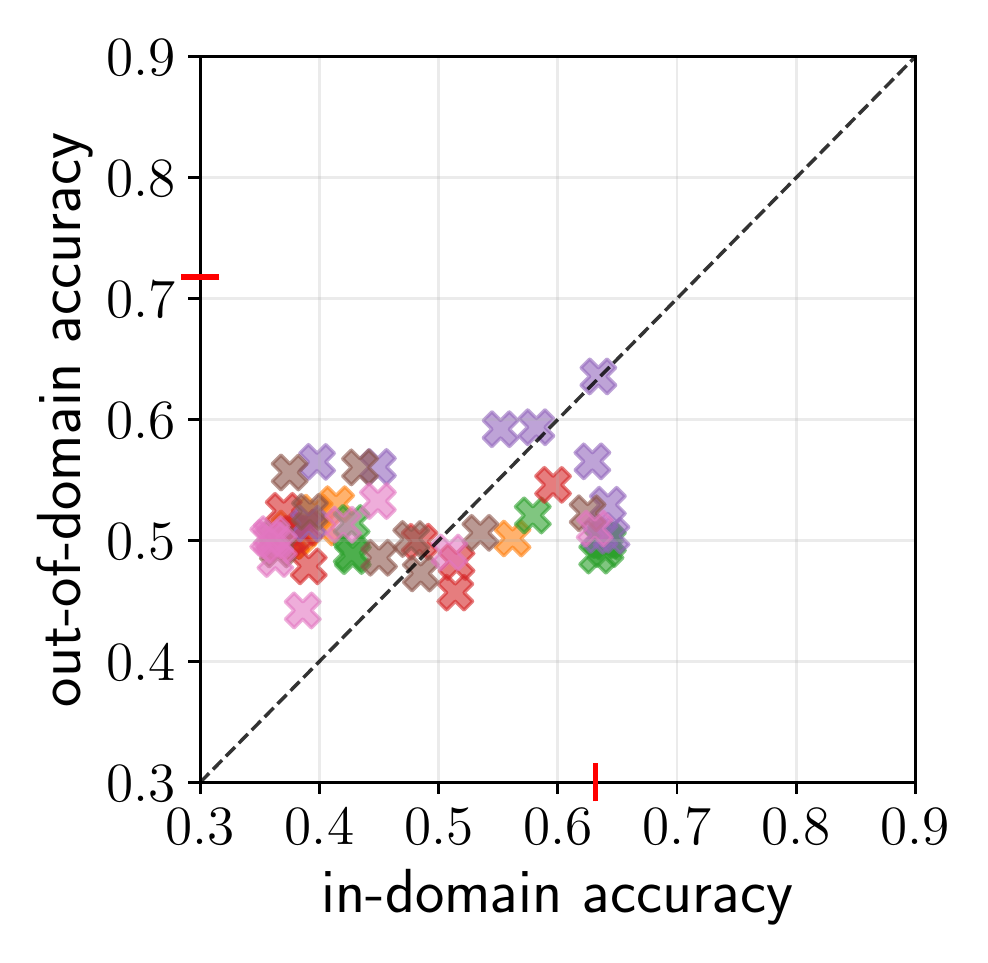} \\
      
      \small{PBFT} & 
      \includegraphics[valign=m,width=1.5in]{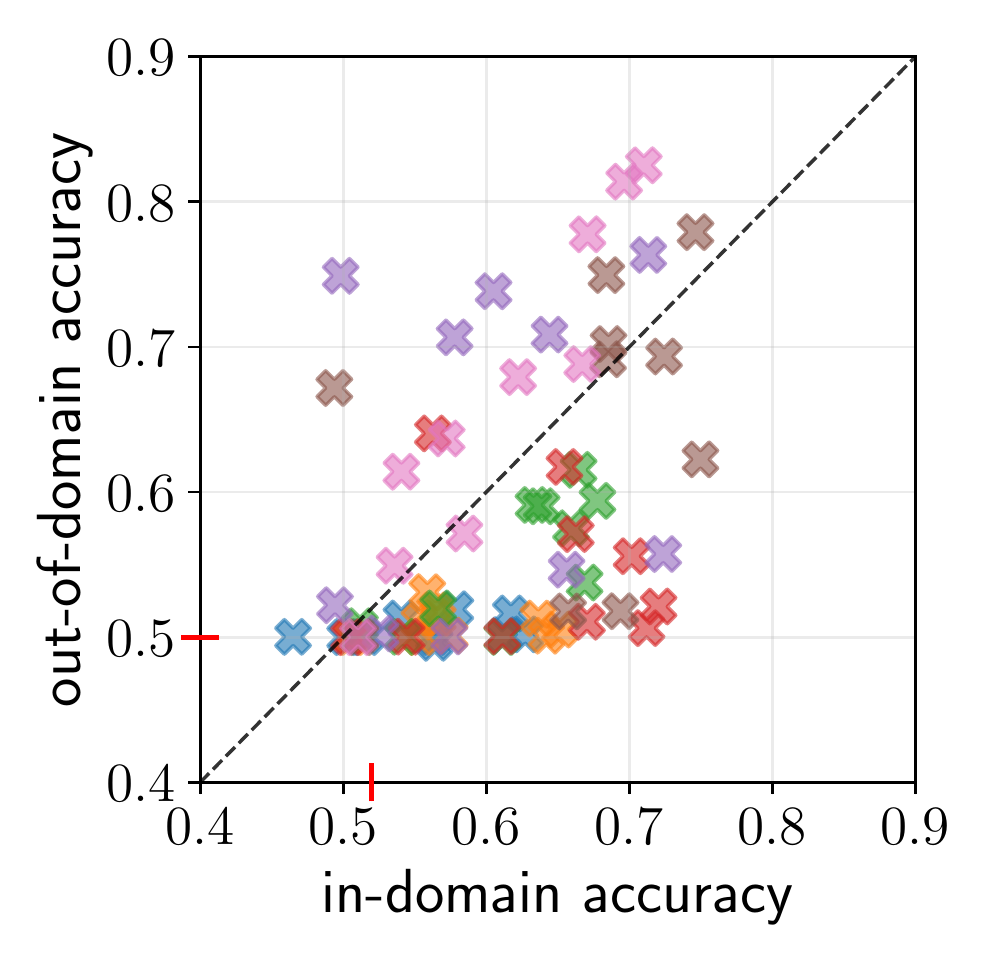} &
      \includegraphics[valign=m,width=1.5in]{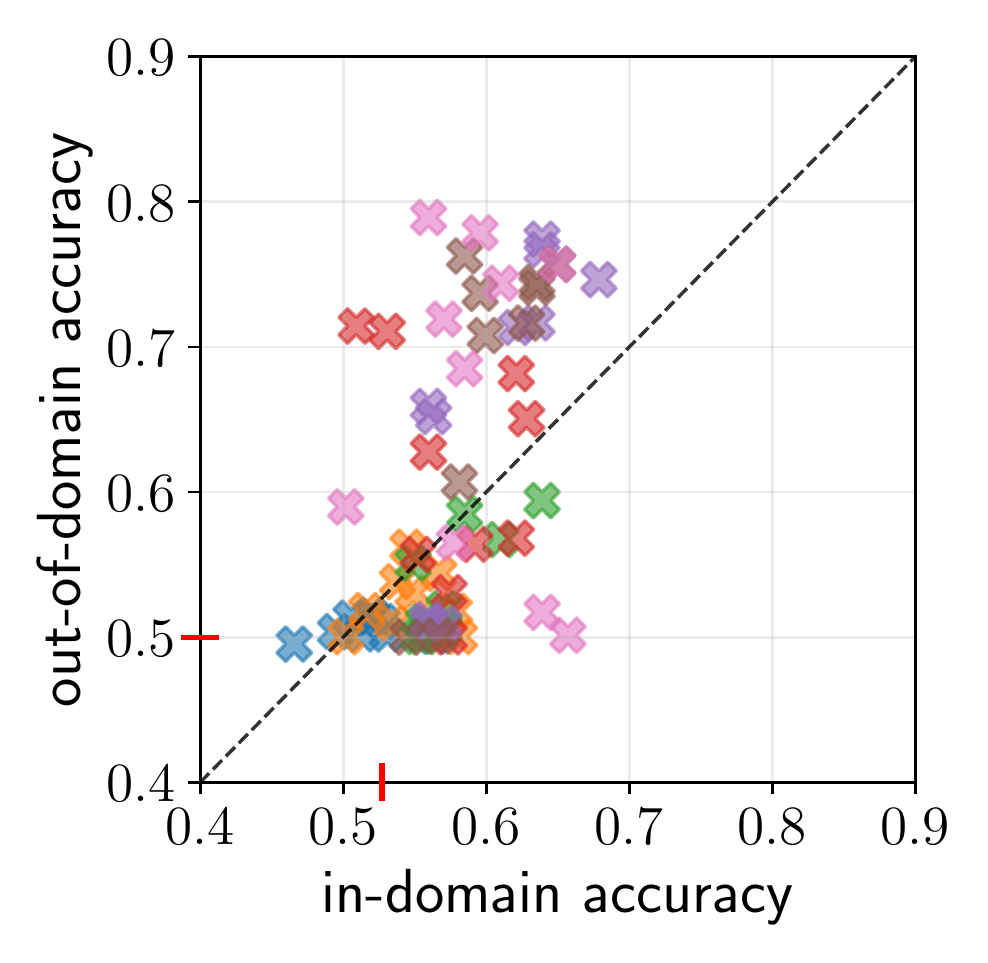} &
      \includegraphics[valign=m,width=1.5in]{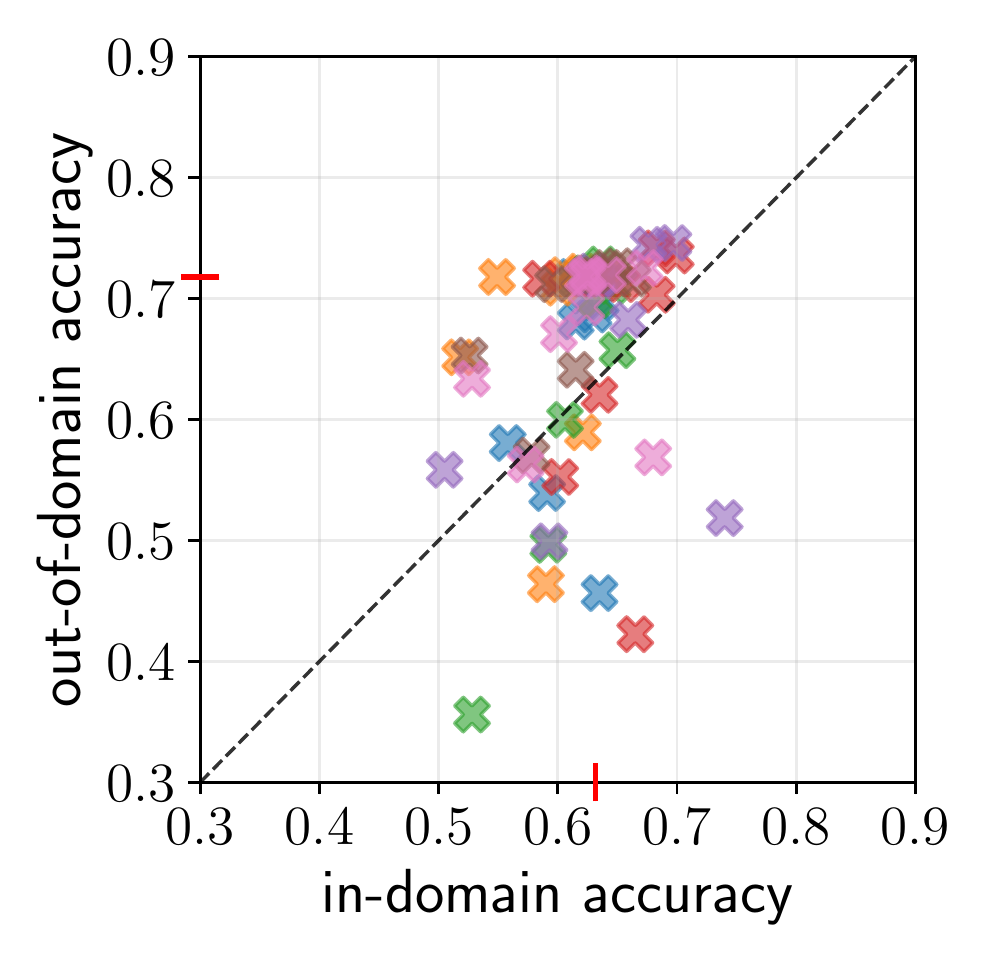} \\
     
  \end{tabular}
  \caption{ICL and FT results for OPT models of various sizes. 
    For each approach, we use 16 examples and perform model selection according to OOD performance. We plot 10 runs per model size which differ only in the data seed.
    \textcolor{red}{$\boldsymbol{-}$} in the x- and y-axis indicates majority class accuracy.
    }
    \label{fig:results-summary}
\end{figure*}

\begin{table*}
    \begin{subtable}{.49\textwidth}\centering
        {
            \resizebox{1\textwidth}{!}{%
            \begin{tabular}{llccccccc}
        \toprule
        & 
        & \multicolumn{7}{c}{\textbf{FT}} \\ \cmidrule{3-9} 
        & & \textbf{125M} & \textbf{350M} & \textbf{1.3B} & \textbf{2.7B} & \textbf{6.7B} & \textbf{13B} & \textbf{30B} 
        \\ \midrule
        \multirow{7}{*}{\rotatebox[origin=c]{90}{\textbf{ICL}}} & \textbf{125M}  & $-0.00$ & \cellcolor{blue!35}$\phantom{-}0.01$ & \cellcolor{blue!35}$\phantom{-}0.02$ & \cellcolor{blue!35}$\phantom{-}0.03$ & \cellcolor{blue!35}$\phantom{-}0.12$ & \cellcolor{blue!35}$\phantom{-}0.14$ & \cellcolor{blue!35}$\phantom{-}0.09$  \\
         & \textbf{350M}  & $-0.00$ & \cellcolor{blue!35}$\phantom{-}0.01$ & \cellcolor{blue!35}$\phantom{-}0.02$ & \cellcolor{blue!35}$\phantom{-}0.03$ & \cellcolor{blue!35}$\phantom{-}0.12$ & \cellcolor{blue!35}$\phantom{-}0.14$ & \cellcolor{blue!35}$\phantom{-}0.09$ \\
         & \textbf{1.3B} & $-0.00$ & \cellcolor{blue!35}$\phantom{-}0.01$ & \cellcolor{blue!35}$\phantom{-}0.02$ & \cellcolor{blue!35}$\phantom{-}0.03$ & \cellcolor{blue!35}$\phantom{-}0.12$ & \cellcolor{blue!35}$\phantom{-}0.14$ & \cellcolor{blue!35}$\phantom{-}0.09$  \\
          & \textbf{2.7B} & $-0.00$ & \cellcolor{blue!35}$\phantom{-}0.01$ & \cellcolor{blue!35}$\phantom{-}0.02$ & \cellcolor{blue!35}$\phantom{-}0.03$ & \cellcolor{blue!35}$\phantom{-}0.12$ & \cellcolor{blue!35}$\phantom{-}0.14$ & \cellcolor{blue!35}$\phantom{-}0.09$  \\
         & \textbf{6.7B} & $-0.00$ & \cellcolor{blue!35}$\phantom{-}0.01$ & \cellcolor{blue!35}$\phantom{-}0.02$ & \cellcolor{blue!35}$\phantom{-}0.03$ & \cellcolor{blue!35}$\phantom{-}0.12$ & \cellcolor{blue!35}$\phantom{-}0.14$ & \cellcolor{blue!35}$\phantom{-}0.09$  \\
         & \textbf{13B} & \cellcolor{red!15}$-0.04$ & \cellcolor{red!15}$-0.02$ & $-0.01$ & $-0.00$ & \cellcolor{blue!35}$\phantom{-}0.09$ & \cellcolor{blue!35}$\phantom{-}0.11$ & $\phantom{-}0.05$  \\
         & \textbf{30B} & \cellcolor{red!15}$-0.11$ & \cellcolor{red!15}$-0.09$ & \cellcolor{red!15}$-0.08$ & \cellcolor{red!15}$-0.08$ & $\phantom{-}0.02$ & $\phantom{-}0.03$ & $-0.02$  \\
         
        \bottomrule
    \end{tabular}%
    }
        }
        \caption{RTE}\label{tab:appendix-statistical-tests-in-domain-ood-rte}
    \end{subtable}%
    ~~ 
    \begin{subtable}{.49\textwidth}\centering
        {
            \resizebox{1\textwidth}{!}{%
            \begin{tabular}{llccccccc}
        \toprule
        & 
        & \multicolumn{7}{c}{\textbf{FT}} \\ \cmidrule{3-9} 
        & & \textbf{125M} & \textbf{350M} & \textbf{1.3B} & \textbf{2.7B} & \textbf{6.7B} & \textbf{13B} & \textbf{30B} 
        \\ \midrule
        \multirow{7}{*}{\rotatebox[origin=c]{90}{\textbf{ICL}}} & \textbf{125M}  & $-0.00$ & $\phantom{-}0.00$ & \cellcolor{blue!35}$\phantom{-}0.02$ & $\phantom{-}0.01$ & \cellcolor{blue!35}$\phantom{-}0.10$ & \cellcolor{blue!35}$\phantom{-}0.11$ & \cellcolor{blue!35}$\phantom{-}0.07$  \\
         & \textbf{350M}  & $-0.00$ & $\phantom{-}0.00$ & \cellcolor{blue!35}$\phantom{-}0.02$ & $\phantom{-}0.01$ & \cellcolor{blue!35}$\phantom{-}0.10$ & \cellcolor{blue!35}$\phantom{-}0.11$ & \cellcolor{blue!35}$\phantom{-}0.07$ \\
         & \textbf{1.3B} & $-0.01$ & $-0.00$ & \cellcolor{blue!35}$\phantom{-}0.01$ & $\phantom{-}0.01$ & \cellcolor{blue!35}$\phantom{-}0.10$ & \cellcolor{blue!35}$\phantom{-}0.11$ & \cellcolor{blue!35}$\phantom{-}0.07$  \\
          & \textbf{2.7B} & $-0.01$ & $-0.00$ & $\phantom{-}0.01$ & $\phantom{-}0.01$ & \cellcolor{blue!35}$\phantom{-}0.09$ & \cellcolor{blue!35}$\phantom{-}0.10$ & \cellcolor{blue!35}$\phantom{-}0.07$  \\
         & \textbf{6.7B} & \cellcolor{red!15}$-0.01$ & $-0.01$ & $\phantom{-}0.01$ & $\phantom{-}0.00$ & \cellcolor{blue!35}$\phantom{-}0.09$ & \cellcolor{blue!35}$\phantom{-}0.10$ & \cellcolor{blue!35}$\phantom{-}0.06$  \\
         & \textbf{13B} & \cellcolor{red!15}$-0.03$ & \cellcolor{red!15}$-0.03$ & $-0.02$ & $-0.02$ & \cellcolor{blue!35}$\phantom{-}0.07$ & \cellcolor{blue!35}$\phantom{-}0.08$ & $\phantom{-}0.04$  \\
         & \textbf{30B} & \cellcolor{red!15}$-0.07$ & \cellcolor{red!15}$-0.07$ & $-0.05$ & \cellcolor{red!15}$-0.06$ & $\phantom{-}0.03$ & $\phantom{-}0.04$ & $\phantom{-}0.00$  \\
         
        \bottomrule
    \end{tabular}%
    }
        }
        \caption{MNLI}\label{tab:appendix-statistical-tests-in-domain-ood-mnli}
    \end{subtable}%
    \caption{Difference between average \textbf{out-of-domain performance} of ICL and FT on RTE (a) and MNLI (b) across model sizes. We use 16 examples and 10 random seeds for both approaches. For ICL, we use the \texttt{gpt-3} pattern. For FT, we use pattern-based fine-tuning (PBFT) and select checkpoints according to in-domain performance. We perform a Welch's t-test and color cells according to whether: \textcolor{red!45}{ICL performs significantly better than FT}, \textcolor{blue!75}{FT performs significantly better than ICL}. For cells without color, there is no significant difference.\looseness-1
    } \label{tab:appendix-statistical-tests-in-domain-ood}
\end{table*}
\paragraph{Few-shot setup}

We follow the same procedure for both approaches.
We randomly sample $n \in \{2, 16, 32, 64, 128\}$ examples from the in-domain training set of a given dataset (unless stated otherwise).\footnote{We sample an equal number of examples per label.}
Due to the high sensitivity of both approaches to the used pattern, as well as to the ordering of the demonstrations in ICL \citep{webson-pavlick-2022-prompt,lu-etal-2022-fantastically}, we sample 10 different sets of examples for each $n$. We also experiment with 3 different patterns, resulting in 30 runs per $n$ and adaption method.\footnote{Except for QQP, where we experiment with only 2 patterns, as one of the patterns is not applicable.} \Cref{tab:appendix-in-context-learning-patterns} in \Cref{sec:appendix:icl-details} provides an overview of the patterns and verbalizers for each task.

\paragraph{FT setup} 
\label{sec:ft-setup}

We perform few-shot PBFT using a minimal pattern \citep{logan-iv-etal-2022-cutting}, which simply adds a question mark at the end of every example. 
For the NLI verbalizer, we use \texttt{Yes} and \texttt{No}, which we map to the task's labels \texttt{entailment} and \texttt{not-entailment} respectively. For QQP, we also use \texttt{Yes} and \texttt{No} and map them to \texttt{not-duplicate} and \texttt{duplicate}.\footnote{Preliminary experiments showed that \texttt{Yes} and \texttt{No} is a strong verbalizer for binary classification tasks. This is consistent with previous findings \citep{webson-pavlick-2022-prompt}.}
We follow the recommendations of \citet{mosbach-etal-2021-on} and fine-tune all models for 40 epochs using a learning rate of $10^{-5}$ which increases linearly (warmup) for the first $10\%$ of the training steps and is kept constant afterward. Details of all hyper-parameters are provided in \cref{sec:appendix-ft-details}. 
\paragraph{ICL setup}
\label{sec:in-context-setup}

Given OPT's fixed context size of 2048 tokens we are limited in the number of examples used for demonstration. Our main experiments focus on 16 demonstrations, but we also present additional experiments using 2 and 32 demonstrations in  \Cref{appendix:additional-results}.\footnote{With the exception of RTE, where 32 examples do not fit OPT's context size}. 
We consider a prediction to be correct if the probability assigned to the verbalizer token of the ground-truth label is larger than the probability of the other verbalizer token. We use the same verbalizer tokens as for fine-tuning.
\section{Results}
\label{sec:results}

We present the results for in-domain and OOD model performance in \cref{fig:results-summary}, comparing both ICL and FT. We perform task adaptation using 16 examples for both strategies. For ICL, we provide additional results that demonstrate the importance of choosing the right pattern and number of demonstrations in \cref{appendix:additional-results-in-context}. For FT, we provide more details, ablations and discussion of various choices later in this section.\looseness=-1

\paragraph{In-domain performance} 

For MNLI and RTE, both ICL and FT exhibit in-domain performance above the majority baseline for most model sizes. Focusing on ICL, MNLI and RTE in-domain performance improves as model size increases.
On MNLI the largest model (30B) obtains an average performance of $71.4\%$ and a maximum performance of $74.9\%$. On RTE, ICL with the same model achieves an average and maximum performance of $61.7\%$ and $66.8\%$ respectively. On QQP, the trend of improved performance with increasing model size is less clear and most models perform worse than the majority baseline. \Cref{tab:icl-previous-work} (in \Cref{sec:appendix-icl-previous}) compares our ICL results with previous work.\looseness=-1

For FT, we similarly observe that in-domain performance increases with model size. Moreover, across all datasets and model sizes, FT with just 16 examples leads to similar in-domain performance as ICL (see \Cref{tab:appendix-statistical-tests-in-domain-in-domain,tab:appendix-statistical-tests-ood-in-domain} in \Cref{appendix:significance-tests} for statistical tests comparing in-domain performance of FT and ICL on RTE and MNLI).  On QQP, we again observe no clear relationship between model size and performance. Only 10 out of  70 models perform better than the majority baseline.\looseness-1

\begin{figure*}[ht] 
    \centering
  \begin{tabular}{lccc}
        & \hspace{4mm} \small{Best in-domain} & \hspace{4mm} \small{Last checkpoint} & \hspace{4mm} \small{Best out-of-domain} \\
      
      \small{MNLI} & 
      \includegraphics[valign=m,width=1.5in]{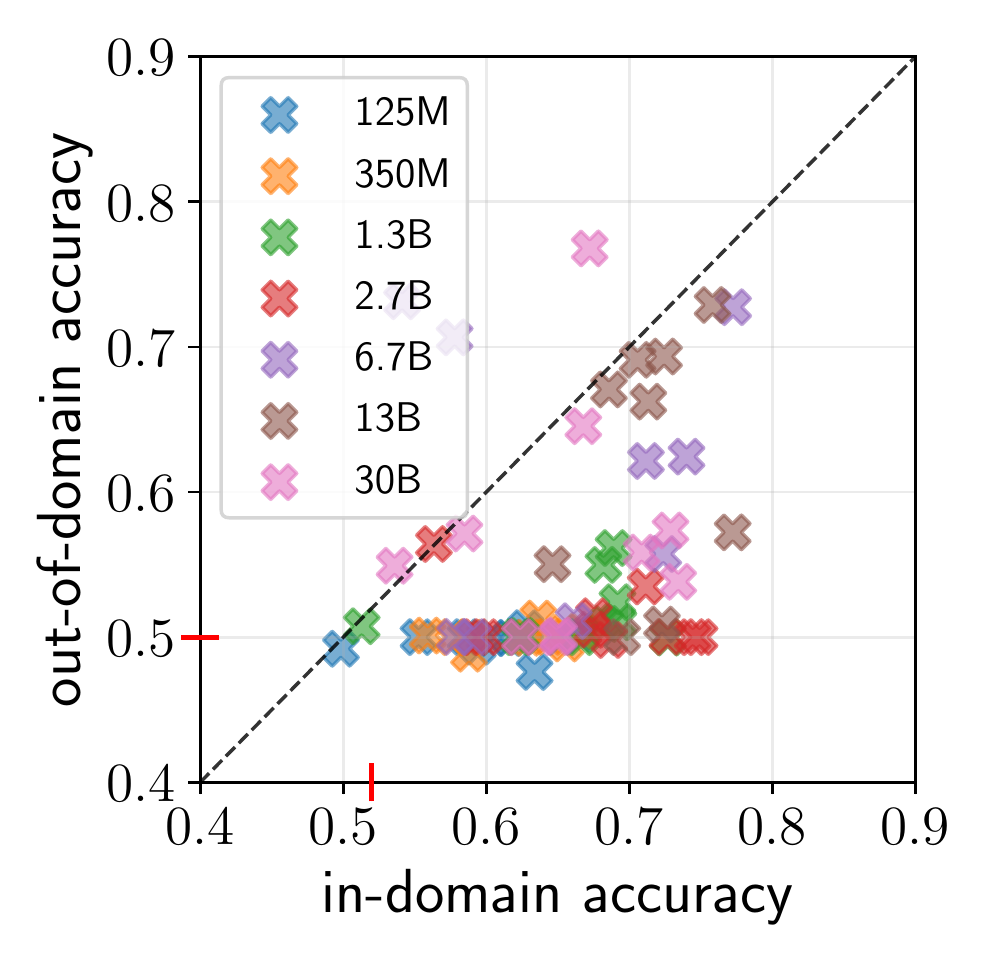} &
      \includegraphics[valign=m,width=1.5in]{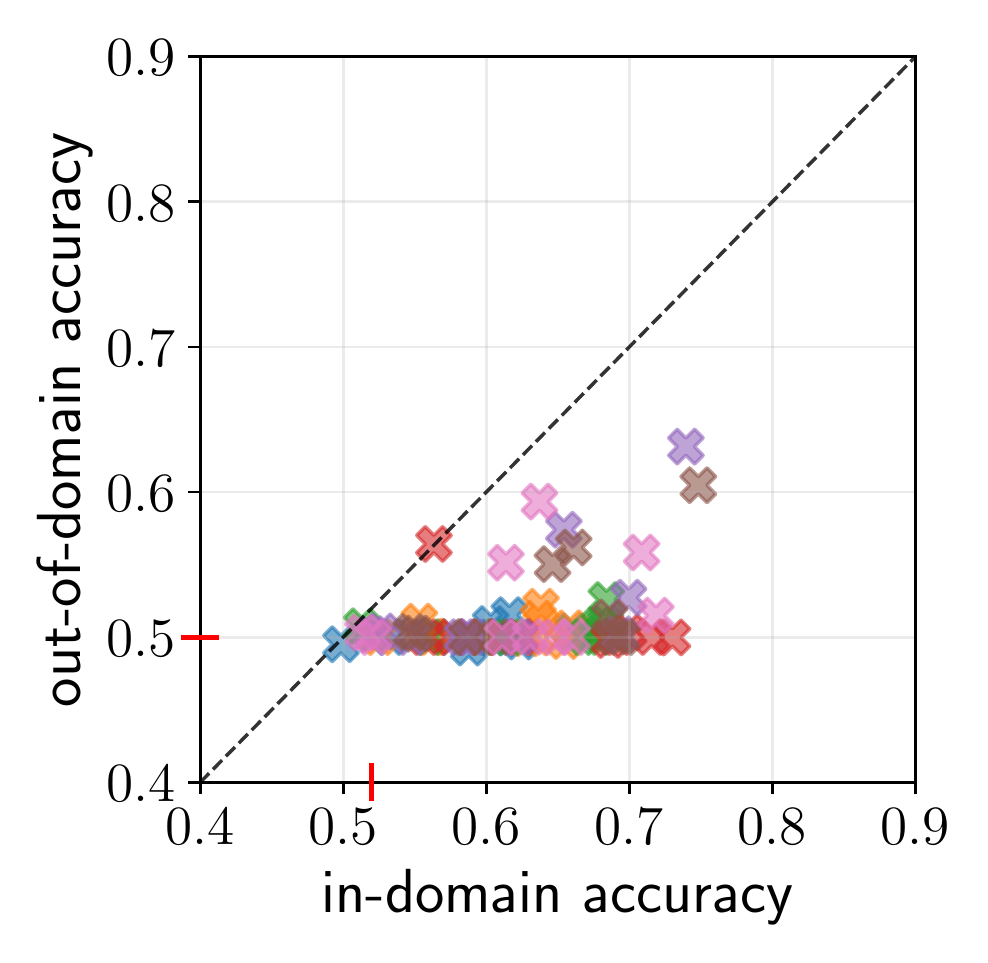} &
      \includegraphics[valign=m,width=1.5in]{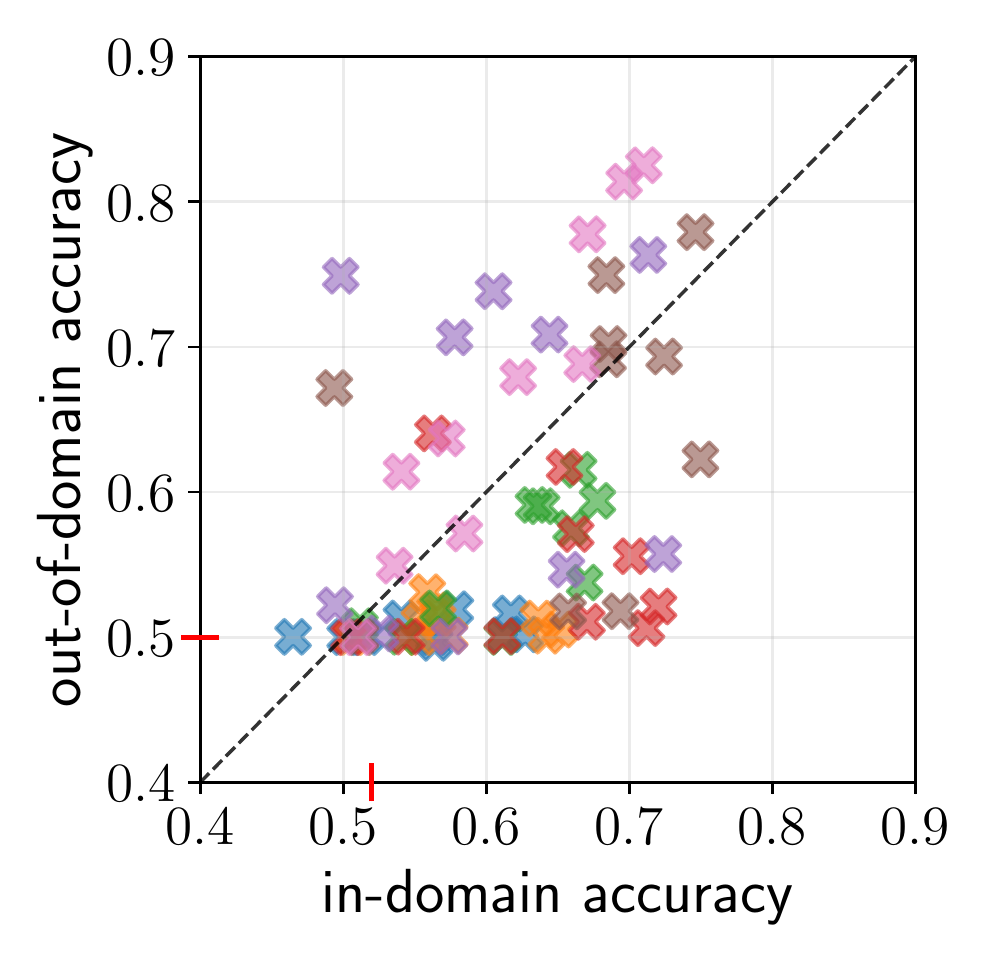} \\
      \small{RTE} & 
      \includegraphics[valign=m,width=1.5in]{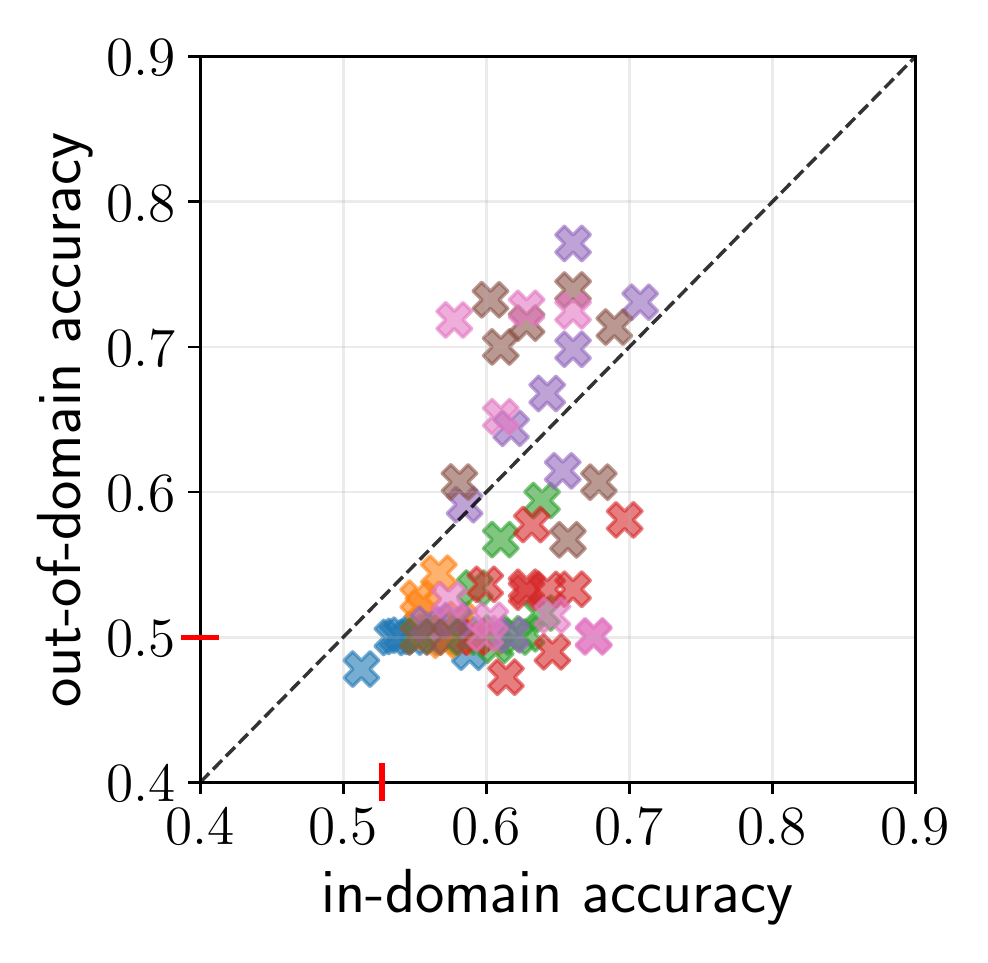} &
      \includegraphics[valign=m,width=1.5in]{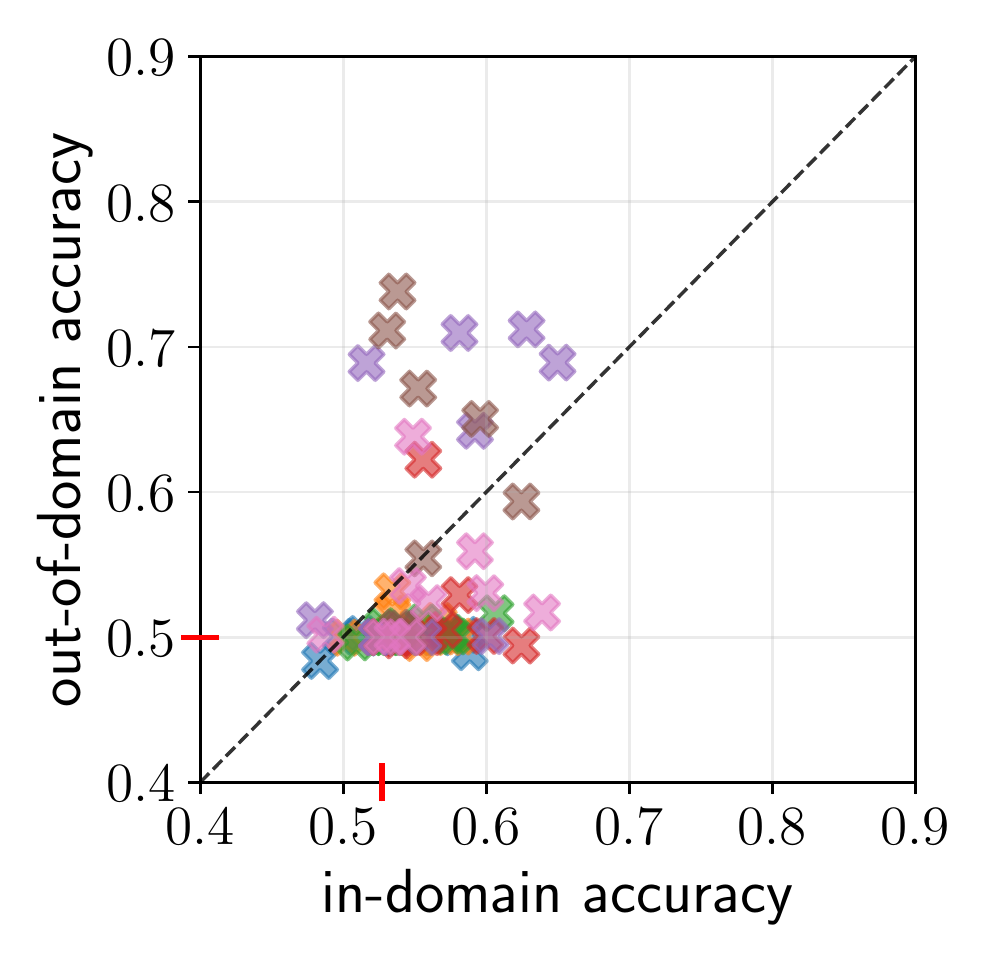} &
      \includegraphics[valign=m,width=1.5in]{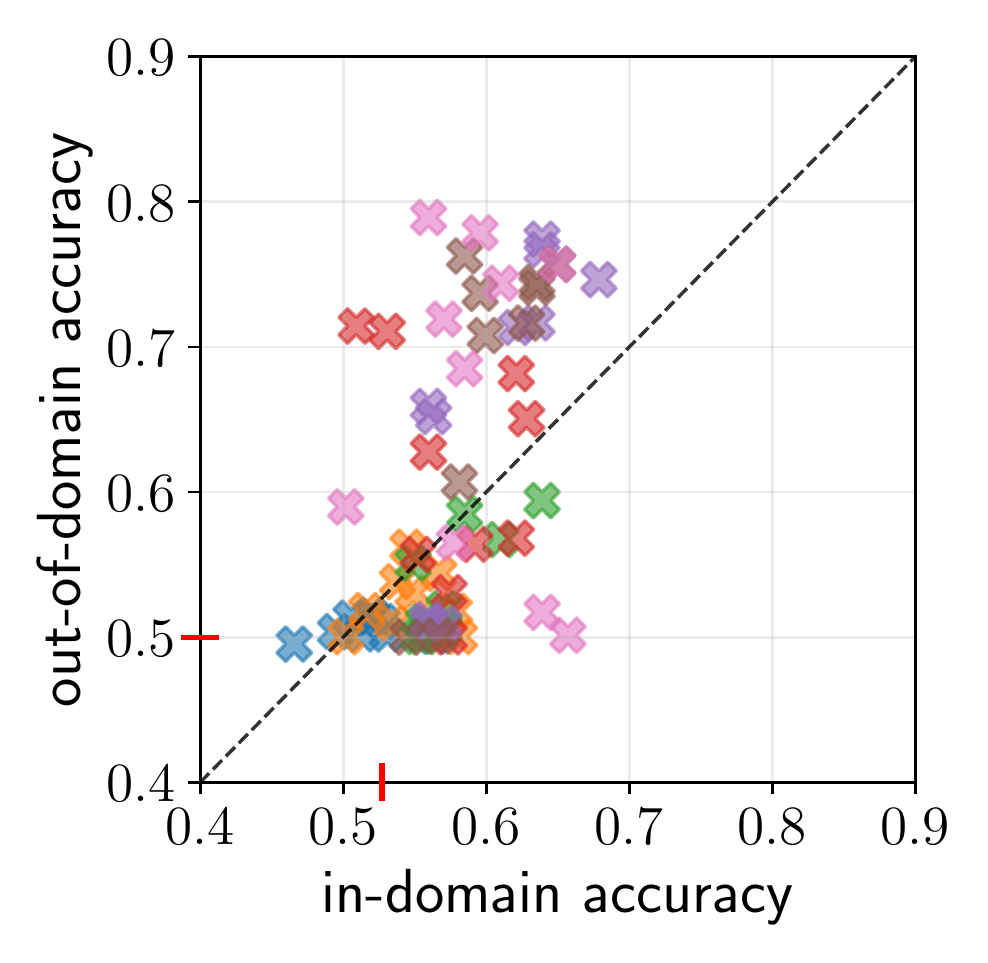} \\
  \end{tabular}
  \caption{Comparing model selection strategies in FT. 
    The first and second rows show results for MNLI and RTE respectively.
    We train on 16 examples and plot results for 10 runs for each model size.
    \textcolor{red}{$\boldsymbol{-}$} in the x- and y-axes indicates majority class accuracy.
    }
    \label{fig:ft-model-selection-mnli}
\end{figure*}
\paragraph{Out-of-domain performance}

Turning to OOD performance, we find that for MNLI and QQP most of the ICL models perform close to the majority baseline. On MNLI, only the largest model (30B) shows good OOD generalization for 4 out of 10 runs. On RTE, in-domain and OOD performance of the 30B model mostly overlap, which is consistent with the findings of \citet{si2023prompting}. In particular, when comparing the relationship between the in-domain and OOD performance of the 30B model to the smallest fine-tuned models (125M and 350M) one might conclude that ICL leads to better OOD performance; for FT on MNLI and RTE, indeed, the smallest models have poor OOD performance.

However, as model size increases, OOD performance increases as well, demonstrating that even in the challenging few-shot setting, fine-tuned models can generalize OOD. Focusing on the largest models (6.7B, 13B, and 30B) fine-tuned on MNLI, we find that for most runs, OOD performance is on par or even better than in-domain performance. On RTE, the trend is even stronger. Even with the 1.3B model, we observe good in-domain and OOD performance, and both improve as the models get larger. Notably, for many models, OOD performance is even better than in-domain performance.

In summary, \textbf{our comparison shows that fine-tuned language models can generalize OOD as well or even better than models adapted via ICL} (see statistical tests comparing them in \Cref{tab:appendix-statistical-tests-in-domain-ood}). This highlights the importance of comparing adaptation approaches using models of the same size.

\subsection{A closer look at FT generalization}
\label{sec:closer-look-ft}

Having established that few-shot FT can also lead to strong in-domain and OOD performance, we now focus on better understanding the individual choices that impact the in-domain and out-of-domain performance of FT. Given that on QQP, most models achieve close to majority accuracy, we focus on MNLI and RTE in the following and present results for QQP in \Cref{appendix:additional-results}.

\paragraph{The role of model selection}

Our FT results in \cref{fig:results-summary} show that many fine-tuned models lead to good out-of-domain generalization. But what is the role of model selection in identifying these checkpoints? To answer this question, we compare selecting the model (a) with the best in-domain performance, (b) at the end of fine-tuning, and (c) with the best out-of-domain performance. \cref{fig:ft-model-selection-mnli} shows the results when fine-tuning on 16 examples. Results for additional sample sizes are shown in \cref{fig:appendix-ft-model-selection-mnli,fig:appendix-ft-model-selection-rte,fig:appendix-ft-model-selection-qqp} in \cref{appendix:additional-results-ft}.

\begin{figure*}[ht] 
    \centering
    \resizebox{0.99\textwidth}{!}{%
  \begin{tabular}{lcccc}
        & \hspace{4mm} \small{16 samples} & \hspace{4mm} \small{32 samples} & \hspace{4mm} \small{64 samples} & \hspace{4mm} \small{128 samples} \\
      \small{MNLI} & 
      \includegraphics[valign=m,width=1.5in]{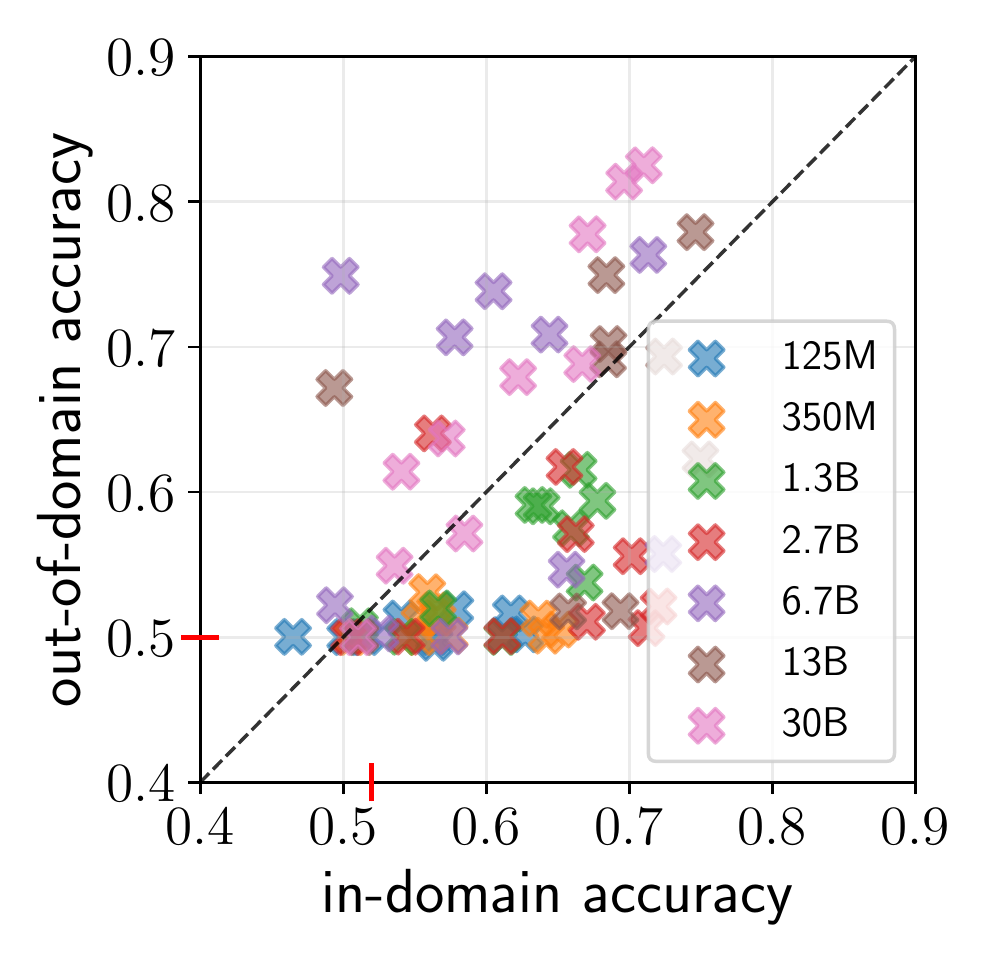} &
      \includegraphics[valign=m,width=1.5in]{figures/ft/all-models_mnli_32_best_out-of-domain_pattern-verbalizer-ft} &
      \includegraphics[valign=m,width=1.5in]{figures/ft/all-models_mnli_64_best_out-of-domain_pattern-verbalizer-ft} &
      \includegraphics[valign=m,width=1.5in]{figures/ft/all-models_mnli_128_best_out-of-domain_pattern-verbalizer-ft} \\
      \small{RTE} & 
      \includegraphics[valign=m,width=1.5in]{figures/ft/all-models_rte_16_best_out-of-domain_pattern-verbalizer-ft} &
      \includegraphics[valign=m,width=1.5in]{figures/ft/all-models_rte_32_best_out-of-domain_pattern-verbalizer-ft} &
      \includegraphics[valign=m,width=1.5in]{figures/ft/all-models_rte_64_best_out-of-domain_pattern-verbalizer-ft} &
      \includegraphics[valign=m,width=1.5in]{figures/ft/all-models_rte_128_best_out-of-domain_pattern-verbalizer-ft} \\
  \end{tabular}
  }%
  \caption{
    Exploring the effect of increasing training examples on FT.
    The first and second rows show results for MNLI and RTE respectively.
    We plot results for 10 runs for each model size and perform model selection according to out-of-domain performance. \textcolor{red}{$\boldsymbol{-}$} in the x- and y-axes indicates majority class accuracy.
    }
    \vspace{-10pt}
    \label{fig:ft-more-data-mnli}
\end{figure*}

Our results show that when performing model selection according to in-domain performance, only the largest models achieve good OOD performance.
On the other hand, when performing model selection according to OOD performance, smaller models can also generalize well (e.g. for the 2.7B model on RTE, 7 out of 10 models have equal or even better OOD than in-domain performance), and this trend persists as model size increases. Interestingly, on RTE, we also observe models with a strong OOD performance when selecting the last checkpoint, which typically leads to poor OOD performance on MNLI.

\paragraph{Training on more data} 

In contrast to ICL, where the maximum number of demonstrations is limited by the context size of a model, FT allows us to perform task adaptation using arbitrary amounts of data. Here, we analyze how the relationship between in-domain and OOD performance is impacted by training on more data. \cref{fig:ft-more-data-mnli} shows the results for MNLI and RTE, and results for QQP are provided in \cref{fig:appendix-ft-model-selection-qqp} in \cref{appendix:additional-results-ft}. For the smallest models, we find that while in-domain performance increases with more training data, OOD performance remains low, which is consistent with previous work \citep{utama-etal-2021-avoiding}. However, for larger models, OOD performance improves as the amount of training data increases and the same trend can be observed when performing model selection according to in-domain performance (see \cref{fig:appendix-ft-model-selection-mnli,fig:appendix-ft-model-selection-rte,fig:appendix-ft-model-selection-qqp} in \cref{appendix:additional-results-ft}).

\paragraph{How much OOD data is needed?}

\begin{figure}[t]
    \centering
    \begin{subfigure}[b]{0.85\columnwidth}
        \centering
        \includegraphics[width=\columnwidth]{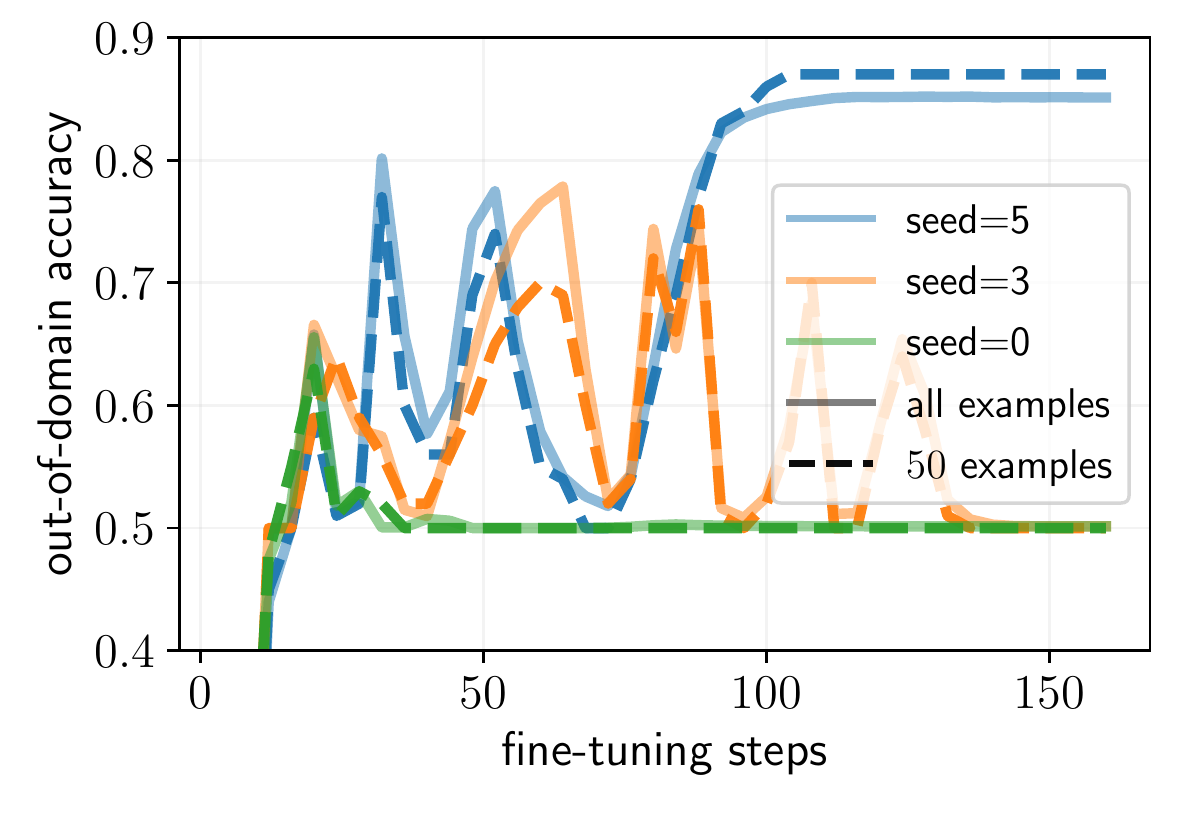}
    \end{subfigure}
    \caption{Estimating OOD performance using less data. We compare OOD performance estimated using all vs. 50 examples when fine-tuning OPT 13B on RTE. Each color corresponds to a run with a different data seed.
    }
    \label{fig:less-ood-data}
\end{figure}

In the experiments so far, we evaluated the models on the full evaluation set (unless mentioned otherwise). Further, we selected FT models based on this evaluation; choosing the best model according to its in-domain or OOD performance in this entire set. 
This setup is not realistic, since in such a scenario where large amounts of data are available for evaluation, it can be used more effectively for training \citep{zhu-etal-2023-weaker}. Hence, in this experiment, we quantify the ability to estimate a model's performance on OOD data using smaller evaluation sets. We fine-tune OPT 13B on MNLI using 128 examples using three different data seeds and plot the OOD generalization in \cref{fig:less-ood-data}. Our results show that using just 50 randomly selected examples is sufficient to distinguish checkpoints that generalize well from those that do not, which would allow us to select, with only these 50 examples, the best OOD checkpoint in a model's training run. This is also reflected in the Pearson correlation of the OOD performance during FT when evaluating it on all vs. 50 examples, which is very high:~$0.99$.

\begin{figure*}[ht] 
    \centering
  \begin{tabular}{lccc}
    & \hspace{4mm} \small{Vanilla FT} & \hspace{4mm} \small{PBFT} & \hspace{4mm} \small{PBFT + LoRA} \\
      & 
      \includegraphics[valign=m,width=1.5in]{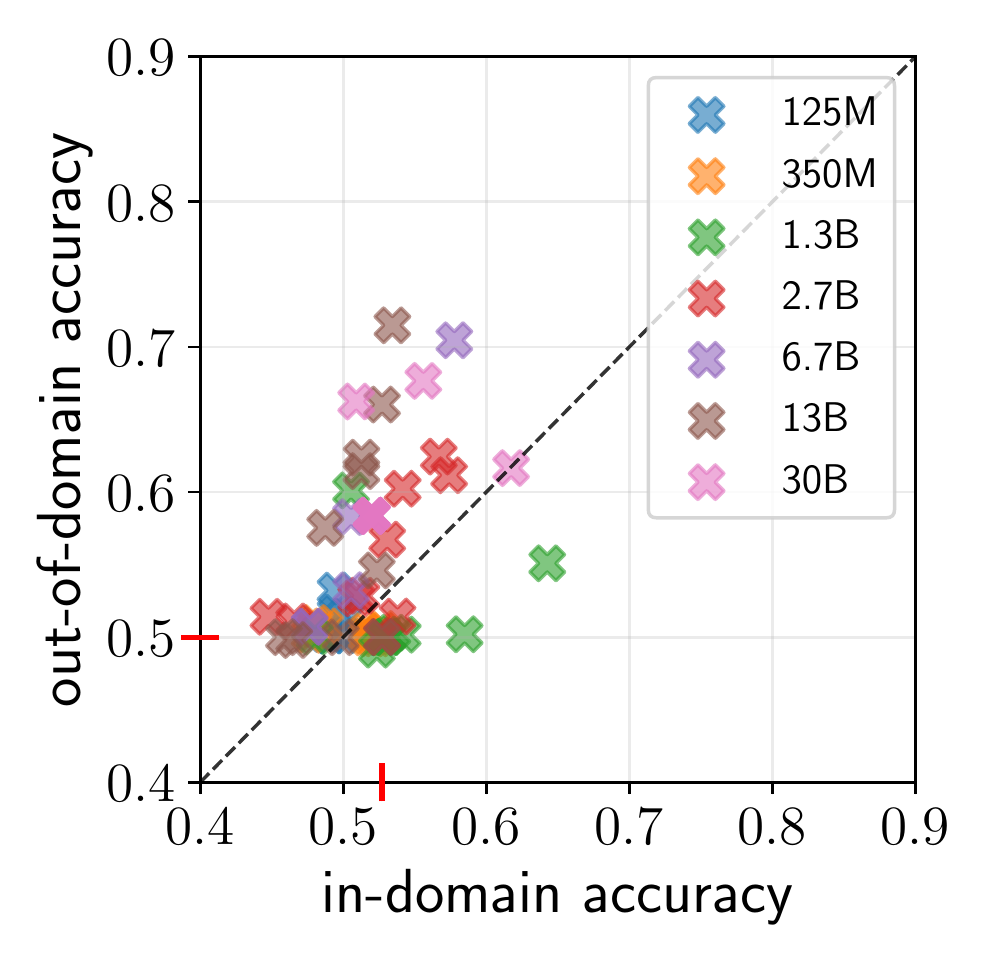} &
      \includegraphics[valign=m,width=1.5in]{figures/ft/all-models_rte_16_best_out-of-domain_pattern-verbalizer-ft.pdf} &
      \includegraphics[valign=m,width=1.5in]{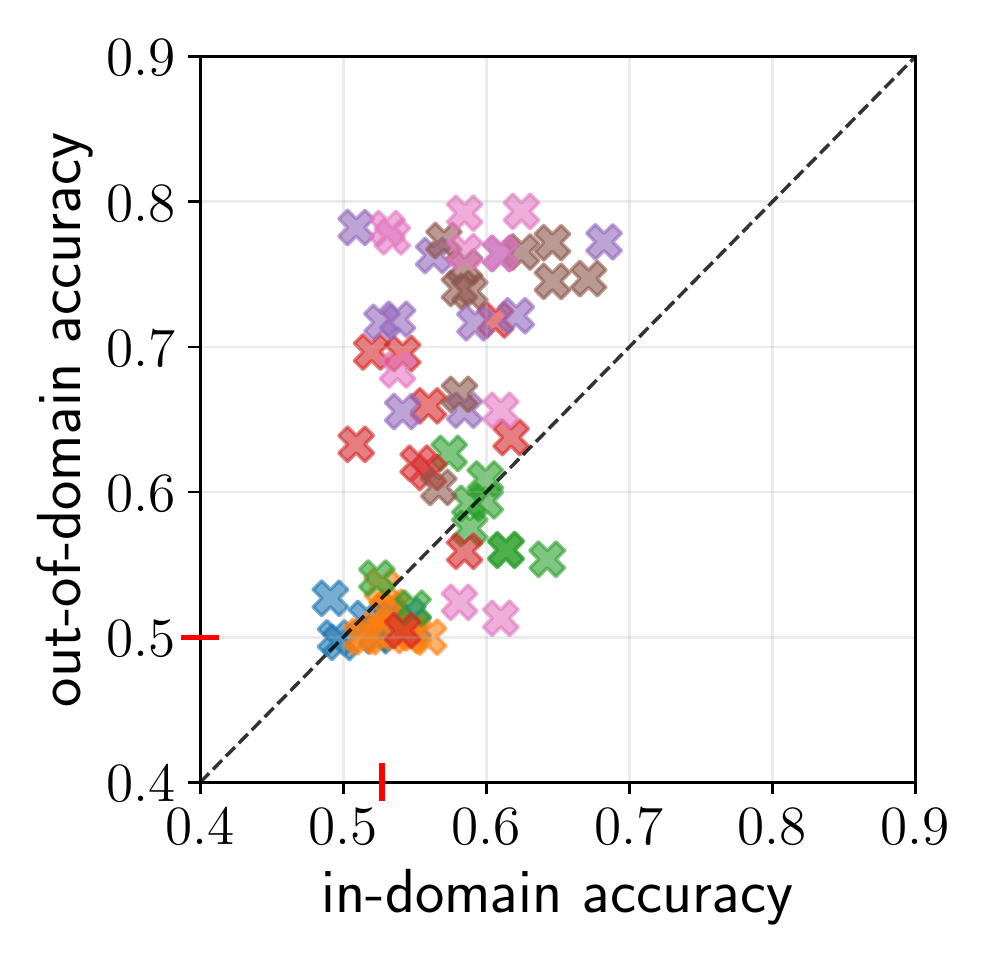} \\
  \end{tabular}
  \caption{Comparing FT approaches on RTE. We use 16 examples and perform model selection according to out-of-domain performance.
  \textcolor{red}{$\boldsymbol{-}$} in the x- and y-axes indicates majority class accuracy.
    }
    \label{fig:ft-comparing-approaches-rte-128}
\end{figure*}
\subsection{Comparing fine-tuning approaches} 

Lastly, we investigate the importance of performing pattern-based FT instead of vanilla FT by fine-tuning a model with a randomly initialized classification head \citep{howard-ruder-2018-universal, devlin-etal-2019-bert}. Further, as an extra fine-tuning strategy, we also apply LoRA \citep{hu2022lora} -- a recently proposed approach for parameter-efficient fine-tuning -- on top of pattern-based FT for comparison. This makes adaptation via FT more similar to adaptation via ICL as it allows the re-use of a large fraction of the weights of a pre-trained language model across tasks.\footnote{We provide more details on both approaches in \cref{sec:appendix-ft-details}.} We fine-tune all models on 16 examples from RTE and present the results in \cref{fig:ft-comparing-approaches-rte-128}. For all FT approaches, we observe a clear improvement in both in-domain and OOD performance as models become larger. Compared to vanilla FT, pattern-based FT leads to better overall performance. When combined with LoRA, pattern-based FT leads to very similar performance as training all parameters. These results demonstrate the generality of our findings beyond a specific FT method. 

\subsection{Our findings generalize beyond OPT}
\label{sec:pythia-results}
\begin{figure}[t]
    \centering
\resizebox{0.99\columnwidth}{!}{%
  \begin{tabular}{cc}
    \small{ICL -- 16 samples} & \small{FT -- 16 samples} \\
      \includegraphics[valign=m,width=1.5in]{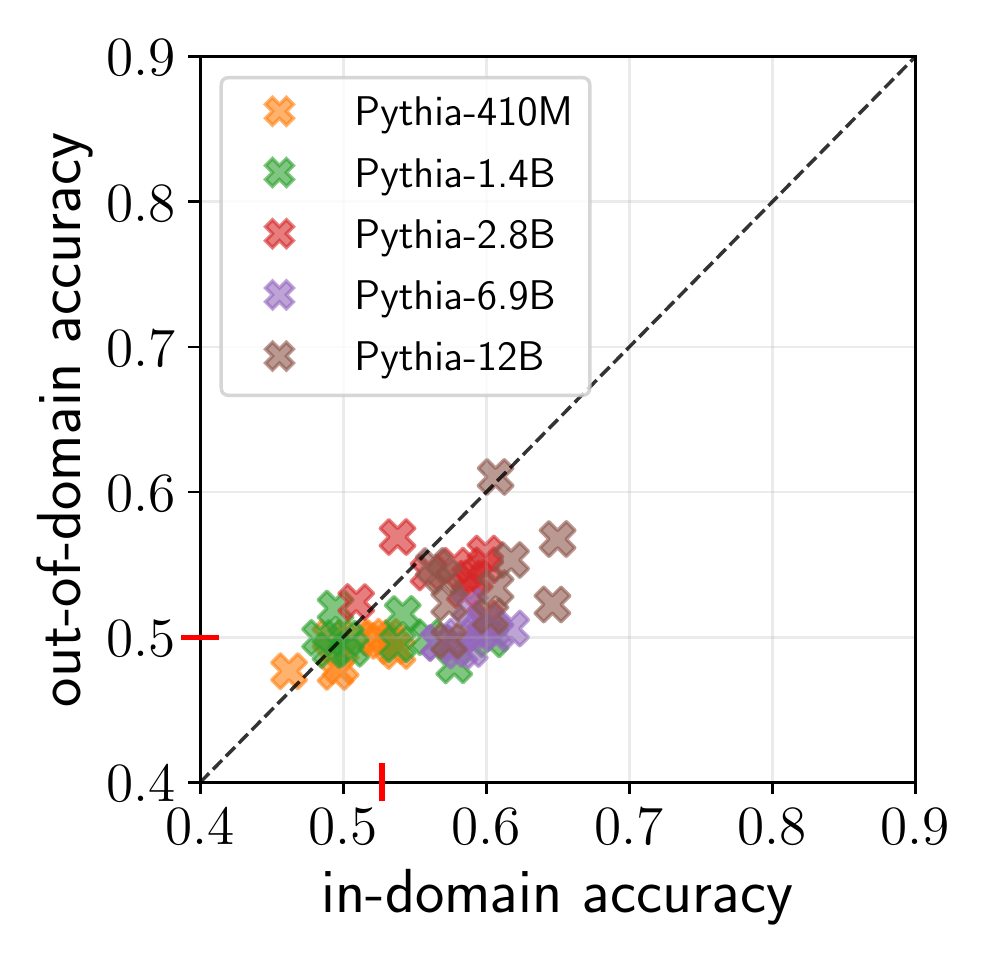} &
      \includegraphics[valign=m,width=1.5in]{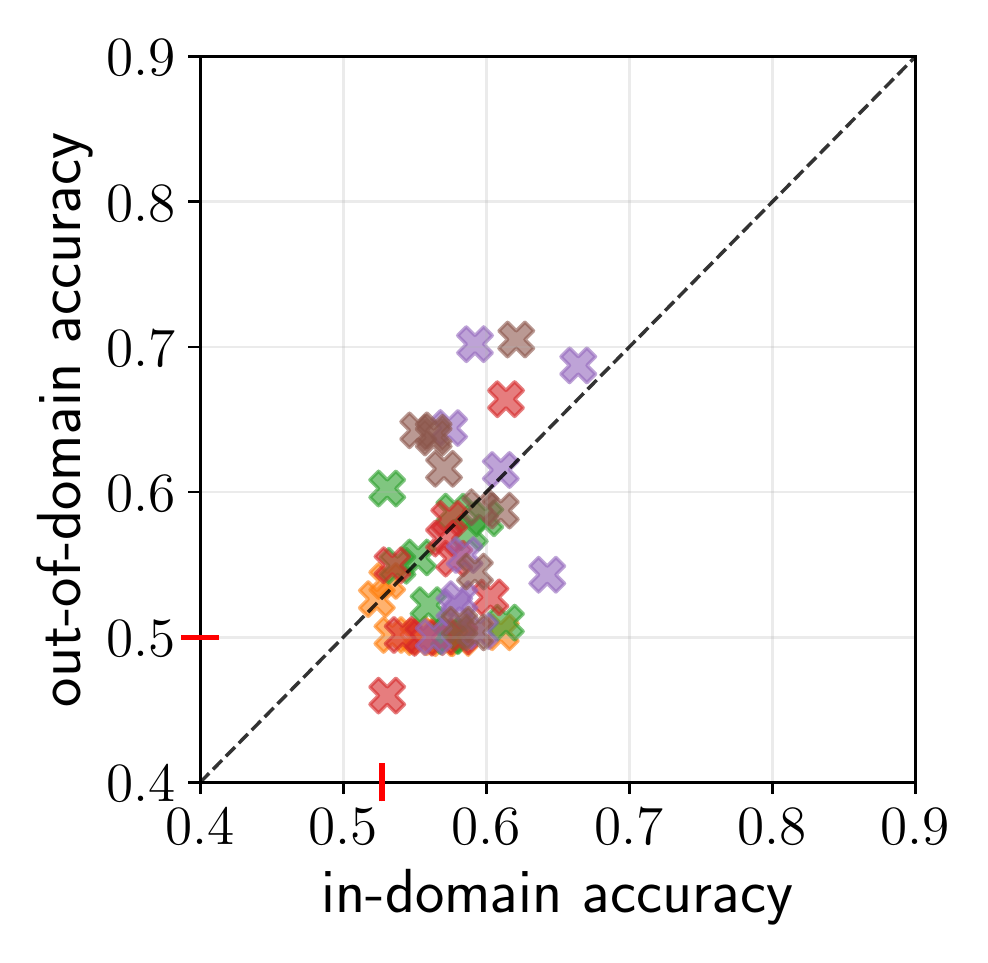} \\
    
  \end{tabular}
  }%
  \caption{ICL and FT results for Pythia models on RTE. We fine-tune models using PBFT. For each approach, we use 16 examples and perform model selection according to 
  in-domain performance. 
  We plot 10 runs per model size which differ only in the data seed.
    \textcolor{red}{$\boldsymbol{-}$} in the x- and y-axes indicates majority class accuracy. 
    } 
    \label{fig:pythia-results}
 \end{figure}

\noindent
\Cref{fig:pythia-results} provides a comparison of ICL and FT using Pythia models\footnote{We use the non-deduped models.} of different sizes ranging from 410M to 12B parameters \citep{biderman2023pythia}. The corresponding significance tests for OOD performance are shown in \Cref{tab:appendix-statistical-tests-pythia-ood-ood} (significance tests for in-domain performance are in \Cref{sec:appendix-pythia}). Similar to OPT, all Pythia models have been trained on the same data, and in the same order. 
We fine-tune using PBFT and select models according to in-domain performance. 
The results for additional patterns, model selection strategies, and sample sizes are discussed in \Cref{sec:appendix-pythia}.

Similarly to OPT, we observe a clear effect of model size on both in-domain and OOD performance. For most model sizes, FT leads to significantly better OOD performance than ICL and both the in-domain and OOD performance of Pythia models improve drastically as we fine-tune on more data (see \Cref{fig:appendix-pythia}). This demonstrates the generality of our findings beyond a single model.\looseness-1

\begin{table}[t]
    \centering
    \resizebox{1.0\columnwidth}{!}{%
    \begin{tabular}{llccccc}
        \toprule
        & 
        & \multicolumn{5}{c}{\textbf{FT}} \\ \cmidrule{3-7} 
        & & \textbf{410M} & \textbf{1.4B} & \textbf{2.8B} & \textbf{6.9B} & \textbf{12B} 
        \\ \midrule
        \multirow{5}{*}{\rotatebox[origin=c]{90}{\textbf{ICL}}} & \textbf{410M}  & \cellcolor{blue!35}$0.02$ & \cellcolor{blue!35}$0.06$ & \cellcolor{blue!35}$0.05$ & \cellcolor{blue!35}$0.09$ & \cellcolor{blue!35}$0.11$ \\
         & \textbf{1.4B}  & $0.01$ & \cellcolor{blue!35}$0.05$ & \cellcolor{blue!35}$0.04$ & \cellcolor{blue!35}$0.08$ & \cellcolor{blue!35}$0.10$ \\
         & \textbf{2.8B} & \cellcolor{red!15}$-0.03$ & $0.01$ & $-0.00$ & $0.04$ & \cellcolor{blue!35}$0.06$ \\
          & \textbf{6.9B} & $0.01$ & \cellcolor{blue!35}$0.05$ & \cellcolor{blue!35}$0.04$ & \cellcolor{blue!35}$0.08$ & \cellcolor{blue!35}$0.10$ \\
         & \textbf{12B} & \cellcolor{red!15}$-0.03$ & $0.01$ & $-0.00$ & $0.04$ & \cellcolor{blue!35}$0.06$ \\
        \bottomrule
    \end{tabular}%
    }
    \caption{Difference between average \textbf{out-of-domain performance} of ICL and FT with Pythia models on RTE. We use 16 examples and 10 random seeds for both approaches. For ICL, we use the \texttt{gpt-3} pattern. For FT, we use pattern-based fine-tuning (PBFT) and select checkpoints according to in-domain performance. We perform a Welch's t-test and color cells according to whether: \textcolor{red!45}{ICL performs significantly better than FT}, \textcolor{blue!75}{FT performs significantly better than ICL}. For cells without color, there is no significant difference.\looseness-1
    } \label{tab:appendix-statistical-tests-pythia-ood-ood}
\end{table}
\section{Discussion}
\label{sec:discussion}

Our findings in the previous section demonstrate that fine-tuned language models can generalize OOD too, highlighting the importance of comparing adaptation approaches fairly. In this section, we present further insights from our experiments and provide a high-level comparison of the pros and cons of adaptation via ICL and FT.\looseness-1

\paragraph{What signal to learn from?} 

Both our ICL and FT results exhibit a large variance in both in-domain and OOD performance. Our results show different OOD behavior during FT when varying only the data seed. In addition, as previous work has shown, the choice of patterns and verbalizers impact both ICL and PBFT performance in unintuitive ways. For instance, \citet{webson-pavlick-2022-prompt} find that pattern-based fine-tuned models perform well even when using misleading patterns. Here, we find that ICL's generalization is heavily dependent on the choice of pattern and verbalizer. This shows the importance of the choice of training data and patterns for task adaptation.\looseness-1

\paragraph{Advances in task adaptation}

The success of ICL led to the development of new methods for improving on it further, such as calibration \citep{pmlr-v139-zhao21c}, and chain-of-thought prompting \citep{wei2022chain}. In this work, we focus on the `vanilla' version of ICL and the fine-tuning approach most similar to it -- pattern-based fine-tuning. Our results suggest that these two approaches are more similar than previously thought, as they achieve similar performance both in-domain and OOD. As such, new methods for ICL can also be applied to PBFT, and we expect them to achieve similar results.

\paragraph{Analyzing the fine-tuning loss surface}

Looking at the OOD generalization curves throughout fine-tuning (in \cref{fig:less-ood-data} and additional plots in \cref{sec:appendix-individual-ft-runs}), we observe that for some runs, OOD performance fluctuates heavily and models change their generalization `strategy' during FT. In \cref{fig:less-ood-data}, we can see that some fine-tuning runs undergo a dramatic change in OOD performance after 75 steps. We leave it to future work to further study this behavior and the relationship between the FT loss surface and OOD generalization \cite{shwartz-ziv2022pretrain, juneja2023linear}.\looseness=-1

\section{Comparing FT and ICL}
\label{sec:ft-icl-comparison}
\begin{table}[t]
    \centering
    \resizebox{1.\columnwidth}{!}{%
    \begin{tabular}{lll}
        \toprule
                          \textbf{Feature}  & \textbf{FT}              & \textbf{ICL}         \\ \midrule
            \multirow{2}{*}{Users}          & Experts         & Experts \& \\
            & & Non-experts \\
            Interaction    & Pre-defined    & Textual \\
            Reusability  & Medium & High \\
            Applicability to  & \multirow{2}{*}{High} & \multirow{2}{*}{Limited} \\
            low-resource languages & & \\
            \midrule
            Requires training        & Yes           & No \\
            \multirow{2}{*}{Inference time}  & \multirow{2}{*}{$|$test example$|$}    & $|$test example$|$ + \\
            & & $|$demonstrations$|$ \\ 
            $|$Demonstrations$|$ & Unlimited       & $\leq$100   \\
            Variance    & High & High  \\
            \midrule
            SOTA           & Yes             & Yes         \\
            Size scaling & Standard & Standard \\
            $|$Demonstrations$|$ scaling & Standard & Limited \\
            Invented & 2018 & 2020 \\
            Well understood & No & No \\
        \bottomrule
    \end{tabular}
    }%
    \caption{A high-level comparison between key features of fine-tuning and in-context learning.}
    \label{tab:comparison}
    \vspace{-0.1in}
\end{table}

\noindent
This section examines the key features for task adaptation and compares FT and ICL. We summarize our findings in \Cref{tab:comparison}. We begin by discussing features related to user interaction, which can be found in the first part of the table. FT requires expertise in model training, whereas ICL only requires natural language, i.e., non-experts can use this approach more easily.
ICL is also highly reusable as it does not modify the pre-trained model and hence, the same model can be used for many tasks; FT, however, is not as reusable (with the exception of parameter-efficient methods) and typically results in a specialized model per task. 
Unfortunately, despite its user-friendliness and reusability, ICL does not work out of the box for some tasks which require more sophisticated prompting \citep{wei2022chain}.\looseness-1 

ICL requires large models to work in contrast to FT, which works well even with small models \citep{devlin-etal-2019-bert}. This hinders the applicability of ICL to models developed for low-resource languages, as training billion parameter-scale models requires huge amounts of training data, which are simply unavailable for many languages.
As such, FT is still the dominating adaptation approach in this setting \citep[\emph{inter alia}]{pfeiffer-etal-2022-lifting, alabi-etal-2022-adapting}.\looseness=-1 

Next, we compare technical details regarding the training and inference of such approaches.
While FT requires training (which when dealing with large models can become expensive), ICL does not. On the other hand, the inference time of fine-tuned models is much smaller than ICL, since it only includes the time that it takes to process the minimal pattern and the test instance. When using ICL, each test instance has to include all of the demonstrations as well, which increases the inference time. The fixed context size of the model also limits the number of demonstrations that can be used\footnote{Note that some methods allow an infinite context \cite[e.g.][]{press2022train, martins-etal-2022-former}. Most current successful LMs, however, have limited context sizes.}\looseness=-1, while FT allows for unlimited training examples. We show in this work that both methods can achieve strong performance on both in-domain and OOD datasets.
Both approaches improve with model size, but FT benefits more from additional samples than ICL does, as was also shown in previous work  \cite{min-etal-2022-rethinking}.

Finally, we highlight that both methods are relatively recent: vanilla FT was invented in 2018 \cite{howard-ruder-2018-universal} and ICL in 2020 \cite{brown-etal-2020-language}.\footnote{PBFT was also invented in 2020 \citep{schick-schutze-2021-exploiting}.\looseness=-1}
As such, these methods are still poorly understood, and more research is required on both approaches to better understand their strengths and weaknesses.\looseness=-1

\section{Related work}
\label{sec:related-work}

\citet{brown-etal-2020-language} compare GPT-3's few-shot in-context learning performance with fine-tuned language models trained in the fully supervised setting, finding that both approaches lead to similar results in question answering. However, the fine-tuned models they compare ICL to are smaller models, making the task adaptation comparison unfair. For SuperGLUE, while using smaller models, they find that FT largely outperforms ICL. This is consistent with our findings. Even in the few-shot setting, fine-tuned language models can outperform ICL when comparing models of the same size. Recently, \citet{liu2022fewshot} compared parameter-efficient few-shot FT of T0 \citep{sanh-etal-2022-multitask} to ICL with GPT-3, finding that their parameter-efficient FT approach outperforms ICL. This is consistent with our findings; however, unlike our work, they only consider in-domain performance. %

Focusing on OOD performance, \citet{si2023prompting} investigate the generalization of GPT-3 along various axes, including generalization under covariate shift -- as we do. However, they compare models of different sizes, i.e., RoBERTa-large and GPT-3 (which has 500 times the number of parameters), and different training settings, i.e., fully supervised for FT vs. few-shot for ICL. They observe much better OOD performance for ICL than FT, concluding that ICL with GPT-3 is more robust than FT using BERT or RoBERTa. While this conclusion is valid, it holds for specific models, rather than the methods in general. We show how important it is to compare methods fairly. Based on our comparable results, fine-tuning language models results in similar or even better OOD generalization. Another work that compares the OOD generalization of different adaptation approaches is \citet{awadalla-etal-2022-exploring}. Unlike our choice of MNLI and RTE, they investigate the robustness of question answering models under various types of distribution shifts and find that ICL is more robust to distribution shifts than FT. Moreover, they argue that for FT, increasing model size does not have a strong impact on generalization. However, they don't scale beyond 1.5B parameters. Our findings suggest that the relationship between in-domain and OOD performance does depend on model size.

While we focus on the task adaptation of decoder-only models, \citet{utama-etal-2021-avoiding} investigate the OOD generalization of encoder-only models adapted via pattern-based few-shot FT. For MNLI and HANS, they find that these models adopt similar inference heuristics to those trained with vanilla FT and hence perform poorly OOD. They observe that models rely even more on heuristics when fine-tuned on more data. This is in contrast to our results where we find that pattern-based few-shot FT can lead to good OOD generalization, and OOD generalization improves as we train on more data. We attribute this to the fact that they experiment with a smaller model (RoBERTa-large; 350M).\footnote{This is also related to \citeposs{warstadt-etal-2020-learning} results, who show that better pre-trained models are less prone to rely on superficial (and potentially spurious) features for predictions.\looseness=-1}
Lastly, \citet{bandel-etal-2022-lexical} show that masked language models can generalize well on HANS if fine-tuned for a sufficient number of steps. While they focus on fine-tuning on the entire dataset, their findings provide additional evidence that fine-tuned language models can generalize well OOD.\looseness-1

\section{Conclusion}
\label{sec:conclusion}

We perform a fair comparison between in-domain and OOD generalization of two alternative task adaptation strategies: Few-shot ICL and FT. We compare OPT models \cite{zhang2022opt} ranging from 125M to 30B parameters on three classification datasets across two tasks. We find that for both approaches, performance improves as models become larger. For the largest models we experiment with (OPT-30B), we find that FT outperforms ICL on both in-domain and OOD performance and even improves further as we train on more data. %
However, our results also demonstrate that the performance of both FT and ICL exhibits high variance, highlighting that truly robust task adaptation remains an open challenge.
We end by providing a high-level comparison between the two approaches, listing the benefits and limitations of each, and discussing some future directions.\looseness-1

\section{Limitations}
\label{sec:limitations}

In this work, we focus on a specific type of OOD generalization, namely, covariate shift \citep{hupkes-etal-2022-state-of}. Under this setup, we refer to OOD as the specific challenge datasets we use. As such, different conclusions might be reached by repeating the experiments and evaluating different datasets.

We focus specifically on OPT decoder-only models as the goal of our work is to compare the generalization of adaptation via fine-tuning vs. in-context learning using the same pre-trained model.
To the best of our knowledge, existing encoder-only models do not have strong in-context learning abilities. For encoder--decoder models such as T5, only recent variants such as Flan-T5 \citep{chung2022scaling} demonstrate the ability to respond well to instructions. However, these models require an additional supervised fine-tuning step on instruction data. This makes it challenging to attribute generalization abilities (or the lack thereof) to specific adaptation techniques (fine-tuning vs in-context learning). 
Hence, we focus on decoder-only models pre-trained exclusively with a language modeling objective. 

Many recent papers that experiment with in-context learning use GPT-3. While fine-tuning GPT-3 is possible via an API, it is unclear what fine-tuning approach is used behind that API. Since this makes a fair comparison difficult, we chose not to experiment with GPT-3.

While similarly large models (e.g. OPT-175B) are publicly available, we do not have the computational resources to run such models. While we expect the trends we observe in this work to hold with larger models, we are not able to empirically test that. Moreover, we only experiment with English language models as, to the best of our knowledge, there are no publicly available models which are similar to OPT (decoder-only models of various sizes trained on the same data) for other languages.\looseness=-1

Finally, we only experiment with basic FT and ICL methods. However, for both approaches there exist more advanced techniques which we do not consider \citep[e.g. calibration;][]{pmlr-v139-zhao21c}. We note that such techniques can typically be applied for both adaptation approaches. Hence we expect an improvement for one method to improve the other as well.

\section*{Acknowledgments}
\label{sec:acknowledgments}

We are grateful to Vagrant Gautam for their valuable feedback and patience when proofreading our work. We also thank Badr Abdullah for his help with proofreading and feedback during early stages of this work. Marius Mosbach acknowledges funding from the Deutsche Forschungsgemeinschaft (DFG, German Research Foundation) –
Project-ID 232722074 – SFB 1102.

\bibliography{anthology,custom}
\bibliographystyle{acl_natbib}

\newpage
\appendix

\onecolumn

\section{Experimental details}
\label{appendix:experimental-details}

We access all models via huggingface transformers \citep{wolf-etal-2020-transformers} and use its DeepSpeed \citep{deepspeed} integration for efficient distributed training and evaluation.

\subsection{Hardware}

We run our experiments on 8x A100 GPUs with 80GB of memory.

\subsection{Label distribution}

Table \ref{tab:majority_baselines} shows the accuracy of the majority class label on each of the datasets. Note that MNLI (when merging the neutral and contradiction classes) and PAWS-QQP are highly unbalanced.

\begin{table}[h]
    \centering
    \resizebox{0.55\textwidth}{!}{%
    \begin{tabular}{lc}
        \toprule
         \textbf{Dataset} & \textbf{Majority class} \\ \midrule
         MNLI (remove neutral) &  $0.512$ \\ %
         MNLI (merge neutral and contradiction) &  $0.645$ \\ %
         RTE &  $0.527$ \\ %
         QQP &  $0.632$ \\ \midrule
         HANS & $0.500$ \\ %
         PAWS-QQP & $0.718$ \\ %
        \bottomrule
    \end{tabular}
    }%
    \caption{Accuracy of the majority class label for each dataset.}
    \label{tab:majority_baselines}
\end{table}

\subsection{In-context learning: Additional details}
\label{sec:appendix:icl-details}

\paragraph{Patterns} We present the patterns used for ICL in Table \ref{tab:appendix-in-context-learning-patterns}. We obtain the \texttt{GPT-3} pattern from \citet{brown-etal-2020-language}. The \texttt{eval-harness} pattern is based on \citet{eval-harness}. 

\begin{table*}[h]
    \centering
    \resizebox{1.0\textwidth}{!}{%
    \begin{tabular}{lllll}
        \toprule
         \textbf{Dataset(s)} & \textbf{Pattern} & \textbf{Text} & \textbf{Answer prefix} & \textbf{Target tokens} \\ \midrule
         MNLI, RTE & \texttt{minimal} & \{premise\} \{hypothesis\} ? & -- & Yes, No \\
         MNLI, RTE& \texttt{gpt-3} & \{premise\} question: \{hypothesis\} Yes or No? & answer: & Yes, No \\
         MNLI, RTE & \texttt{eval-harness} & \{premise\} \textbackslash n Question: \{hypothesis\} True or False? & \textbackslash n Answer: & True, False\\ \midrule
         QQP & \texttt{minimal} & \{question 1\} \{question 2\} ? & -- & Yes, No\\
         \multirow{2}{*}{QQP} & \multirow{2}{*}{\texttt{eval-harness}} & Question 1: \{question 1\} \textbackslash n Question 2:\{question 2\} & \multirow{2}{*}{Answer:} & \multirow{2}{*}{Yes, No} \\
          &  & \textbackslash n Question: Do both questions ask the same thing? & \\
         \bottomrule
    \end{tabular}
    }%
    \caption{Patterns used for ICL. The minimal patterns are used for PBFT as well.}
    \label{tab:appendix-in-context-learning-patterns}
\end{table*}

\subsection{In-context learning: Comparison with previous work}
\label{sec:appendix-icl-previous}

\Cref{tab:icl-previous-work} compares our ICL results to results from previous work. On RTE and MNLI we achieve comparable performance to previous work. On QQP, our ICL results are much worse (and very close to the majority class classifier). We hypothesize that this is due to the difference in model size (GPT-3 with 175B parameters vs. OPT with 30B parameters) and hence focus on MNLI and RTE for most of our experiments.

\begin{table}[h]
    \centering
    \resizebox{0.55\textwidth}{!}{%
    \begin{tabular}{llcc}
        \toprule
         \textbf{Model} & \textbf{Dataset} & \textbf{In-domain} & \textbf{Out-of-domain} \\ \midrule
         GPT-3 175B  & MNLI & $77.6$  & $75.3$ \\ %
         OPT 30B & RTE & $62.0$ & -- \\ %
         GPT-3 175B  & QQP & $83.5$ & $73.7$ \\ \midrule
         OPT 30B & MNLI & $71.4$ ($74.9$) & $56.7$ ($72.3$) \\ %
         OPT 30B & RTE & $61.7$ ($66.8$) & $60.5$ ($65.4$) \\ %
         OPT 30B & QQP & $42.0$ ($63.1$) & $49.8$ ($53.3$) \\
        \bottomrule
    \end{tabular}
    }%
    \caption{Comparing ICL results from previous work (first three rows) with ours (last three rows). In our results we report average and maximum performance (in parentheses) of the largest model. Previous results are from \citet{si2023prompting} for GPT-3 and \citet{zhang2022opt} for OPT.
    }
    \label{tab:icl-previous-work}
\end{table}
\subsection{Fine-tuning: Additional details}
\label{sec:appendix-ft-details}

\paragraph{Vanilla FT}
Vanilla FT \citep{howard-ruder-2018-universal, devlin-etal-2019-bert} is one of the most commonly used task adaptation approaches for pre-trained language models.
During FT we typically: (i) replace the model's language modeling head with a new randomly initialized classification head; (ii) update all model parameters, as well as the new head's, on the downstream task's training data.\footnote{We will refer to any FT approach that uses a randomly initialized classifier as vanilla FT.}
When trained on entire datasets, fine-tuned language models dominate academic leaderboards, such as GLUE \cite{wang-etal-2018-glue} and SuperGLUE \cite{wang-etal-2019-superglue}. %
However, despite their strong in-domain performance, fine-tuned language models tend to generalize poorly OOD, which is often attributed to adopting inference heuristics during FT \citep{mccoy-etal-2019-right,elazar-etal-2021-back}. %

\paragraph{Parameter-efficient FT} Parameter-efficient FT methods update only a small number of parameters relative to the total number of parameters of the pre-trained model \citep[\emph{inter alia}]{pmlr-v97-houlsby19a, ben-zaken-etal-2022-bitfit, hu2022lora}.
Such approaches can be applied to either vanilla or prompt-based FT and are appealing since they allow large parts of a model to be re-used across tasks. %

\begin{table}[ht]
    \centering
    \resizebox{0.55\textwidth}{!}{%
    \begin{tabular}{ll}
        \toprule
         \textbf{Hyperparameter} & \textbf{Value} \\ \midrule
         Optimizer & AdamW \\
         Learning rate &  $10^{-5}$ \\ %
         Learning rate schedule &  linear warmup then constant \\ %
         Warmup ratio &  $10\%$ of total steps \\ %
         Weight decay & $0.0$ \\ %
         Dropout & $0.1$ \\ %
         Batch size & 32 \\
         Epochs & 40 \\
         Total steps & $\frac{\text{\#samples}}{\text{batch size}} * \text{epochs} $ \\
        \bottomrule
    \end{tabular}
    }
    \caption{FT hyperparameters.}

    \label{tab:hyperparameters}
\end{table}

\paragraph{Hyperparameters} Table \ref{tab:hyperparameters} provides an overview of all hyperparameters used during FT.

\section{Additional results for OPT models}%
\label{appendix:additional-results}

\subsection{Significance tests}
\label{appendix:significance-tests}

\Cref{tab:appendix-statistical-tests-in-domain-in-domain,tab:appendix-statistical-tests-ood-in-domain,tab:appendix-statistical-tests-in-domain-ood,tab:appendix-statistical-tests-ood-ood} show the results of a Welch's t-test comparing the average in-domain and out-of-domain performance of ICL and PBFT on RTE and MNLI. We use 16 samples and 10 different seeds for each approach and consider a p-value of 0.05 to be statistically significant. For FT, we compare two different approaches of model selection: (1) based on in-domain performance and (2) based on out-of-domain performance (note that these are the same models as those shown in the first row of \cref{fig:teaser}). 

For RTE, our results show that ICL outperforms FT only when comparing large models to smaller models. However, when comparing models of the same size, FT performs at least equally well to ICL, and in some cases even significantly better. For MNLI, for larger models (6.7B onwards) ICL outperforms FT in terms of in-domain performance. Looking at OOD performance, however, we again see that ICL only outperforms FT when comparing large models to much smaller models.

\begin{table*}
    \begin{subtable}{.49\textwidth}\centering
        {
            \resizebox{1\textwidth}{!}{%
            \begin{tabular}{llccccccc}
        \toprule
        & 
        & \multicolumn{7}{c}{\textbf{FT}} \\ \cmidrule{3-9} 
        & & \textbf{125M} & \textbf{350M} & \textbf{1.3B} & \textbf{2.7B} & \textbf{6.7B} & \textbf{13B} & \textbf{30B} 
        \\ \midrule
        \multirow{7}{*}{\rotatebox[origin=c]{90}{\textbf{ICL}}} & \textbf{125M}  & \cellcolor{blue!35}$\phantom{-}0.03$ & \cellcolor{blue!35}$\phantom{-}0.04$ & \cellcolor{blue!35}$\phantom{-}0.08$ & \cellcolor{blue!35}$\phantom{-}0.11$ & \cellcolor{blue!35}$\phantom{-}0.10$ & \cellcolor{blue!35}$\phantom{-}0.09$ & \cellcolor{blue!35}$\phantom{-}0.10$  \\
         & \textbf{350M}  & \cellcolor{blue!35}$\phantom{-}0.03$ & \cellcolor{blue!35}$\phantom{-}0.05$ & \cellcolor{blue!35}$\phantom{-}0.08$ & \cellcolor{blue!35}$\phantom{-}0.11$ & \cellcolor{blue!35}$\phantom{-}0.10$ & \cellcolor{blue!35}$\phantom{-}0.10$ & \cellcolor{blue!35}$\phantom{-}0.10$ \\
         & \textbf{1.3B} & \cellcolor{blue!35}$\phantom{-}0.03$ & \cellcolor{blue!35}$\phantom{-}0.04$ & \cellcolor{blue!35}$\phantom{-}0.08$ & \cellcolor{blue!35}$\phantom{-}0.10$ & \cellcolor{blue!35}$\phantom{-}0.10$ & \cellcolor{blue!35}$\phantom{-}0.09$ & \cellcolor{blue!35}$\phantom{-}0.09$  \\
          & \textbf{2.7B} & $\phantom{-}0.00$ & $\phantom{-}0.02$ & \cellcolor{blue!35}$\phantom{-}0.05$ & \cellcolor{blue!35}$\phantom{-}0.08$ & \cellcolor{blue!35}$\phantom{-}0.07$ & \cellcolor{blue!35}$\phantom{-}0.07$ & \cellcolor{blue!35}$\phantom{-}0.07$  \\
         & \textbf{6.7B} & \cellcolor{red!15}$-0.06$ & \cellcolor{red!15}$-0.04$ & $-0.01$ & $\phantom{-}0.02$ & $\phantom{-}0.01$ & $\phantom{-}0.01$ & $\phantom{-}0.01$  \\
         & \textbf{13B} & \cellcolor{red!15}$-0.06$ & \cellcolor{red!15}$-0.05$ & $-0.01$ & $\phantom{-}0.02$ & $\phantom{-}0.01$ & $\phantom{-}0.00$ & $\phantom{-}0.01$ \\
         & \textbf{30B} & \cellcolor{red!15}$-0.06$ & \cellcolor{red!15}$-0.04$ & $-0.01$ & $\phantom{-}0.02$ & $\phantom{-}0.01$ & $\phantom{-}0.01$ & $\phantom{-}0.01$  \\
         
        \bottomrule
    \end{tabular}%
    }
        }
        \caption{RTE}\label{tab:appendix-statistical-tests-in-domain-in-domain-rte}
    \end{subtable}%
    ~~ 
    \begin{subtable}{.49\textwidth}\centering
        {
            \resizebox{1\textwidth}{!}{%
            \begin{tabular}{llccccccc}
        \toprule
        & 
        & \multicolumn{7}{c}{\textbf{FT}} \\ \cmidrule{3-9} 
        & & \textbf{125M} & \textbf{350M} & \textbf{1.3B} & \textbf{2.7B} & \textbf{6.7B} & \textbf{13B} & \textbf{30B} 
        \\ \midrule
        \multirow{7}{*}{\rotatebox[origin=c]{90}{\textbf{ICL}}} & \textbf{125M}  & \cellcolor{blue!35}$\phantom{-}0.07$ & \cellcolor{blue!35}$\phantom{-}0.09$ & \cellcolor{blue!35}$\phantom{-}0.13$ & \cellcolor{blue!35}$\phantom{-}0.14$ & \cellcolor{blue!35}$\phantom{-}0.12$ & \cellcolor{blue!35}$\phantom{-}0.17$ & \cellcolor{blue!35}$\phantom{-}0.13$  \\
         & \textbf{350M}  & \cellcolor{blue!35}$\phantom{-}0.05$ & \cellcolor{blue!35}$\phantom{-}0.07$ & \cellcolor{blue!35}$\phantom{-}0.11$ & \cellcolor{blue!35}$\phantom{-}0.13$ & \cellcolor{blue!35}$\phantom{-}0.11$ & \cellcolor{blue!35}$\phantom{-}0.16$ & \cellcolor{blue!35}$\phantom{-}0.11$ \\
         & \textbf{1.3B} & $-0.02$ & $-0.00$ & $\phantom{-}0.03$ & \cellcolor{blue!35}$\phantom{-}0.05$ & $\phantom{-}0.03$ & \cellcolor{blue!35}$\phantom{-}0.08$ & $\phantom{-}0.03$  \\
          & \textbf{2.7B} & $\phantom{-}0.01$ & \cellcolor{blue!35}$\phantom{-}0.03$ & \cellcolor{blue!35}$\phantom{-}0.07$ & \cellcolor{blue!35}$\phantom{-}0.08$ & \cellcolor{blue!35}$\phantom{-}0.06$ & \cellcolor{blue!35}$\phantom{-}0.11$ & \cellcolor{blue!35}$\phantom{-}0.06$  \\
         & \textbf{6.7B} & \cellcolor{red!15}$-0.06$ & \cellcolor{red!15}$-0.04$ & $-0.00$ & $\phantom{-}0.01$ & $-0.01$ & \cellcolor{blue!35}$\phantom{-}0.04$ & $-0.00$  \\
         & \textbf{13B} & \cellcolor{red!15}$-0.13$ & \cellcolor{red!15}$-0.11$ & \cellcolor{red!15}$-0.07$ & \cellcolor{red!15}$-0.06$ & \cellcolor{red!15}$-0.08$ & $-0.03$ & \cellcolor{red!15}$-0.08$  \\
         & \textbf{30B} & \cellcolor{red!15}$-0.11$ & \cellcolor{red!15}$-0.09$ & \cellcolor{red!15}$-0.05$ & \cellcolor{red!15}$-0.04$ & \cellcolor{red!15}$-0.06$ & $-0.01$ & \cellcolor{red!15}$-0.06$  \\
         
        \bottomrule
    \end{tabular}%
    }
        }
        \caption{MNLI}\label{tab:appendix-statistical-tests-in-domain-in-domain-mnli}
    \end{subtable}%
    \caption{Difference between average \textbf{in-domain performance} of ICL and FT on RTE (a) and MNLI (b) across model sizes. We use 16 examples and 10 random seeds for both approaches. For ICL, we use the \texttt{gpt-3} pattern. For FT, we use pattern-based fine-tuning (PBFT) and select checkpoints according to \underline{in-domain performance}. We perform a Welch's t-test and color cells according to whether: \textcolor{red!45}{ICL performs significantly better than FT}, \textcolor{blue!75}{FT performs significantly better than ICL}. For cells without color, there is no significant difference between ICL and FT.\looseness-1
    } \label{tab:appendix-statistical-tests-in-domain-in-domain}
\end{table*}
\begin{table*}
    \begin{subtable}{.49\textwidth}\centering
        {
            \resizebox{1\textwidth}{!}{%
            \begin{tabular}{llccccccc}
        \toprule
        & 
        & \multicolumn{7}{c}{\textbf{FT}} \\ \cmidrule{3-9} 
        & & \textbf{125M} & \textbf{350M} & \textbf{1.3B} & \textbf{2.7B} & \textbf{6.7B} & \textbf{13B} & \textbf{30B} 
        \\ \midrule
        \multirow{7}{*}{\rotatebox[origin=c]{90}{\textbf{ICL}}} & \textbf{125M}  & $-0.01$ & \cellcolor{blue!35}$\phantom{-}0.02$ & \cellcolor{blue!35}$\phantom{-}0.05$ & \cellcolor{blue!35}$\phantom{-}0.05$ & \cellcolor{blue!35}$\phantom{-}0.07$ & \cellcolor{blue!35}$\phantom{-}0.07$ & \cellcolor{blue!35}$\phantom{-}0.07$  \\
         & \textbf{350M}  & $-0.01$ & \cellcolor{blue!35}$\phantom{-}0.02$ & \cellcolor{blue!35}$\phantom{-}0.05$ & \cellcolor{blue!35}$\phantom{-}0.05$ & \cellcolor{blue!35}$\phantom{-}0.08$ & \cellcolor{blue!35}$\phantom{-}0.08$ & \cellcolor{blue!35}$\phantom{-}0.07$ \\
         & \textbf{1.3B} & $-0.01$ & $\phantom{-}0.01$ & \cellcolor{blue!35}$\phantom{-}0.04$ & \cellcolor{blue!35}$\phantom{-}0.04$ & \cellcolor{blue!35}$\phantom{-}0.07$ & \cellcolor{blue!35}$\phantom{-}0.07$ & \cellcolor{blue!35}$\phantom{-}0.06$  \\
          & \textbf{2.7B} & \cellcolor{red!15}$-0.04$ & $-0.01$ & $\phantom{-}0.02$ & $\phantom{-}0.02$ & \cellcolor{blue!35}$\phantom{-}0.05$ & \cellcolor{blue!35}$\phantom{-}0.05$ & \cellcolor{blue!35}$\phantom{-}0.04$  \\
         & \textbf{6.7B} & \cellcolor{red!15}$-0.09$ & \cellcolor{red!15}$-0.07$ & \cellcolor{red!15}$-0.04$ & \cellcolor{red!15}$-0.04$ & $-0.01$ & $-0.01$ & $-0.02$  \\
         & \textbf{13B} & \cellcolor{red!15}$-0.10$ & \cellcolor{red!15}$-0.07$ & \cellcolor{red!15}$-0.04$ & \cellcolor{red!15}$-0.04$ & $-0.02$ & $-0.02$ & $-0.02$ \\
         & \textbf{30B} & \cellcolor{red!15}$-0.10$ & \cellcolor{red!15}$-0.07$ & \cellcolor{red!15}$-0.04$ & \cellcolor{red!15}$-0.04$ & $-0.01$ & $-0.01$ & $-0.02$  \\
         
        \bottomrule
    \end{tabular}%
    }
        }
        \caption{RTE}\label{tab:appendix-statistical-tests-ood-in-domain-rte}
    \end{subtable}%
    ~~ 
    \begin{subtable}{.49\textwidth}\centering
        {
            \resizebox{1\textwidth}{!}{%
            \begin{tabular}{llccccccc}
        \toprule
        & 
        & \multicolumn{7}{c}{\textbf{FT}} \\ \cmidrule{3-9} 
        & & \textbf{125M} & \textbf{350M} & \textbf{1.3B} & \textbf{2.7B} & \textbf{6.7B} & \textbf{13B} & \textbf{30B} 
        \\ \midrule
        \multirow{7}{*}{\rotatebox[origin=c]{90}{\textbf{ICL}}} & \textbf{125M}  & $\phantom{-}0.03$ & \cellcolor{blue!35}$\phantom{-}0.05$ & \cellcolor{blue!35}$\phantom{-}0.09$ & \cellcolor{blue!35}$\phantom{-}0.10$ & \cellcolor{blue!35}$\phantom{-}0.07$ & \cellcolor{blue!35}$\phantom{-}0.13$ & \cellcolor{blue!35}$\phantom{-}0.08$  \\
         & \textbf{350M}  & $\phantom{-}0.01$ & \cellcolor{blue!35}$\phantom{-}0.03$ & \cellcolor{blue!35}$\phantom{-}0.07$ & \cellcolor{blue!35}$\phantom{-}0.09$ & \cellcolor{blue!35}$\phantom{-}0.05$ & \cellcolor{blue!35}$\phantom{-}0.12$ & \cellcolor{blue!35}$\phantom{-}0.06$ \\
         & \textbf{1.3B} & \cellcolor{red!15}$-0.07$ & \cellcolor{red!15}$-0.04$ & $-0.01$ & $\phantom{-}0.01$ & $-0.02$ & $\phantom{-}0.04$ & $-0.01$  \\
          & \textbf{2.7B} & \cellcolor{red!15}$-0.03$ & $-0.01$ & $\phantom{-}0.02$ & $\phantom{-}0.04$ & $\phantom{-}0.01$ & \cellcolor{blue!35}$\phantom{-}0.07$ & $\phantom{-}0.02$  \\
         & \textbf{6.7B} & \cellcolor{red!15}$-0.10$ & \cellcolor{red!15}$-0.08$ & \cellcolor{red!15}$-0.04$ & $-0.03$ & \cellcolor{red!15}$-0.06$ & $\phantom{-}0.00$ & \cellcolor{red!15}$-0.05$  \\
         & \textbf{13B} & \cellcolor{red!15}$-0.17$ & \cellcolor{red!15}$-0.15$ & \cellcolor{red!15}$-0.11$ & \cellcolor{red!15}$-0.10$ & \cellcolor{red!15}$-0.13$ & \cellcolor{red!15}$-0.07$ & \cellcolor{red!15}$-0.12$  \\
         & \textbf{30B} & \cellcolor{red!15}$-0.16$ & \cellcolor{red!15}$-0.13$ & \cellcolor{red!15}$-0.10$ & \cellcolor{red!15}$-0.08$ & \cellcolor{red!15}$-0.11$ & $-0.05$ & \cellcolor{red!15}$-0.10$ \\
         
        \bottomrule
    \end{tabular}%
    }
        }
        \caption{MNLI}\label{tab:appendix-statistical-tests-ood-in-domain-mnli}
    \end{subtable}%
    \caption{Difference between average \textbf{in-domain performance} of ICL and FT on RTE (a) and MNLI (b) across model sizes. We use 16 examples and 10 random seeds for both approaches. For ICL, we use the \texttt{gpt-3} pattern. For FT, we use pattern-based fine-tuning (PBFT) and select checkpoints according to \underline{out-of-domain performance}. We perform a Welch's t-test and color cells according to whether: \textcolor{red!45}{ICL performs significantly better than FT}, \textcolor{blue!75}{FT performs significantly better than ICL}. For cells without color, there is no significant difference between ICL and FT.\looseness-1
    } \label{tab:appendix-statistical-tests-ood-in-domain}
\end{table*}
\begin{table*}
    \begin{subtable}{.49\textwidth}\centering
        {
            \resizebox{1\textwidth}{!}{%
            \begin{tabular}{llccccccc}
        \toprule
        & 
        & \multicolumn{7}{c}{\textbf{FT}} \\ \cmidrule{3-9} 
        & & \textbf{125M} & \textbf{350M} & \textbf{1.3B} & \textbf{2.7B} & \textbf{6.7B} & \textbf{13B} & \textbf{30B} 
        \\ \midrule
        \multirow{7}{*}{\rotatebox[origin=c]{90}{\textbf{ICL}}} & \textbf{125M}  & \cellcolor{blue!35}$\phantom{-}0.01$ & \cellcolor{blue!35}$\phantom{-}0.02$ & \cellcolor{blue!35}$\phantom{-}0.03$ & \cellcolor{blue!35}$\phantom{-}0.11$ & \cellcolor{blue!35}$\phantom{-}0.16$ & \cellcolor{blue!35}$\phantom{-}0.18$ & \cellcolor{blue!35}$\phantom{-}0.16$  \\
         & \textbf{350M}  & \cellcolor{blue!35}$\phantom{-}0.01$ & \cellcolor{blue!35}$\phantom{-}0.02$ & \cellcolor{blue!35}$\phantom{-}0.03$ & \cellcolor{blue!35}$\phantom{-}0.11$ & \cellcolor{blue!35}$\phantom{-}0.16$ & \cellcolor{blue!35}$\phantom{-}0.18$ & \cellcolor{blue!35}$\phantom{-}0.16$ \\
         & \textbf{1.3B} & $\phantom{-}0.01$ & \cellcolor{blue!35}$\phantom{-}0.02$ & \cellcolor{blue!35}$\phantom{-}0.03$ & \cellcolor{blue!35}$\phantom{-}0.11$ & \cellcolor{blue!35}$\phantom{-}0.16$ & \cellcolor{blue!35}$\phantom{-}0.18$ & \cellcolor{blue!35}$\phantom{-}0.16$  \\
          & \textbf{2.7B} & $\phantom{-}0.00$ & \cellcolor{blue!35}$\phantom{-}0.02$ & \cellcolor{blue!35}$\phantom{-}0.03$ & \cellcolor{blue!35}$\phantom{-}0.11$ & \cellcolor{blue!35}$\phantom{-}0.15$ & \cellcolor{blue!35}$\phantom{-}0.18$ & \cellcolor{blue!35}$\phantom{-}0.16$  \\
         & \textbf{6.7B} & \cellcolor{blue!35}$\phantom{-}0.01$ & \cellcolor{blue!35}$\phantom{-}0.02$ & \cellcolor{blue!35}$\phantom{-}0.03$ & \cellcolor{blue!35}$\phantom{-}0.11$ & \cellcolor{blue!35}$\phantom{-}0.15$ & \cellcolor{blue!35}$\phantom{-}0.18$ & \cellcolor{blue!35}$\phantom{-}0.16$  \\
         & \textbf{13B} & \cellcolor{red!15}$-0.03$ & $-0.01$ & $\phantom{-}0.00$ & \cellcolor{blue!35}$\phantom{-}0.08$ & \cellcolor{blue!35}$\phantom{-}0.12$ & \cellcolor{blue!35}$\phantom{-}0.14$ & \cellcolor{blue!35}$\phantom{-}0.13$ \\
         & \textbf{30B} & \cellcolor{red!15}$-0.10$ & \cellcolor{red!15}$-0.08$ & \cellcolor{red!15}$-0.07$ & $\phantom{-}0.01$ & $\phantom{-}0.05$ & \cellcolor{blue!35}$\phantom{-}0.07$ & $\phantom{-}0.06$  \\
         
        \bottomrule
    \end{tabular}%
    }
        }
        \caption{RTE}\label{tab:appendix-statistical-tests-ood-ood-rte}
    \end{subtable}%
    ~~ 
    \begin{subtable}{.49\textwidth}\centering
        {
            \resizebox{1\textwidth}{!}{%
            \begin{tabular}{llccccccc}
        \toprule
        & 
        & \multicolumn{7}{c}{\textbf{FT}} \\ \cmidrule{3-9} 
        & & \textbf{125M} & \textbf{350M} & \textbf{1.3B} & \textbf{2.7B} & \textbf{6.7B} & \textbf{13B} & \textbf{30B} 
        \\ \midrule
        \multirow{7}{*}{\rotatebox[origin=c]{90}{\textbf{ICL}}} & \textbf{125M}  & \cellcolor{blue!35}$\phantom{-}0.00$ & \cellcolor{blue!35}$\phantom{-}0.01$ & \cellcolor{blue!35}$\phantom{-}0.05$ & \cellcolor{blue!35}$\phantom{-}0.04$ & \cellcolor{blue!35}$\phantom{-}0.13$ & \cellcolor{blue!35}$\phantom{-}0.14$ & \cellcolor{blue!35}$\phantom{-}0.17$  \\
         & \textbf{350M}  & \cellcolor{blue!35}$\phantom{-}0.00$ & \cellcolor{blue!35}$\phantom{-}0.01$ & \cellcolor{blue!35}$\phantom{-}0.05$ & \cellcolor{blue!35}$\phantom{-}0.04$ & \cellcolor{blue!35}$\phantom{-}0.13$ & \cellcolor{blue!35}$\phantom{-}0.14$ & \cellcolor{blue!35}$\phantom{-}0.17$ \\
         & \textbf{1.3B} & $\phantom{-}0.00$ & $\phantom{-}0.01$ & \cellcolor{blue!35}$\phantom{-}0.05$ & \cellcolor{blue!35}$\phantom{-}0.04$ & \cellcolor{blue!35}$\phantom{-}0.13$ & \cellcolor{blue!35}$\phantom{-}0.14$ & \cellcolor{blue!35}$\phantom{-}0.16$ \\
          & \textbf{2.7B} & $\phantom{-}0.00$ & $\phantom{-}0.01$ & \cellcolor{blue!35}$\phantom{-}0.05$ & \cellcolor{blue!35}$\phantom{-}0.04$ & \cellcolor{blue!35}$\phantom{-}0.13$ & \cellcolor{blue!35}$\phantom{-}0.14$ & \cellcolor{blue!35}$\phantom{-}0.16$  \\
         & \textbf{6.7B} & $-0.01$ & $-0.00$ & \cellcolor{blue!35}$\phantom{-}0.04$ & \cellcolor{blue!35}$\phantom{-}0.03$ & \cellcolor{blue!35}$\phantom{-}0.12$ & \cellcolor{blue!35}$\phantom{-}0.13$ & \cellcolor{blue!35}$\phantom{-}0.16$  \\
         & \textbf{13B} & \cellcolor{red!15}$-0.03$ & \cellcolor{red!15}$-0.02$ & $\phantom{-}0.02$ & $\phantom{-}0.01$ & \cellcolor{blue!35}$\phantom{-}0.10$ & \cellcolor{blue!35}$\phantom{-}0.11$ & \cellcolor{blue!35}$\phantom{-}0.13$  \\
         & \textbf{30B} & \cellcolor{red!15}$-0.06$ & \cellcolor{red!15}$-0.06$ & $-0.01$ & $-0.02$ & $\phantom{-}0.06$ & \cellcolor{blue!35}$\phantom{-}0.08$ & \cellcolor{blue!35}$\phantom{-}0.10$ \\
         
        \bottomrule
    \end{tabular}%
    }
        }
        \caption{MNLI}\label{tab:appendix-statistical-tests-ood-ood-mnli}
    \end{subtable}%
    \caption{Difference between average \textbf{out-of-domain performance} of ICL and FT on RTE (a) and MNLI (b) across model sizes. We use 16 examples and 10 random seeds for both approaches. For ICL, we use the \texttt{gpt-3} pattern. For FT, we use pattern-based fine-tuning (PBFT) and select checkpoints according to \underline{out-of-domain performance}. We perform a Welch's t-test and color cells according to whether: \textcolor{red!45}{ICL performs significantly better than FT}, \textcolor{blue!75}{FT performs significantly better than ICL}. For cells without color, there is no significant difference between ICL and FT.\looseness-1
    } \label{tab:appendix-statistical-tests-ood-ood}
\end{table*}

\subsection{In-context learning}
\label{appendix:additional-results-in-context}

\begin{figure*}[h]
    \centering
    \begin{subfigure}[b]{0.31\textwidth}
        \centering
        \includegraphics[width=\textwidth]{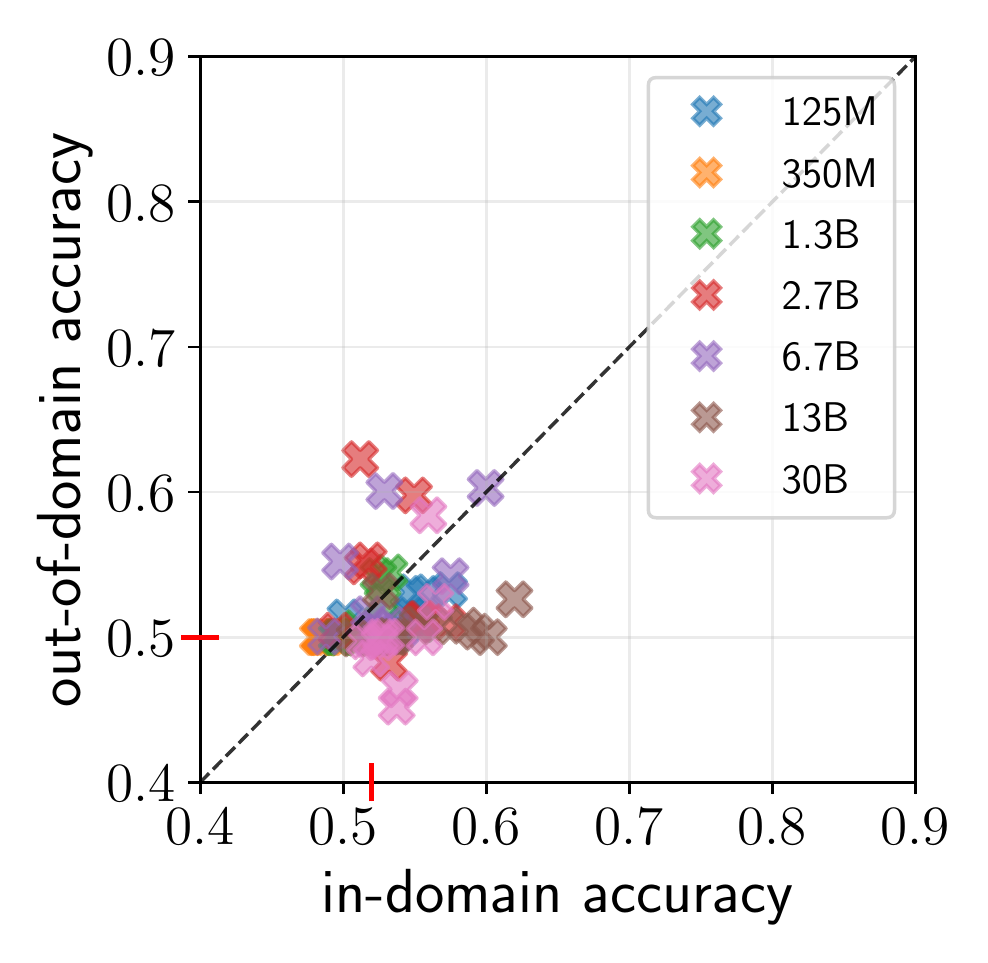}
        \caption{2 samples -- \texttt{minimal}}
    \end{subfigure}
    ~
    \begin{subfigure}[b]{0.31\textwidth}
        \centering
        \includegraphics[width=\textwidth]{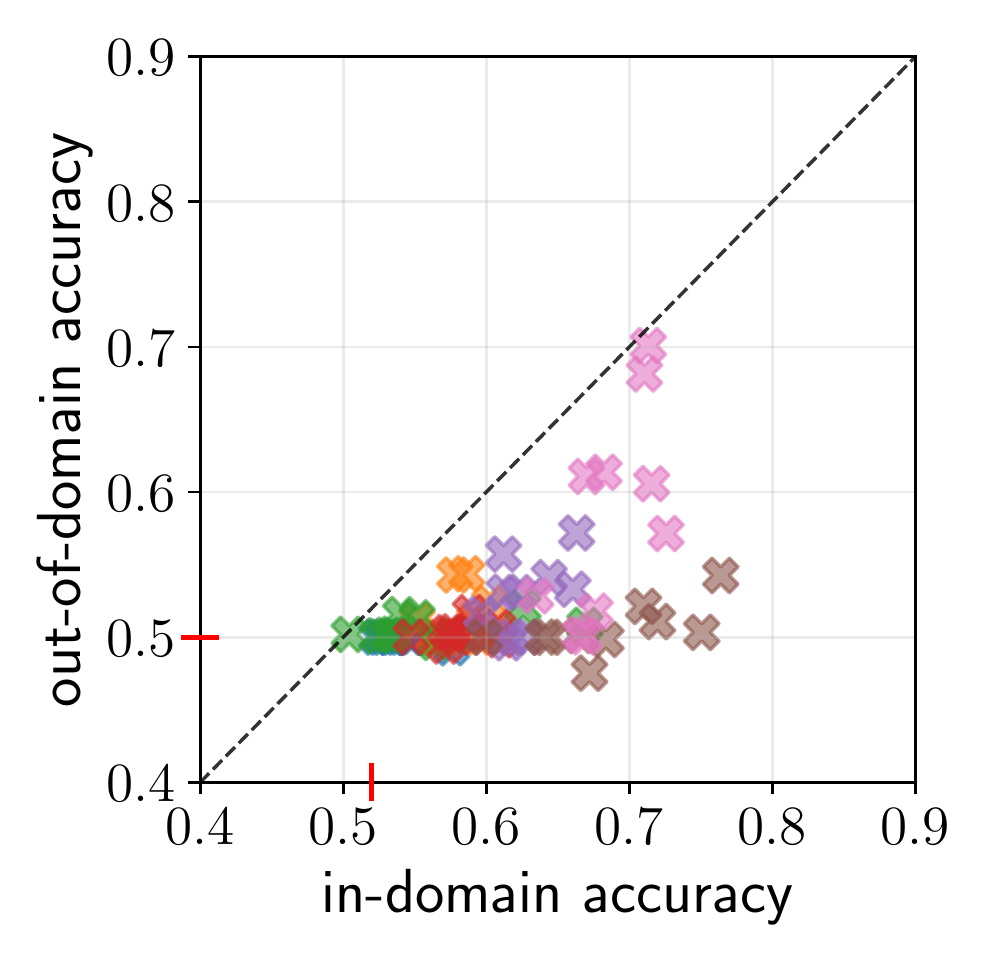}
        \caption{2 samples -- \texttt{gpt-3}}
    \end{subfigure}
    ~
    \begin{subfigure}[b]{0.31\textwidth}
        \centering
        \includegraphics[width=\textwidth]{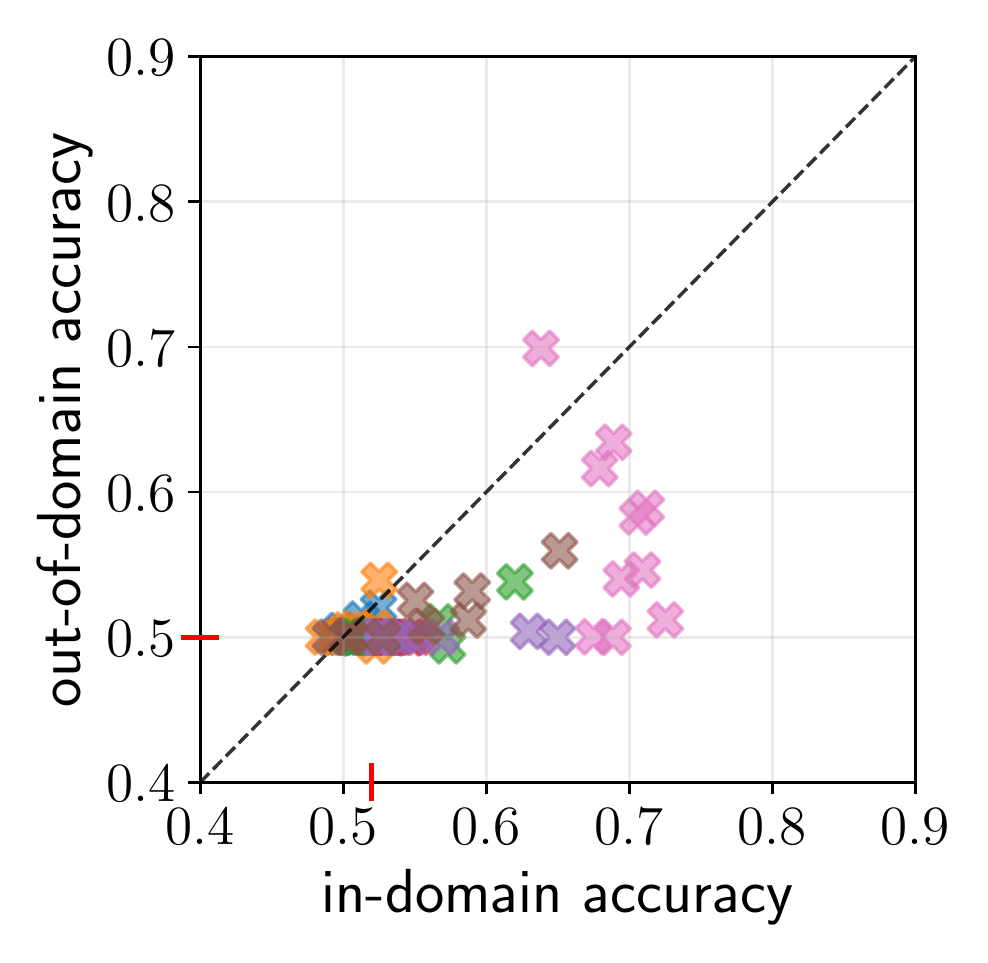}
        \caption{2 samples -- \texttt{eval-harness}}
    \end{subfigure}
    \\
    \begin{subfigure}[b]{0.31\textwidth}
        \centering
        \includegraphics[width=\textwidth]{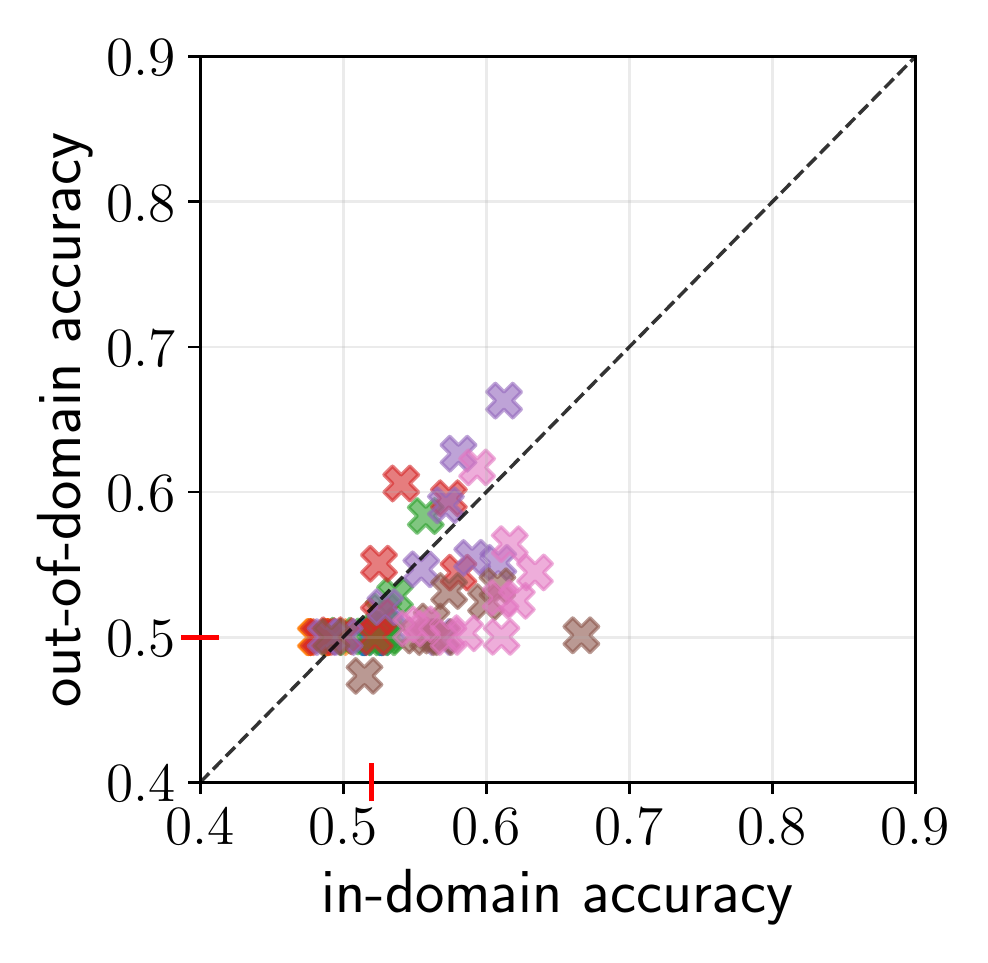}
        \caption{16 samples -- \texttt{minimal}}
    \end{subfigure}
    ~
    \begin{subfigure}[b]{0.31\textwidth}
        \centering
        \includegraphics[width=\textwidth]{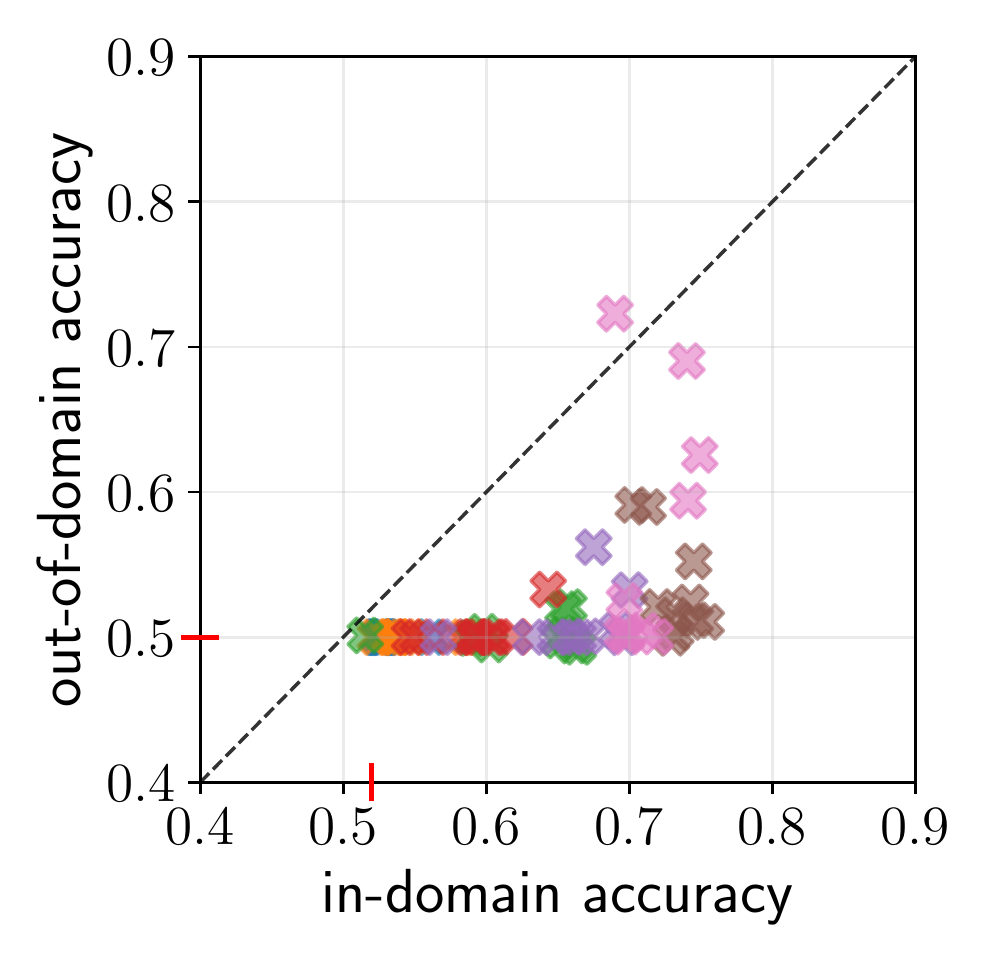}
        \caption{16 samples -- \texttt{gpt-3}}
    \end{subfigure}
    ~
    \begin{subfigure}[b]{0.31\textwidth}
        \centering
        \includegraphics[width=\textwidth]{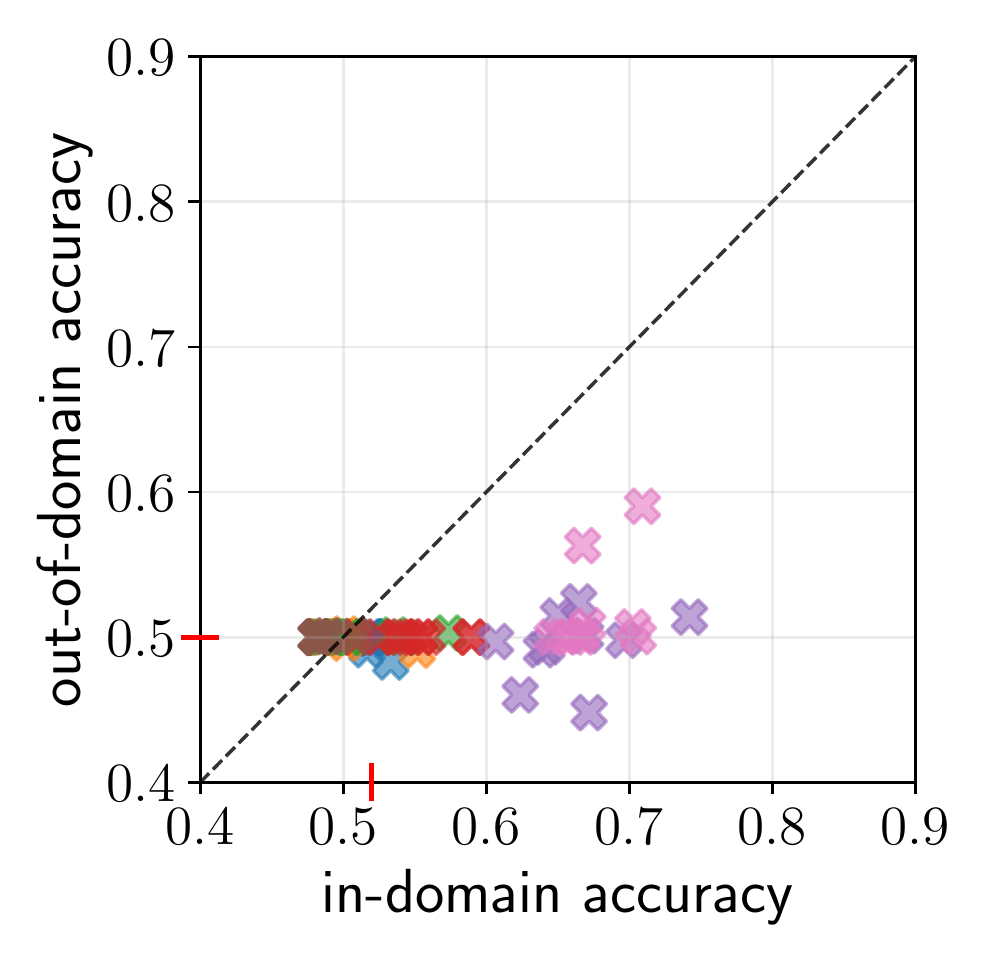}
        \caption{16 samples -- \texttt{eval-harness}}
    \end{subfigure}
    \\
    \begin{subfigure}[b]{0.31\textwidth}
        \centering
        \includegraphics[width=\textwidth]{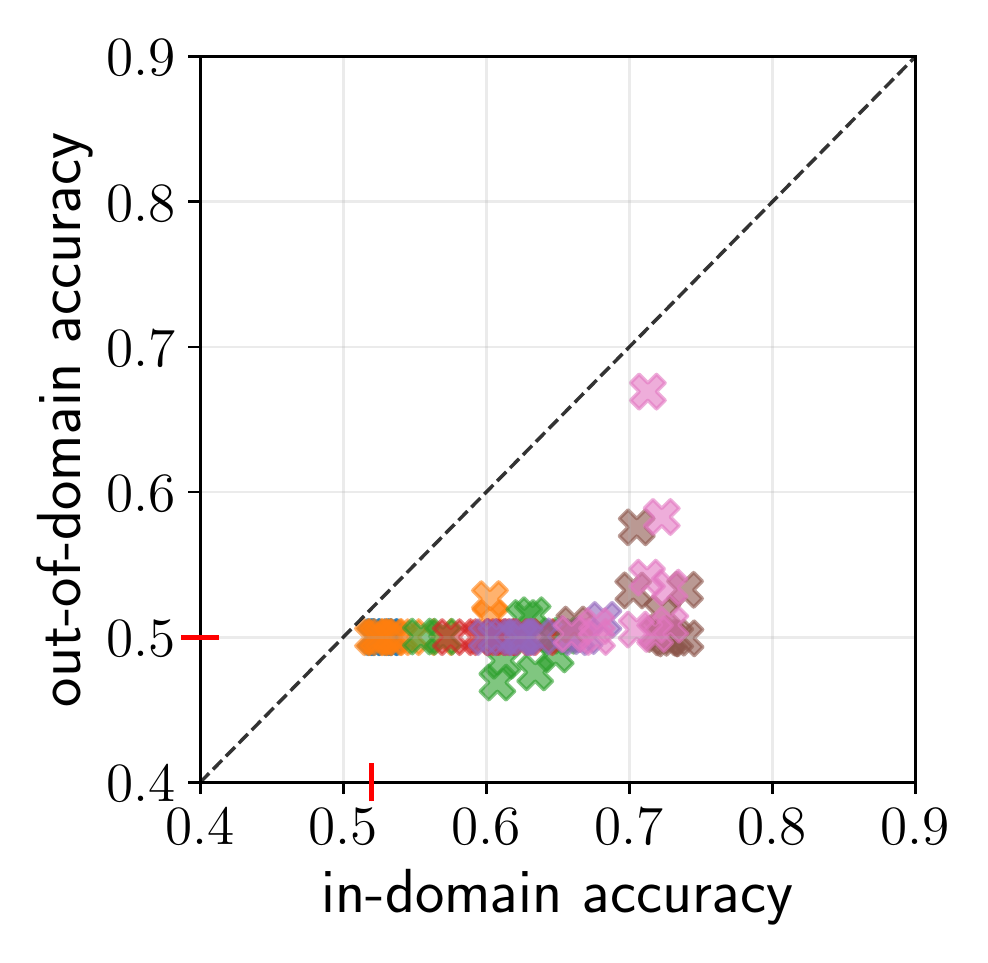}
        \caption{32 samples -- \texttt{minimal}}
    \end{subfigure}
    ~
    \begin{subfigure}[b]{0.31\textwidth}
        \centering
        \includegraphics[width=\textwidth]{figures/in-context/in-context_all-models_32-shots_mnli_gpt-3.pdf}
        \caption{32 samples -- \texttt{gpt-3}}
    \end{subfigure}
    ~
    \begin{subfigure}[b]{0.31\textwidth}
        \centering
        \includegraphics[width=\textwidth]{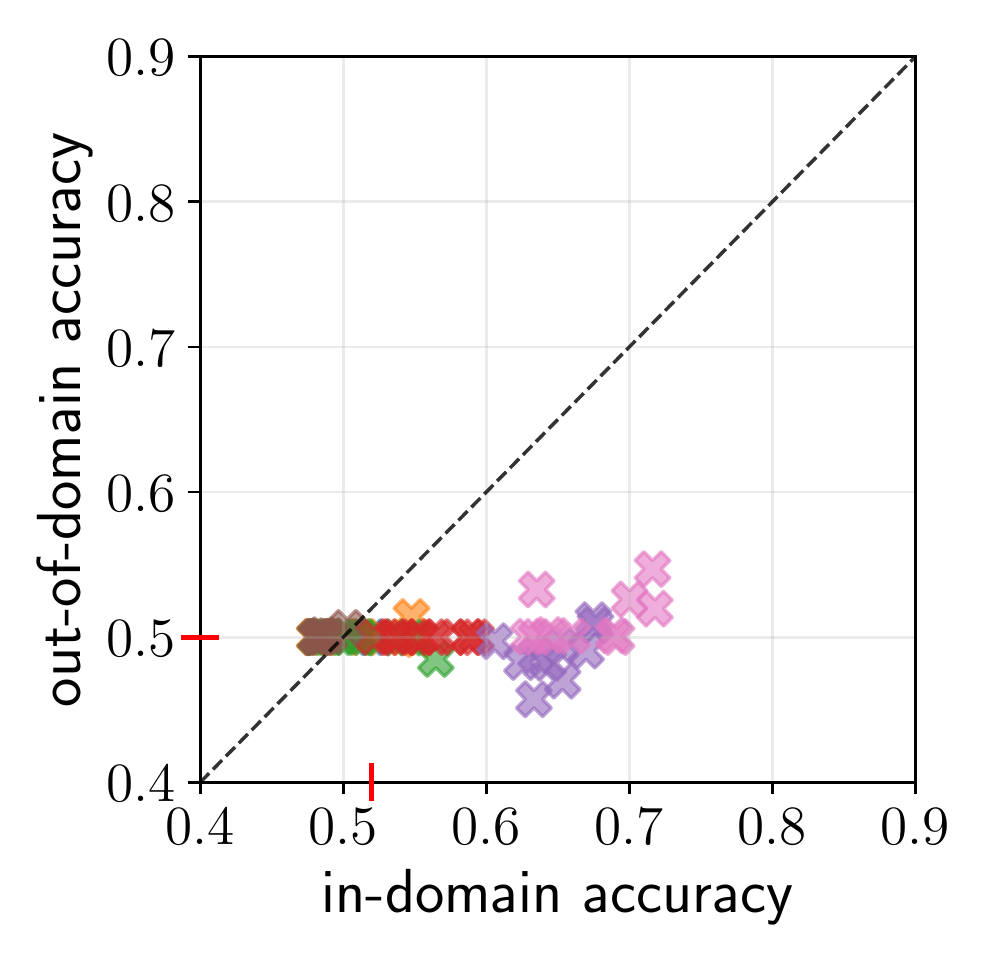}
        \caption{32 samples -- \texttt{eval-harness}}
    \end{subfigure}
    
    \caption{\textbf{Relationship between in-domain and out-of-domain performance of ICL on MNLI} for OPT models of various sizes. Rows vary amount of training data. Columns vary input pattern.
    Colors indicate model size. We run 10 models per setting varying only the data seed.
    \textcolor{red}{$\boldsymbol{-}$} in the x- and y-axis indicates the performance of the majority class label.
    }
    \label{fig:in-context-model-selection-mnli}
\end{figure*}

\begin{figure*}[h]
    \centering
    \begin{subfigure}[b]{0.31\textwidth}
        \centering
        \includegraphics[width=\textwidth]{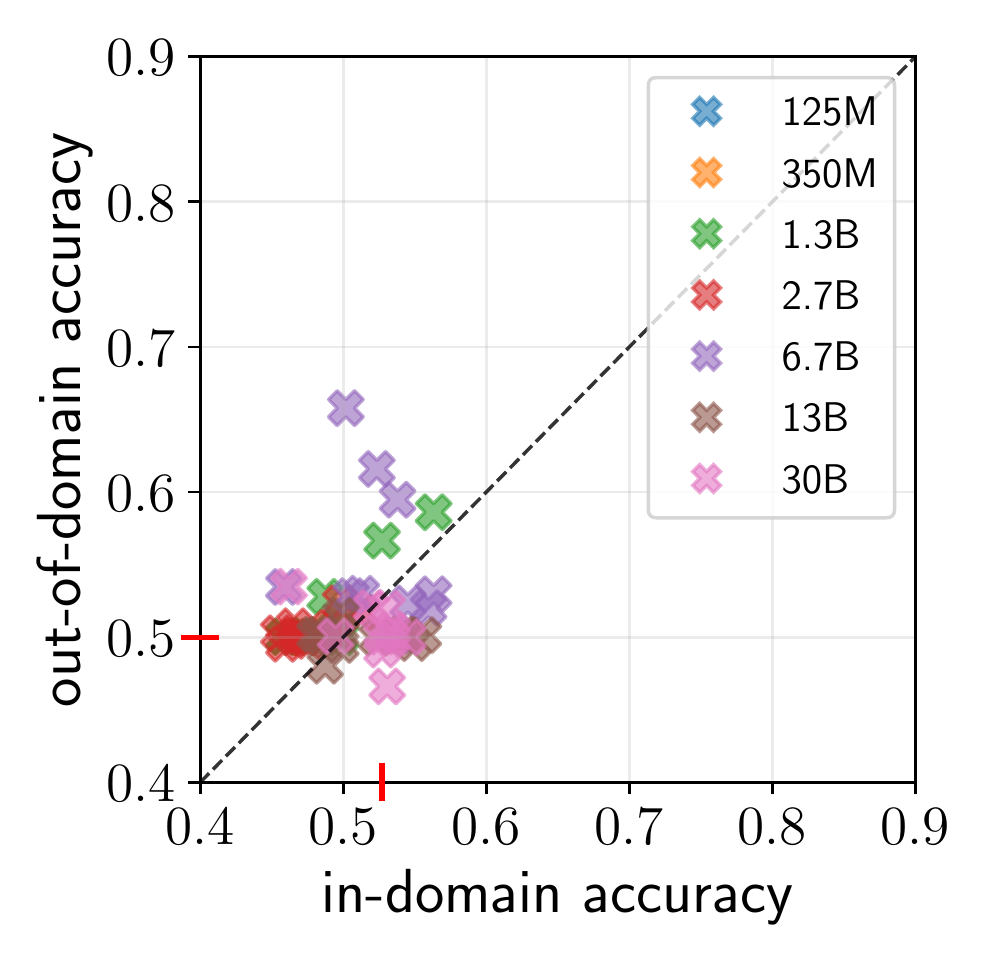}
        \caption{2 samples -- \texttt{minimal}}
    \end{subfigure}
    ~
    \begin{subfigure}[b]{0.31\textwidth}
        \centering
        \includegraphics[width=\textwidth]{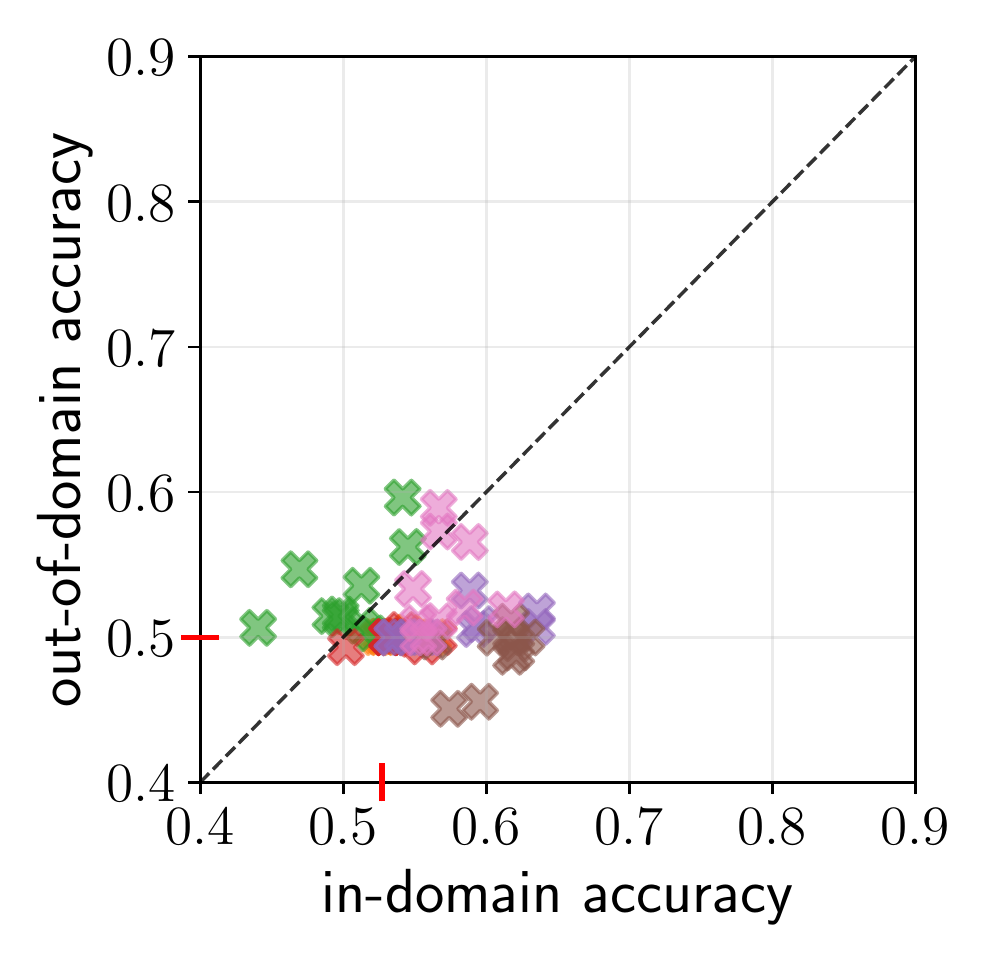}
        \caption{2 samples -- \texttt{gpt-3}}
    \end{subfigure}
    ~
    \begin{subfigure}[b]{0.31\textwidth}
        \centering
        \includegraphics[width=\textwidth]{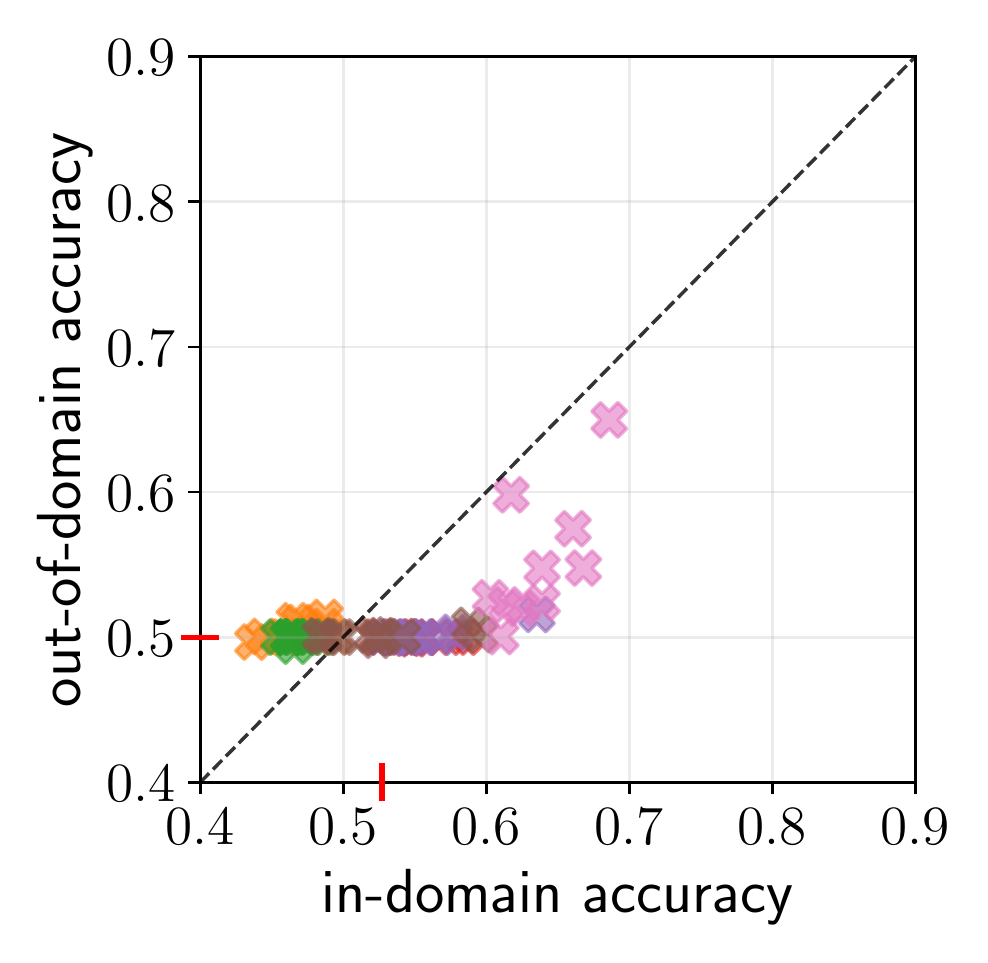}
        \caption{2 samples -- \texttt{eval-harness}}
    \end{subfigure}
    \\
    \begin{subfigure}[b]{0.31\textwidth}
        \centering
        \includegraphics[width=\textwidth]{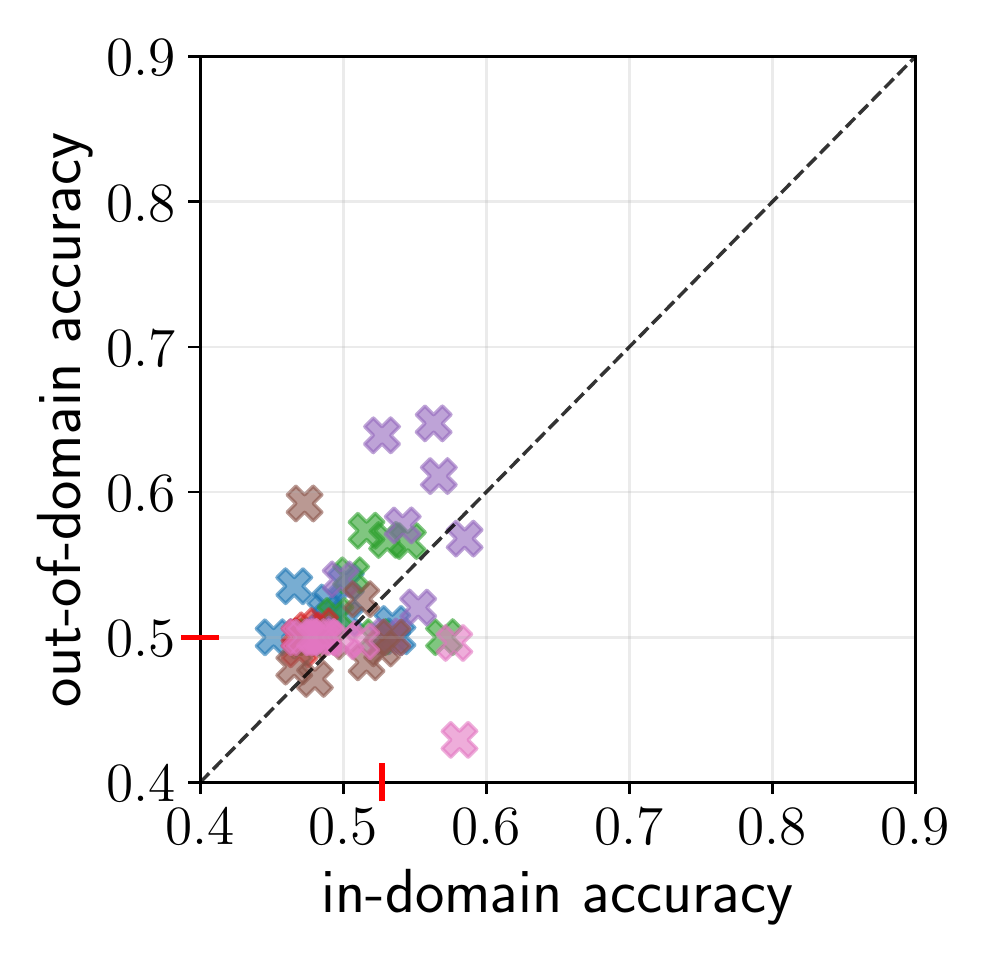}
        \caption{16 samples -- \texttt{minimal}}
    \end{subfigure}
    ~
    \begin{subfigure}[b]{0.31\textwidth}
        \centering
        \includegraphics[width=\textwidth]{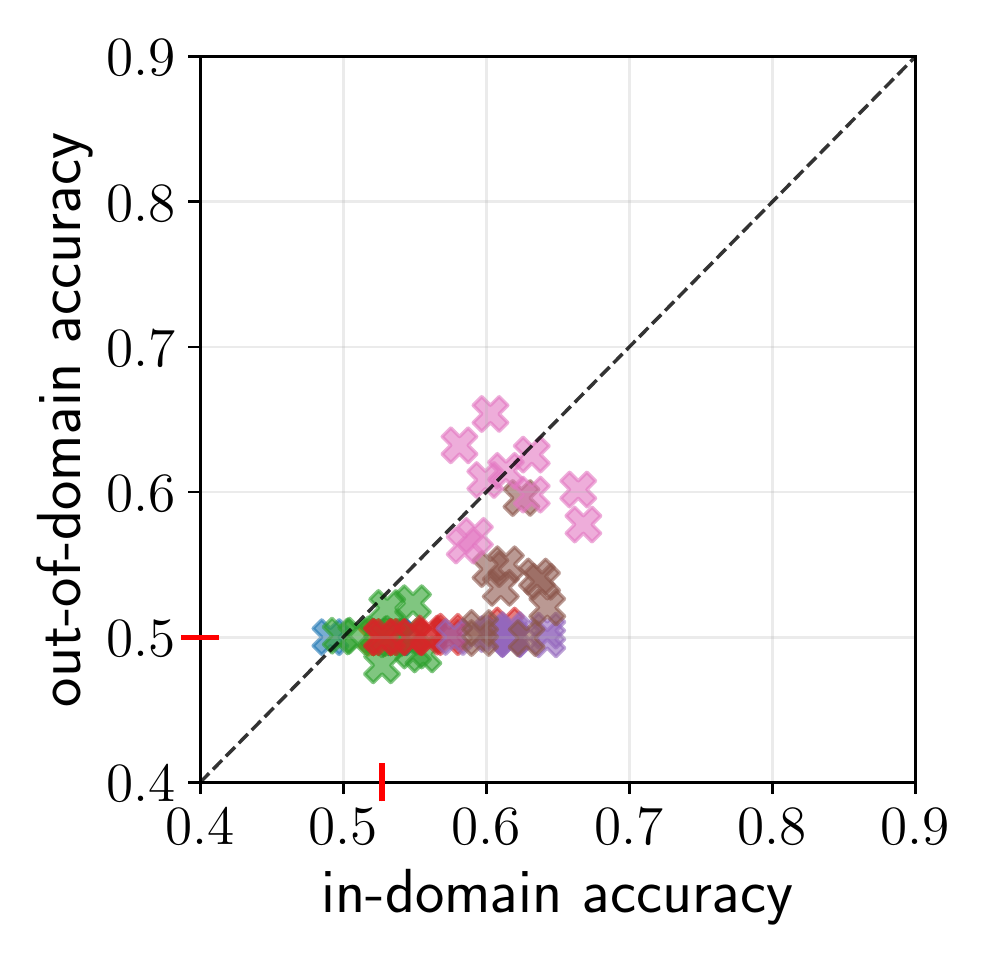}
        \caption{16 samples -- \texttt{gpt-3}}
    \end{subfigure}
    ~
    \begin{subfigure}[b]{0.31\textwidth}
        \centering
        \includegraphics[width=\textwidth]{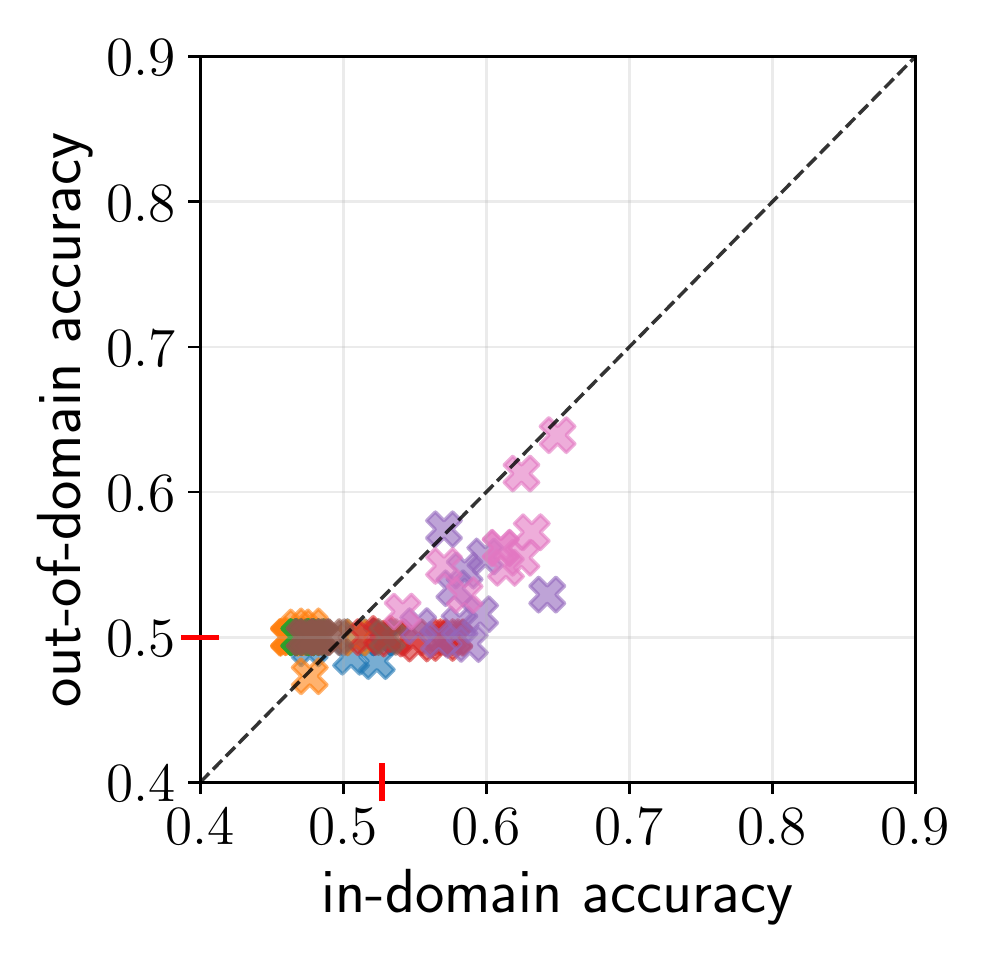}
        \caption{16 samples -- \texttt{eval-harness}}
    \end{subfigure}
    
    \caption{\textbf{Relationship between in-domain and out-of-domain performance of ICL on RTE} for OPT models of various sizes. Rows vary amount of training data. Columns vary input pattern.
    Colors indicate model size. We run 10 models per setting varying only the data seed.
    \textcolor{red}{$\boldsymbol{-}$} in the x- and y-axis indicates the performance of the majority class label.
    }
    \label{fig:in-context-model-selection-rte}
\end{figure*}

\begin{figure*}[h]
    \centering
    \begin{subfigure}[b]{0.31\textwidth}
        \centering
        \includegraphics[width=\textwidth]{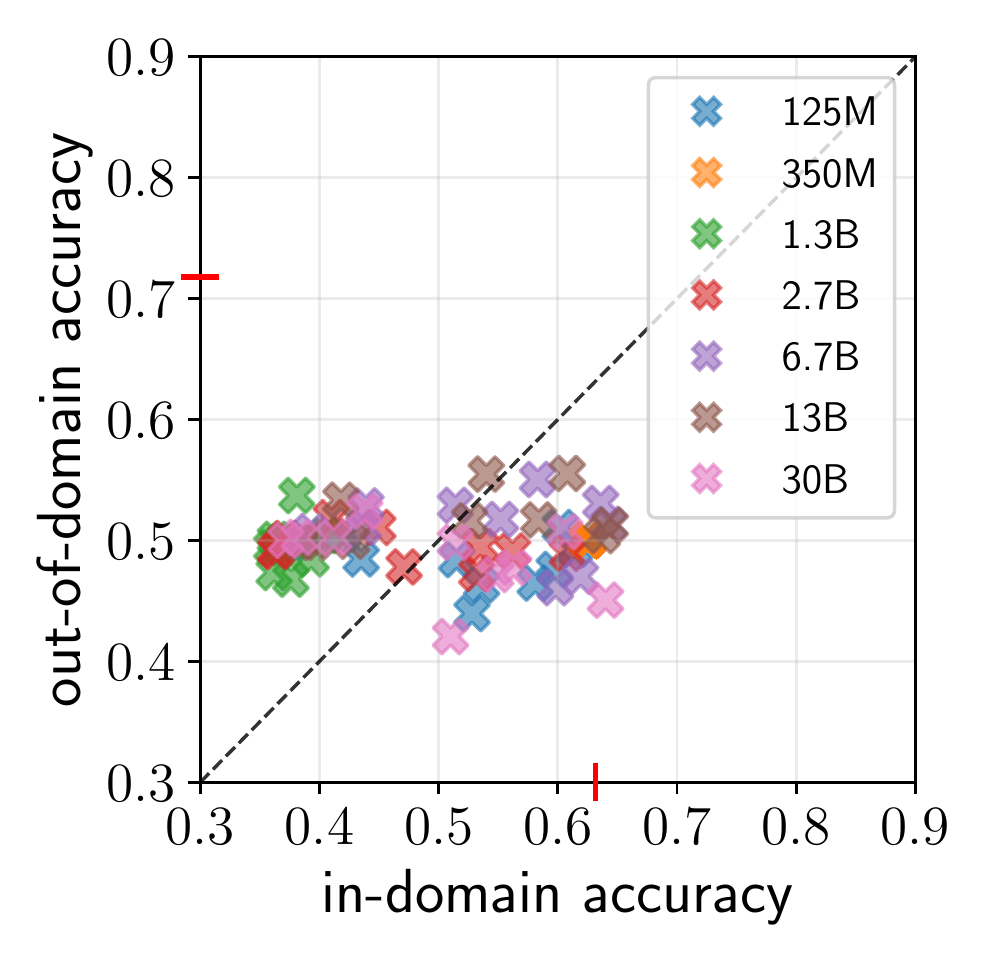}
        \caption{2 samples -- \texttt{minimal}}
    \end{subfigure}
    ~
    \begin{subfigure}[b]{0.31\textwidth}
        \centering
        \includegraphics[width=\textwidth]{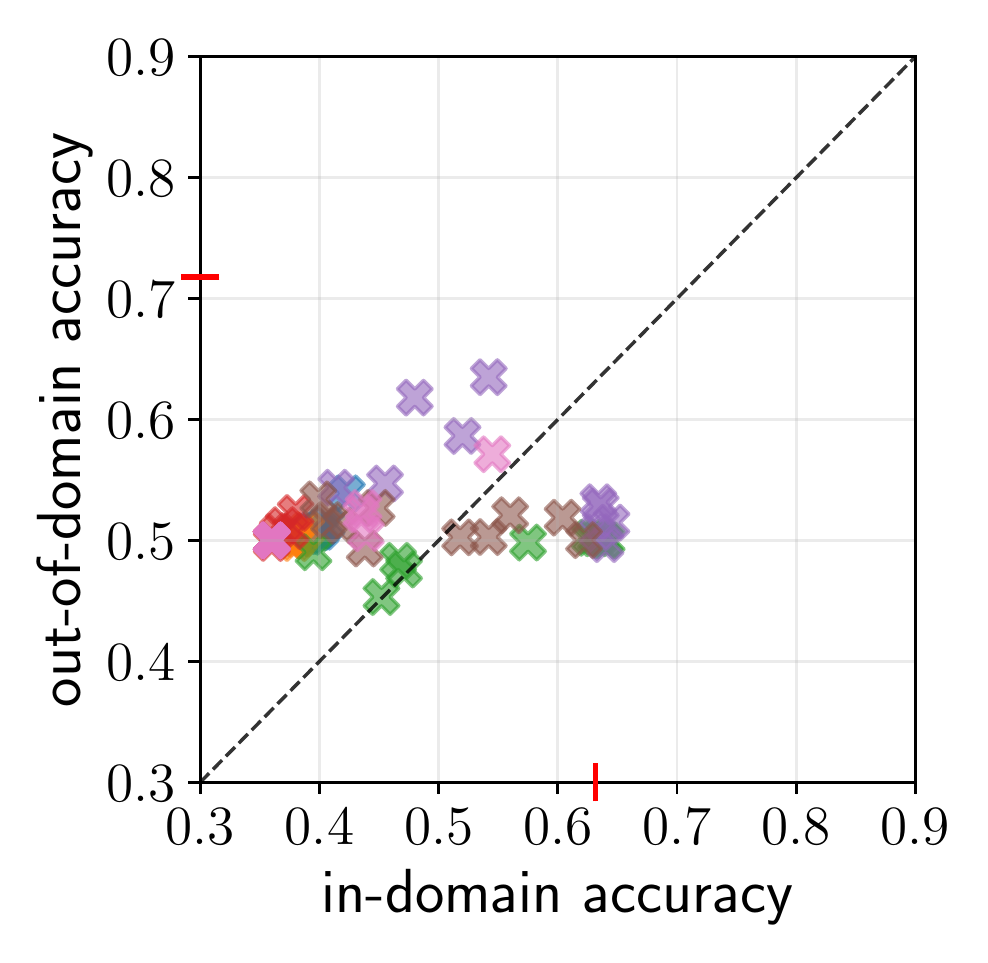}
        \caption{2 samples -- \texttt{eval-harness}}
    \end{subfigure}
    \\
    \begin{subfigure}[b]{0.31\textwidth}
        \centering
        \includegraphics[width=\textwidth]{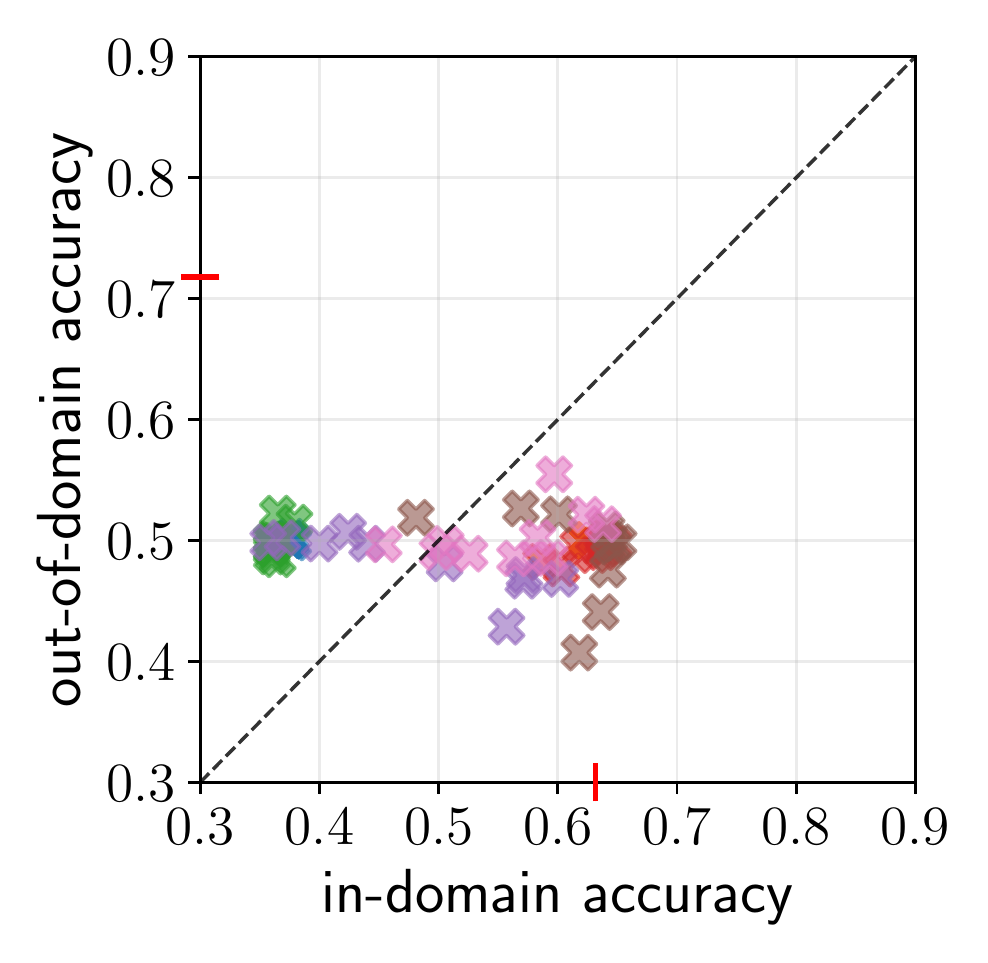}
        \caption{16 samples -- \texttt{minimal}}
    \end{subfigure}
    ~
    \begin{subfigure}[b]{0.31\textwidth}
        \centering
        \includegraphics[width=\textwidth]{figures/in-context/in-context_all-models_16-shots_qqp_eval-harness.pdf}
        \caption{16 samples -- \texttt{eval-harness}}
    \end{subfigure}
    \\
    \begin{subfigure}[b]{0.31\textwidth}
        \centering
        \includegraphics[width=\textwidth]{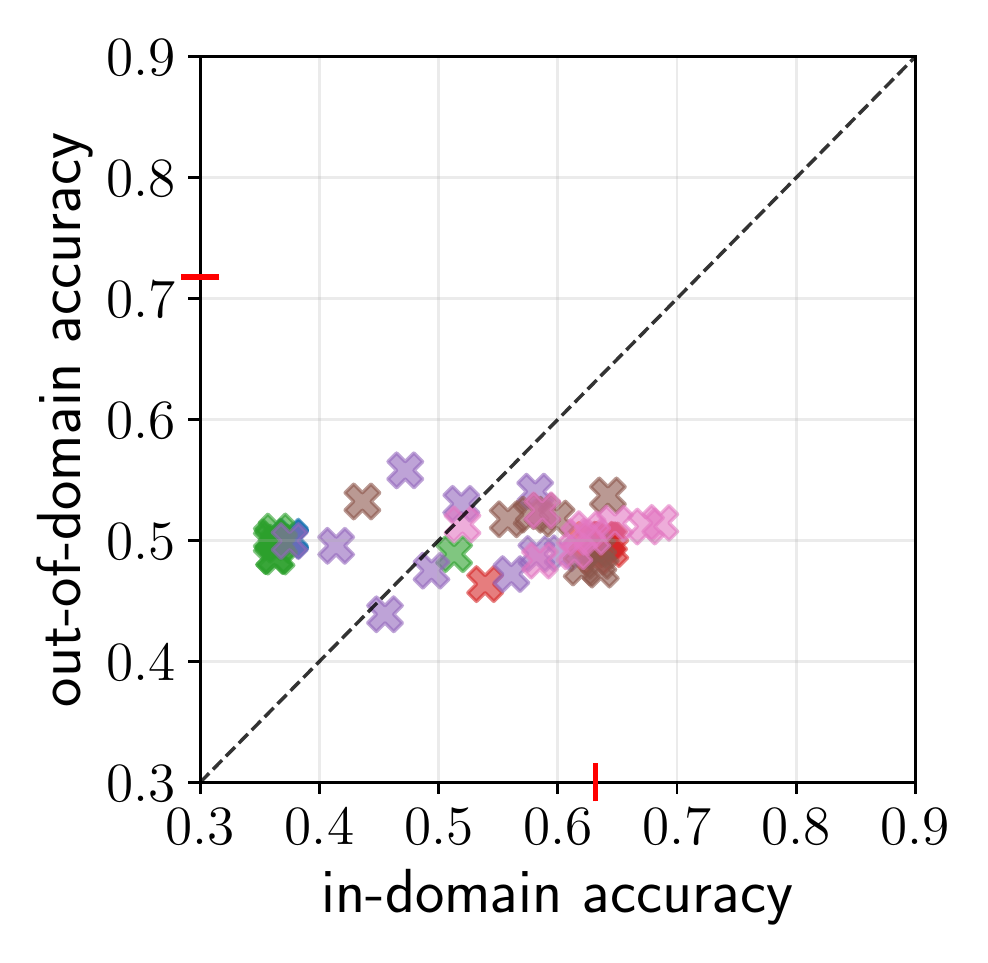}
        \caption{32 samples -- \texttt{minimal}}
    \end{subfigure}
    ~
    \begin{subfigure}[b]{0.31\textwidth}
        \centering
        \includegraphics[width=\textwidth]{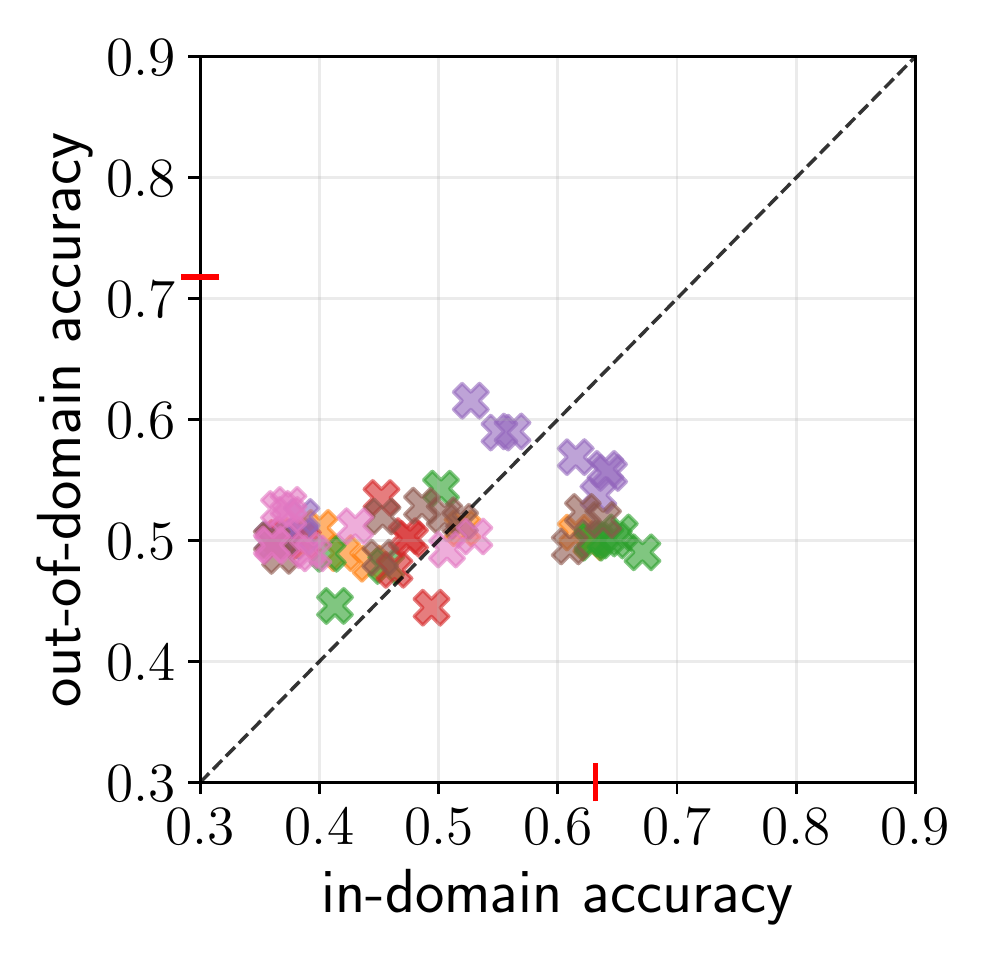}
        \caption{32 samples -- \texttt{eval-harness}}
    \end{subfigure}
    
    \caption{\textbf{Relationship between in-domain and out-of-domain performance of ICL on QQP} for OPT models of various sizes. Rows vary amount of training data. Columns vary input pattern.
    Colors indicate model size. We run 10 models per setting varying only the data seed.
    \textcolor{red}{$\boldsymbol{-}$} in the x- and y-axis indicates the performance of the majority class label.
    }
    \label{fig:in-context-model-selection-qqp}
\end{figure*}

Figures \ref{fig:in-context-model-selection-mnli}, \ref{fig:in-context-model-selection-rte}, and \ref{fig:in-context-model-selection-qqp} show ICL results on MNLI, RTE, and QQP for all OPT model sizes grouped by number of demonstrations and patterns. 

\paragraph{Sensitivity to pattern choice and number of examples}

On MNLI and RTE, we find that only the largest model benefits from the instructive \texttt{gpt-3} and \texttt{eval-harness} patterns. Moreover, on all datasets and for all patterns, models are sensitive to the number of demonstrations and do not necessarily improve with more demonstrations.


\subsection{Fine-tuning}
\label{appendix:additional-results-ft}

\begin{figure*}[h]
    \centering
    \begin{subfigure}[b]{0.31\textwidth}
        \centering
        \includegraphics[width=\textwidth]{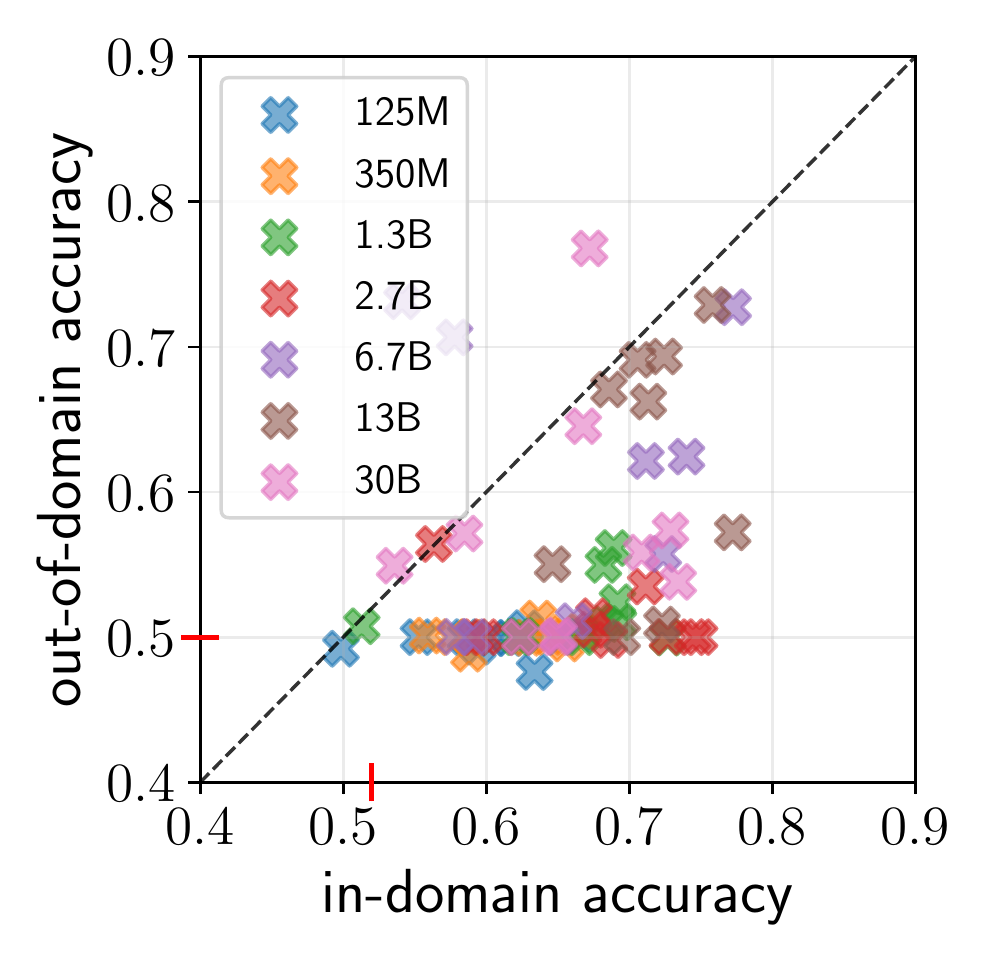}
        \caption{16 -- in-domain}
    \end{subfigure}
    ~
    \begin{subfigure}[b]{0.31\textwidth}
        \centering
        \includegraphics[width=\textwidth]{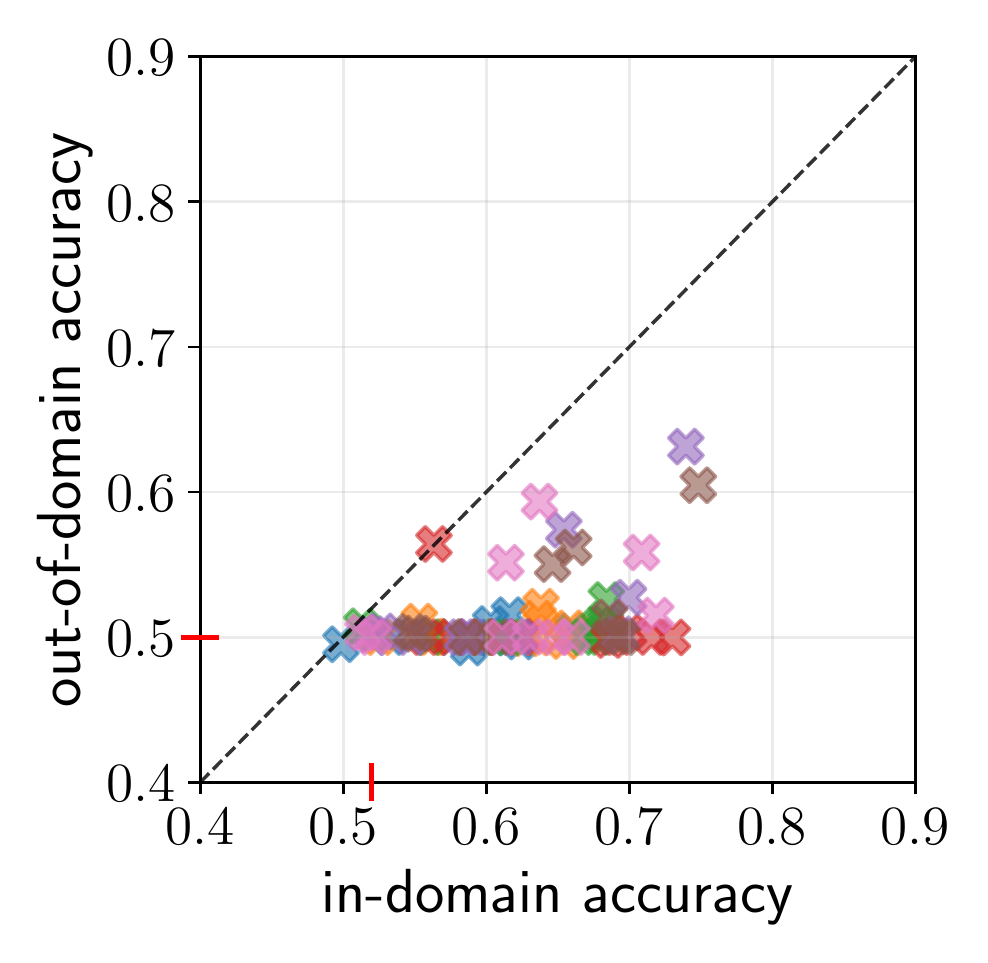}
        \caption{16 -- last checkpoint}
    \end{subfigure}
    ~
    \begin{subfigure}[b]{0.31\textwidth}
        \centering
        \includegraphics[width=\textwidth]{figures/ft/all-models_mnli_16_best_out-of-domain_pattern-verbalizer-ft.pdf}
        \caption{16 -- out-of-domain}
    \end{subfigure}
    \\
    \begin{subfigure}[b]{0.31\textwidth}
        \centering
        \includegraphics[width=\textwidth]{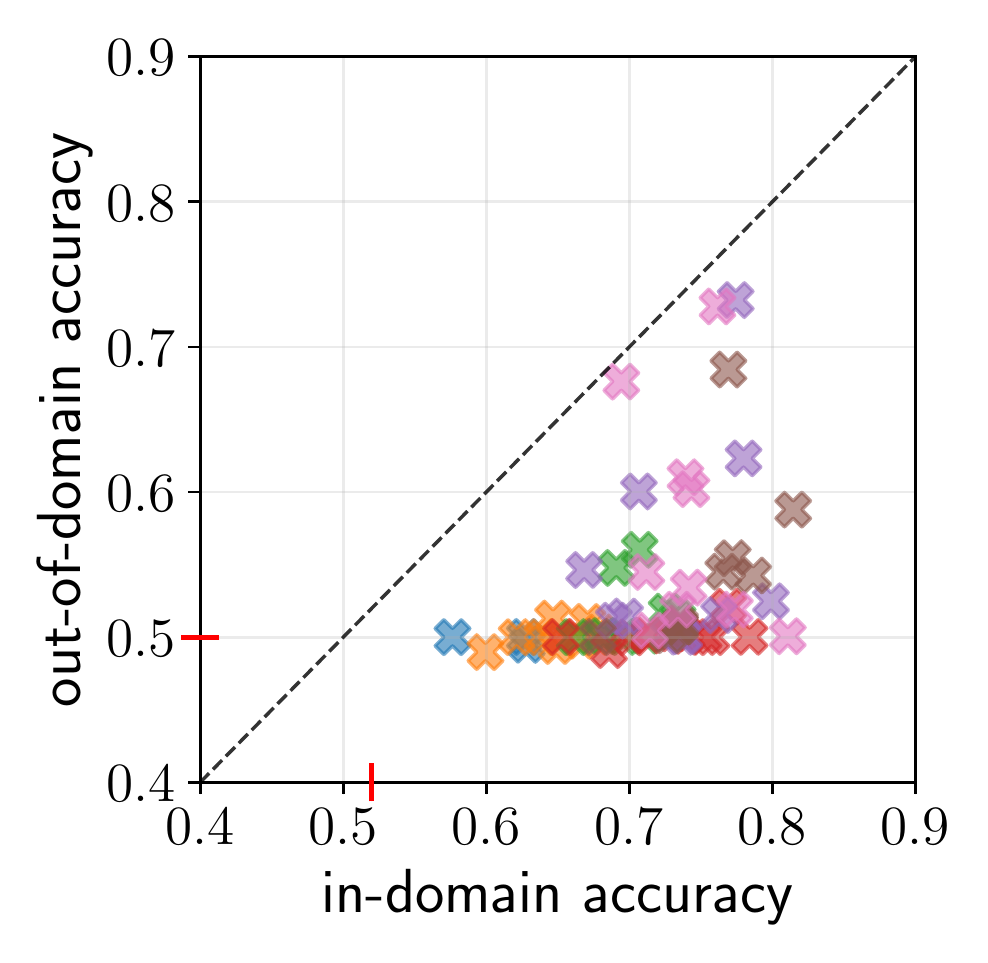}
        \caption{32 -- in-domain}
    \end{subfigure}
    ~
    \begin{subfigure}[b]{0.31\textwidth}
        \centering
        \includegraphics[width=\textwidth]{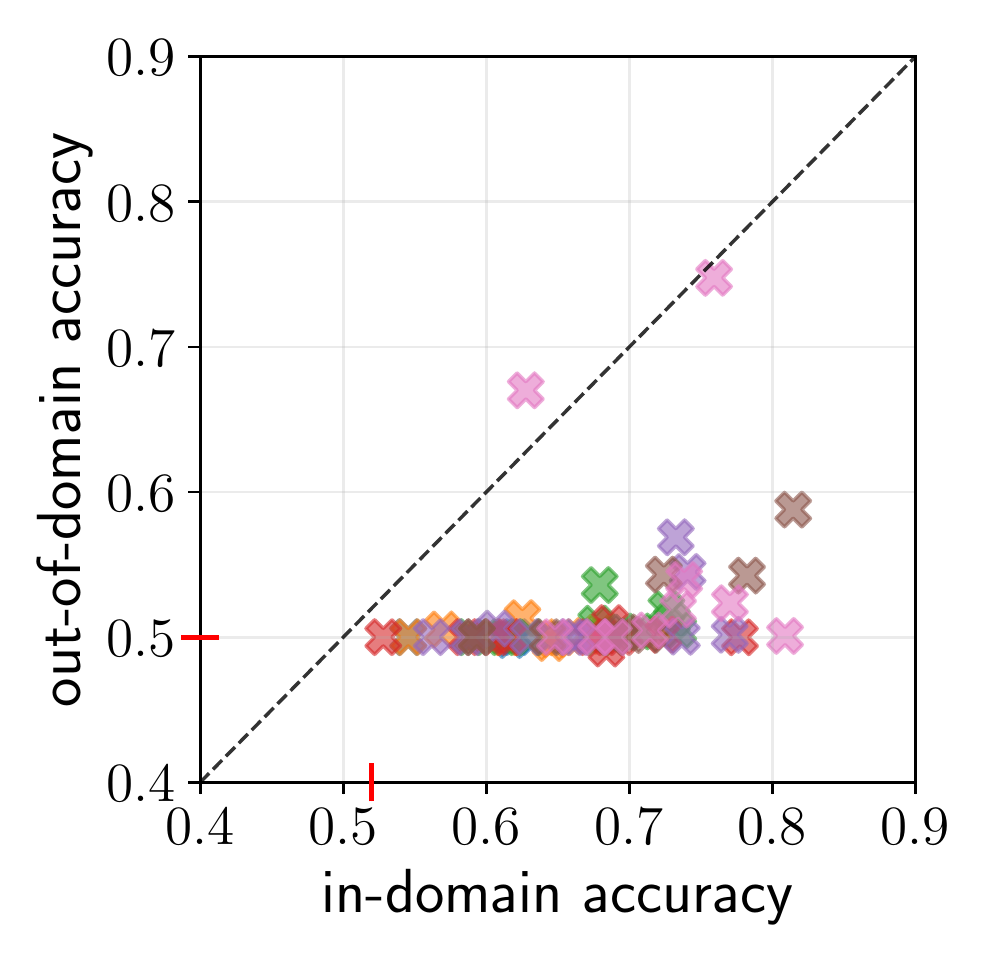}
        \caption{32 -- last checkpoint}
    \end{subfigure}
    ~
    \begin{subfigure}[b]{0.31\textwidth}
        \centering
        \includegraphics[width=\textwidth]{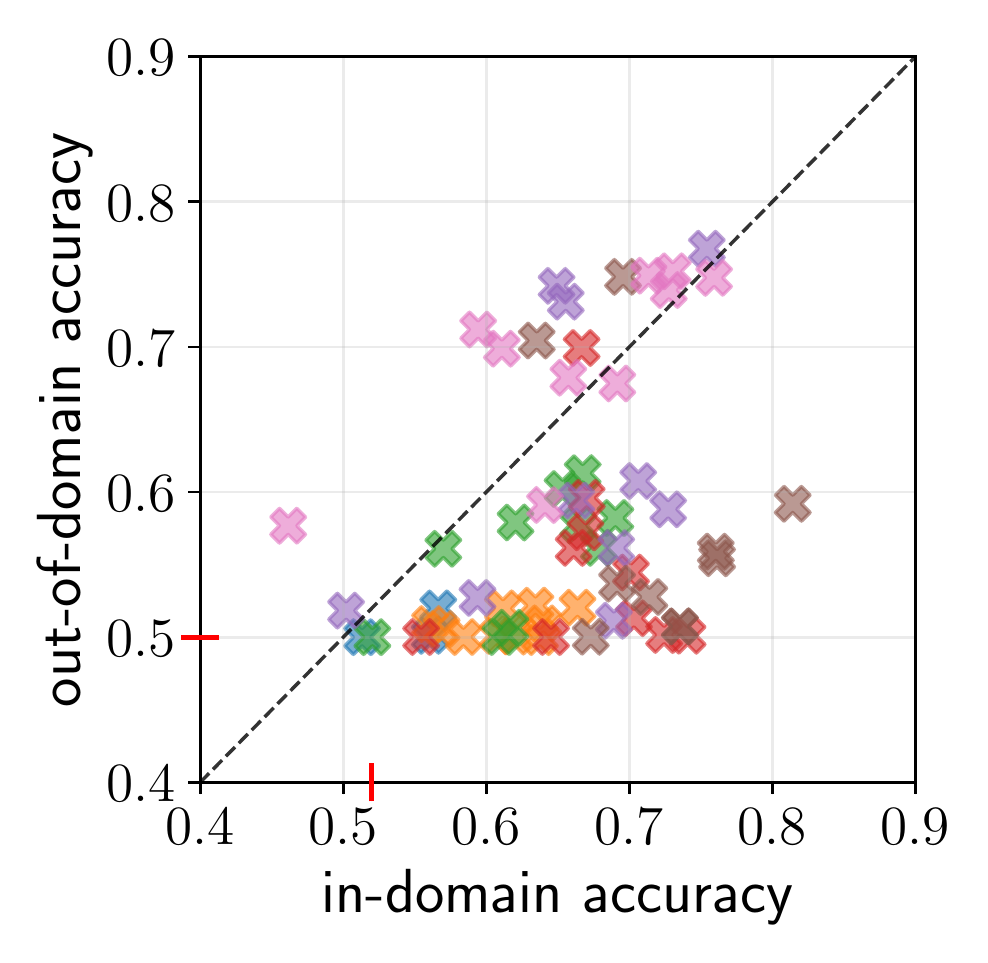}
        \caption{32 -- out-of-domain}
    \end{subfigure}
    \\
    \begin{subfigure}[b]{0.31\textwidth}
        \centering
        \includegraphics[width=\textwidth]{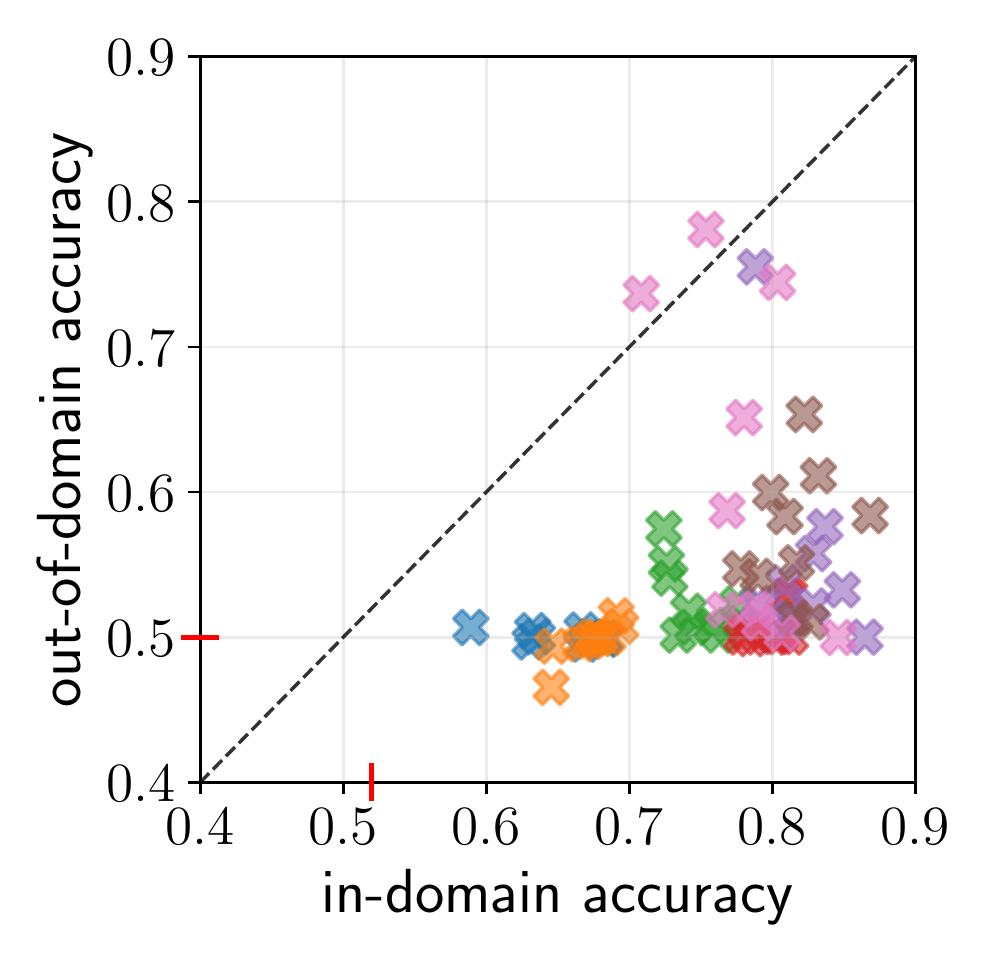}
        \caption{64 -- in-domain}
    \end{subfigure}
    ~
    \begin{subfigure}[b]{0.31\textwidth}
        \centering
        \includegraphics[width=\textwidth]{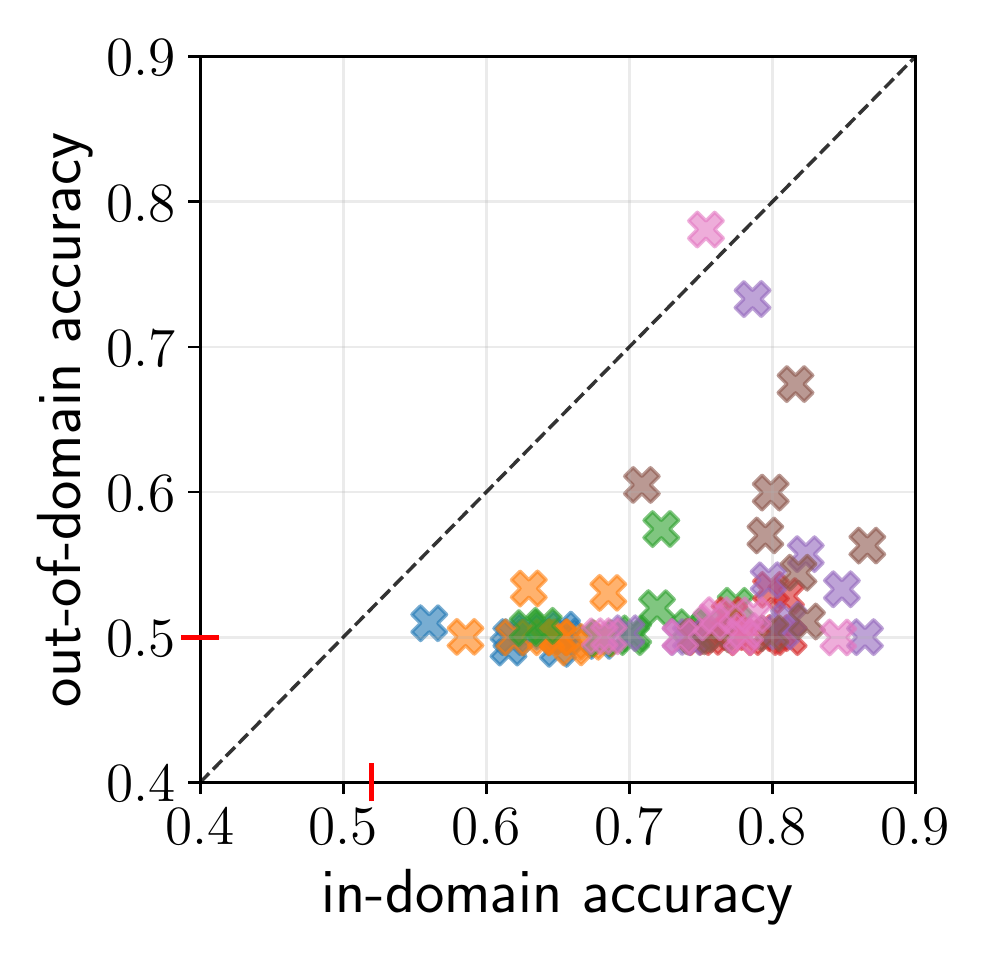}
        \caption{64 -- last checkpoint}
    \end{subfigure}
    ~
    \begin{subfigure}[b]{0.31\textwidth}
        \centering
        \includegraphics[width=\textwidth]{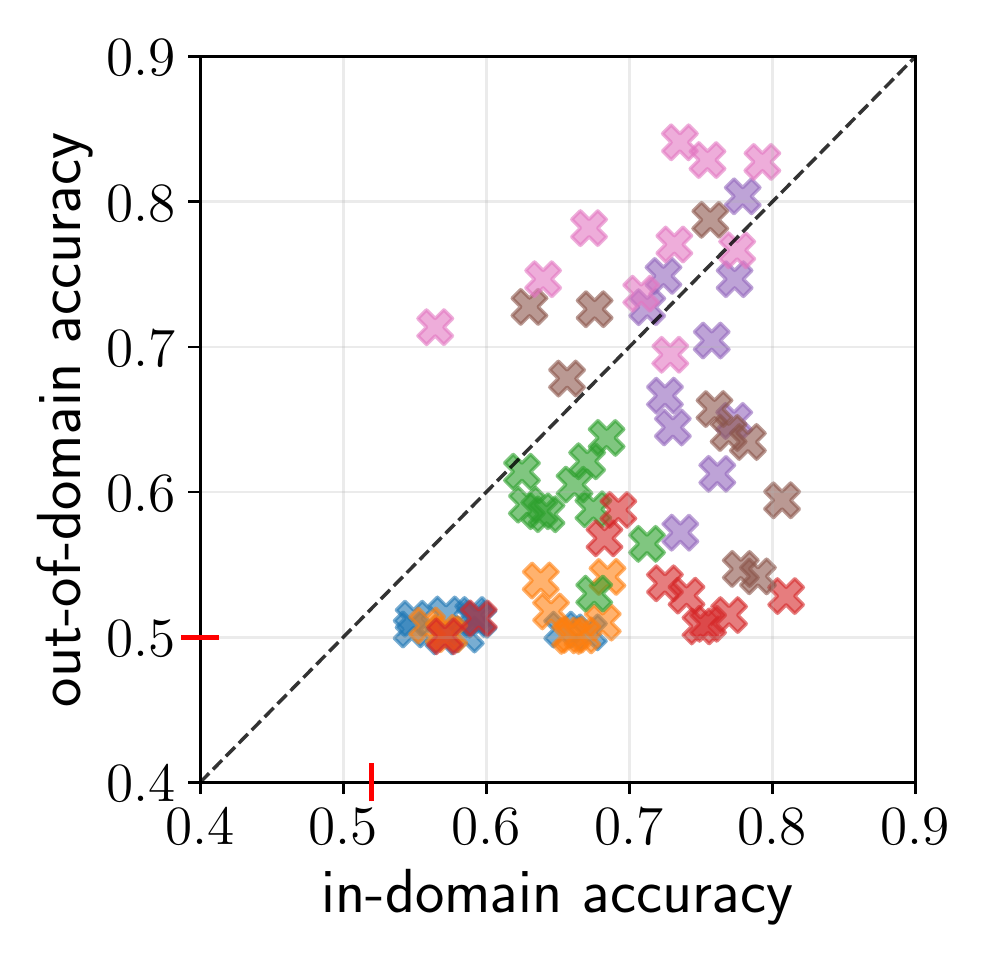}
        \caption{64 -- out-of-domain}
    \end{subfigure}
    \\
    \begin{subfigure}[b]{0.31\textwidth}
        \centering
        \includegraphics[width=\textwidth]{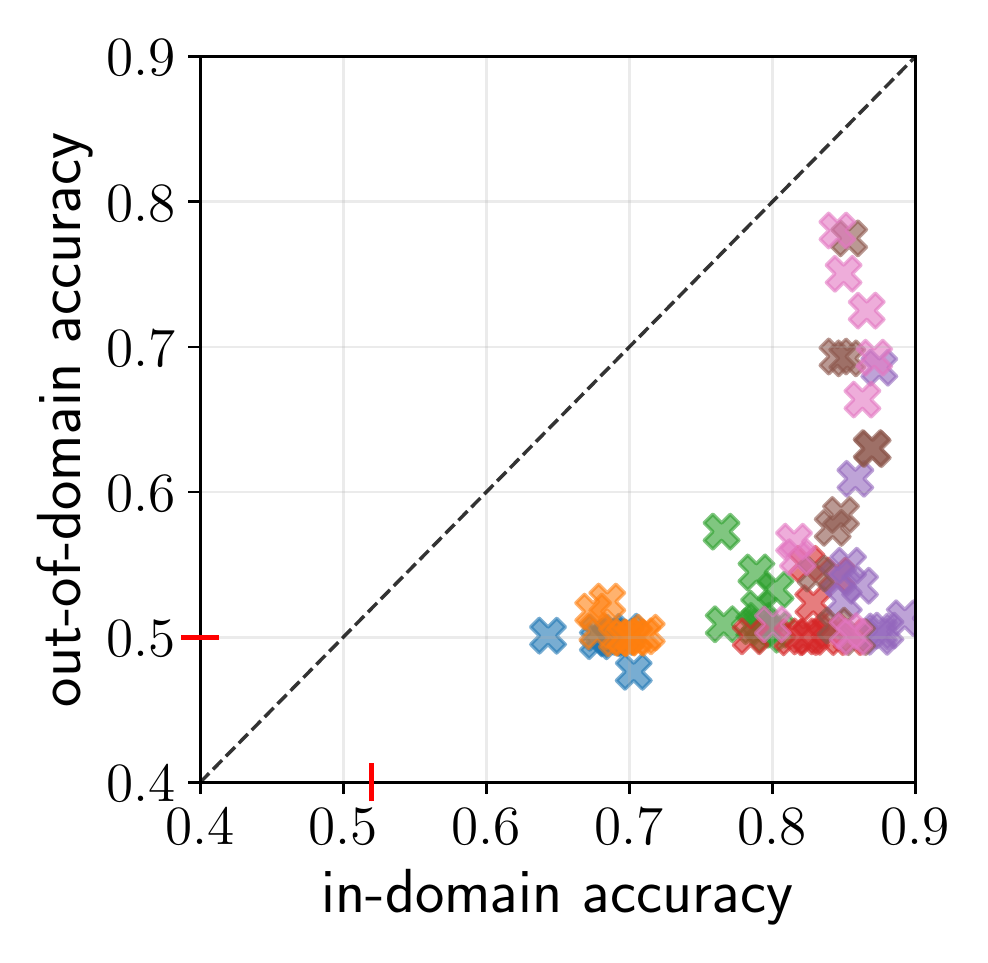}
        \caption{128 -- in-domain}
    \end{subfigure}
    ~
    \begin{subfigure}[b]{0.31\textwidth}
        \centering
        \includegraphics[width=\textwidth]{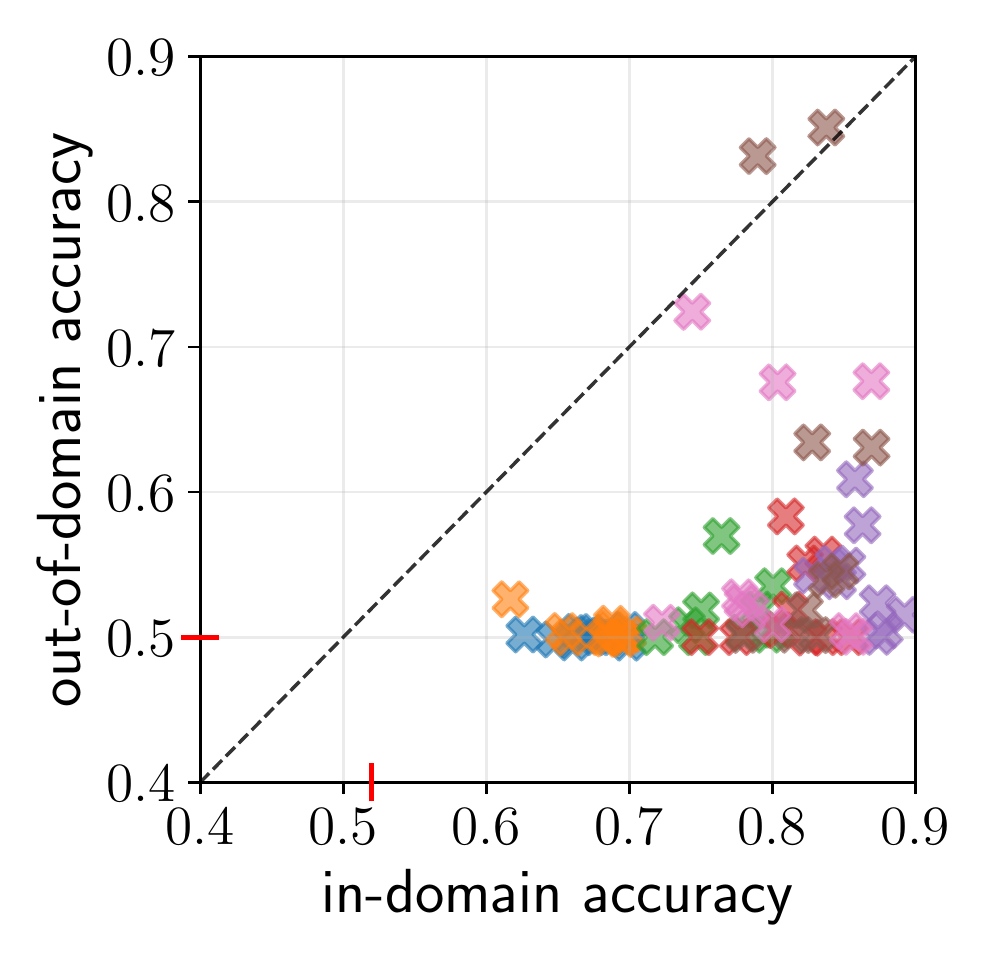}
        \caption{128 -- last checkpoint}
    \end{subfigure}
    ~
    \begin{subfigure}[b]{0.31\textwidth}
        \centering
        \includegraphics[width=\textwidth]{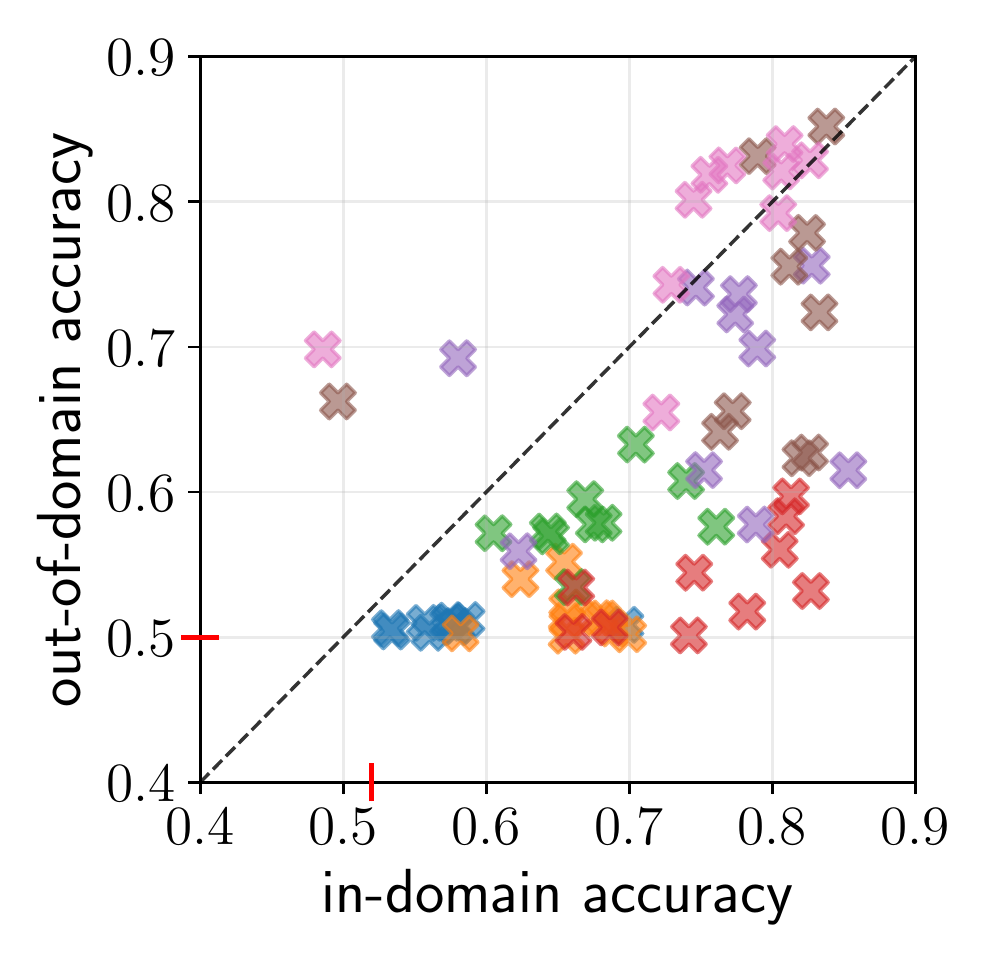}
        \caption{128 -- out-of-domain}
    \end{subfigure}
    
    \caption{\textbf{Relationship between in-domain and out-of-domain performance of PBFT on MNLI} for OPT models of various sizes. Rows vary amount of training data. Columns vary model selection strategy.
    Colors indicate model size. We fine-tune 10 models per setting varying only the data seed.
    \textcolor{red}{$\boldsymbol{-}$} in the x- and y-axis indicates the performance of the majority class label.
    }
    \label{fig:appendix-ft-model-selection-mnli}
\end{figure*}

\begin{figure*}[h]
    \centering
    \begin{subfigure}[b]{0.31\textwidth}
        \centering
        \includegraphics[width=\textwidth]{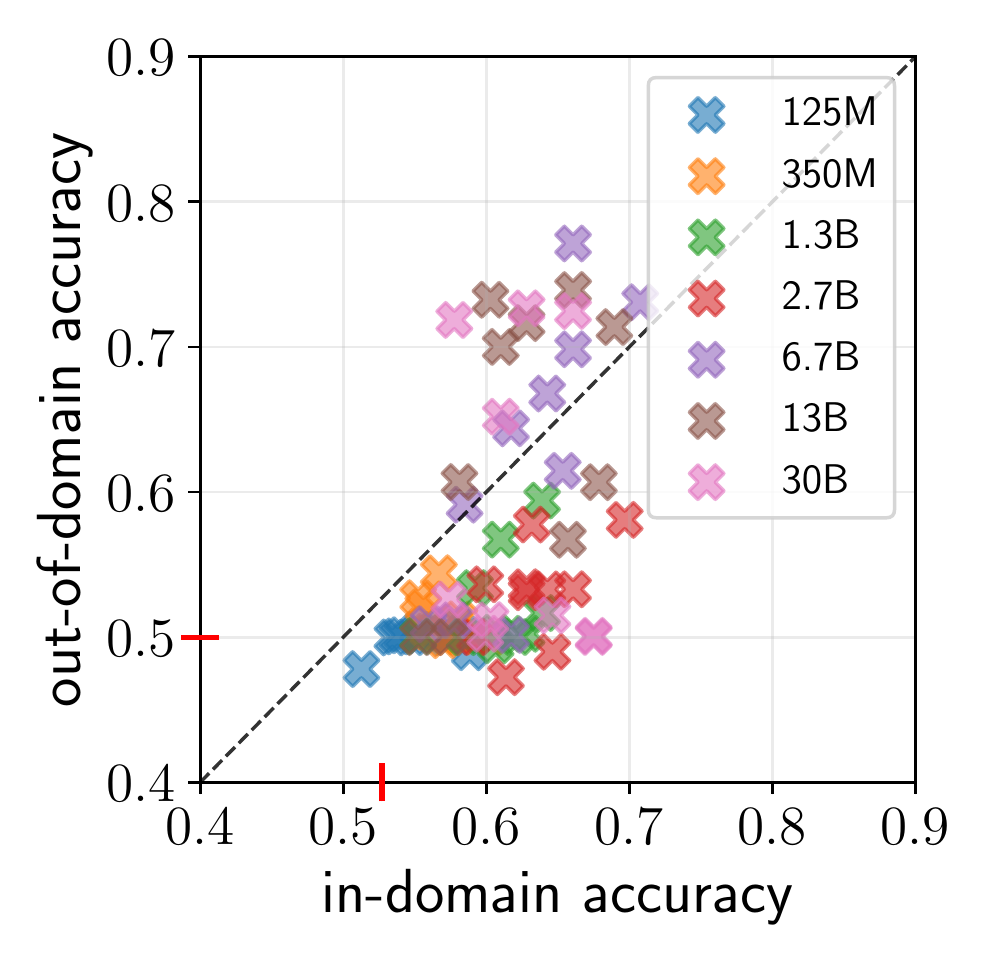}
        \caption{16 -- in-domain}
    \end{subfigure}
    ~
    \begin{subfigure}[b]{0.31\textwidth}
        \centering
        \includegraphics[width=\textwidth]{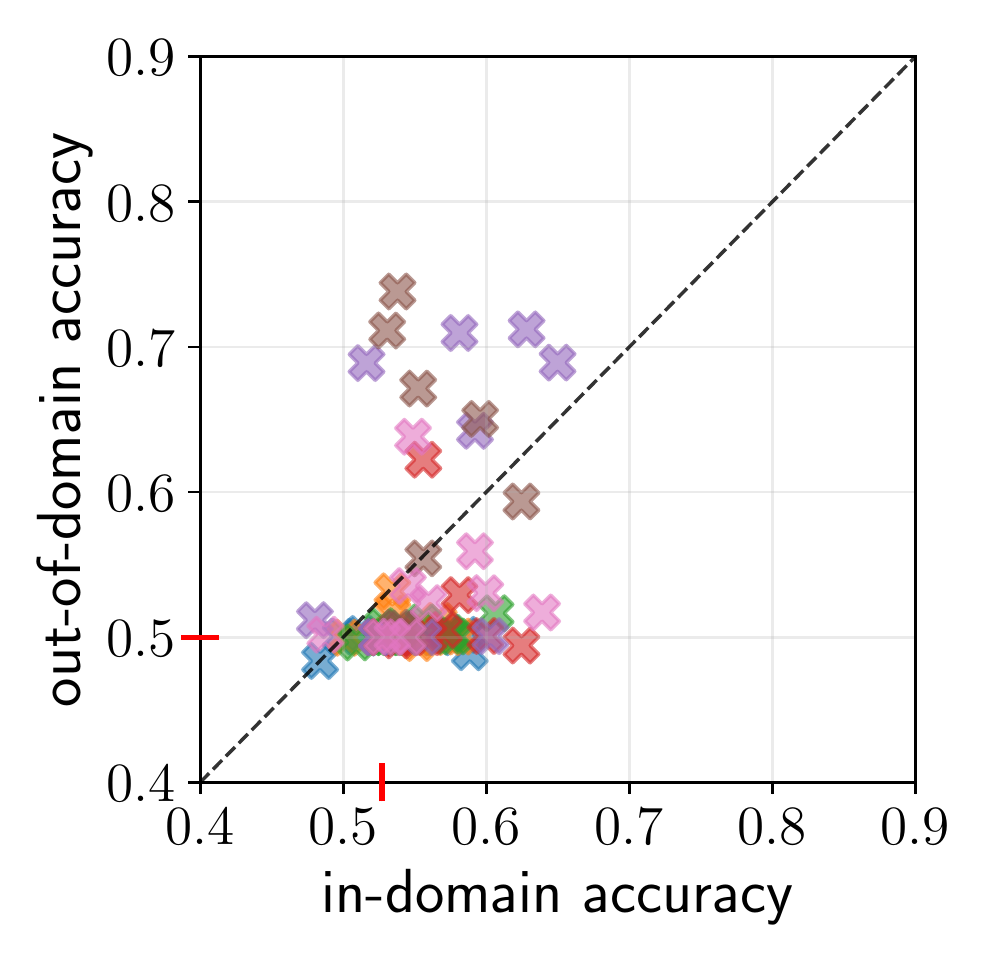}
        \caption{16 -- last checkpoint}
    \end{subfigure}
    ~
    \begin{subfigure}[b]{0.31\textwidth}
        \centering
        \includegraphics[width=\textwidth]{figures/ft/all-models_rte_16_best_out-of-domain_pattern-verbalizer-ft.pdf}
        \caption{16 -- out-of-domain}
    \end{subfigure}
    \\
    \begin{subfigure}[b]{0.31\textwidth}
        \centering
        \includegraphics[width=\textwidth]{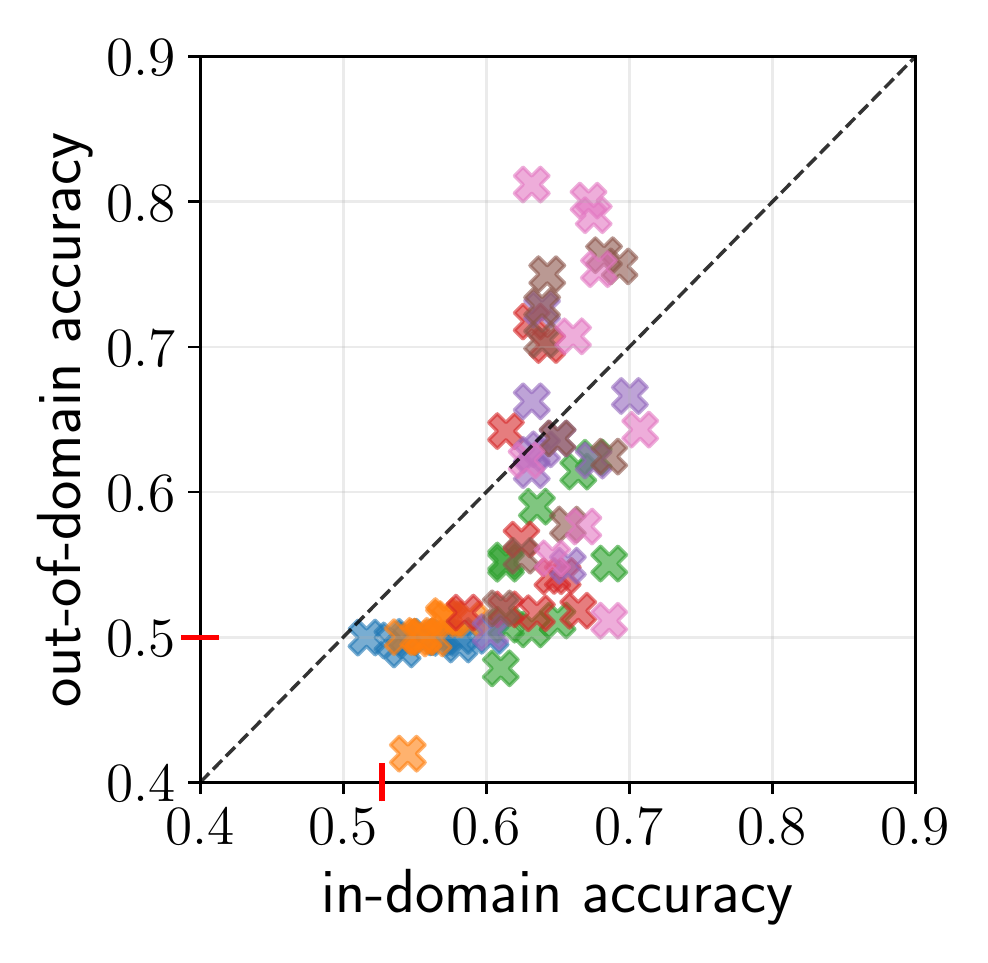}
        \caption{32 -- in-domain}
    \end{subfigure}
    ~
    \begin{subfigure}[b]{0.31\textwidth}
        \centering
        \includegraphics[width=\textwidth]{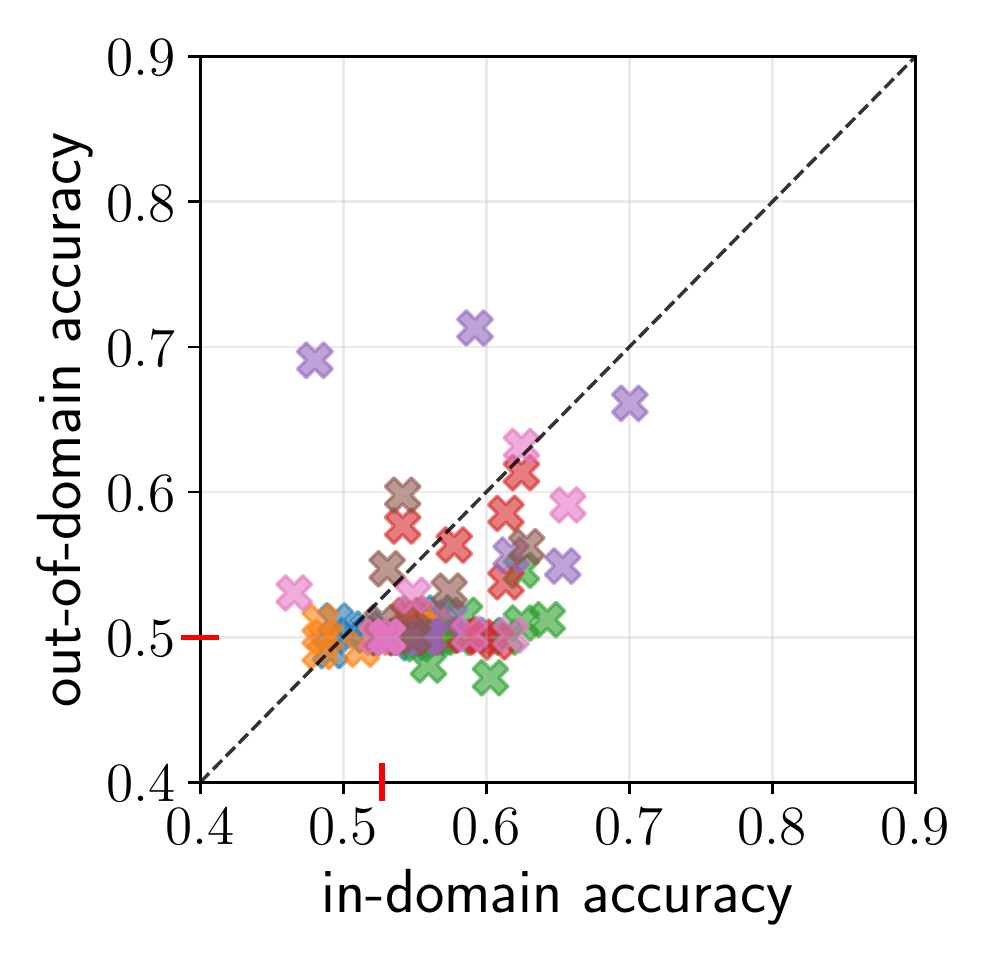}
        \caption{32 -- last checkpoint}
    \end{subfigure}
    ~
    \begin{subfigure}[b]{0.31\textwidth}
        \centering
        \includegraphics[width=\textwidth]{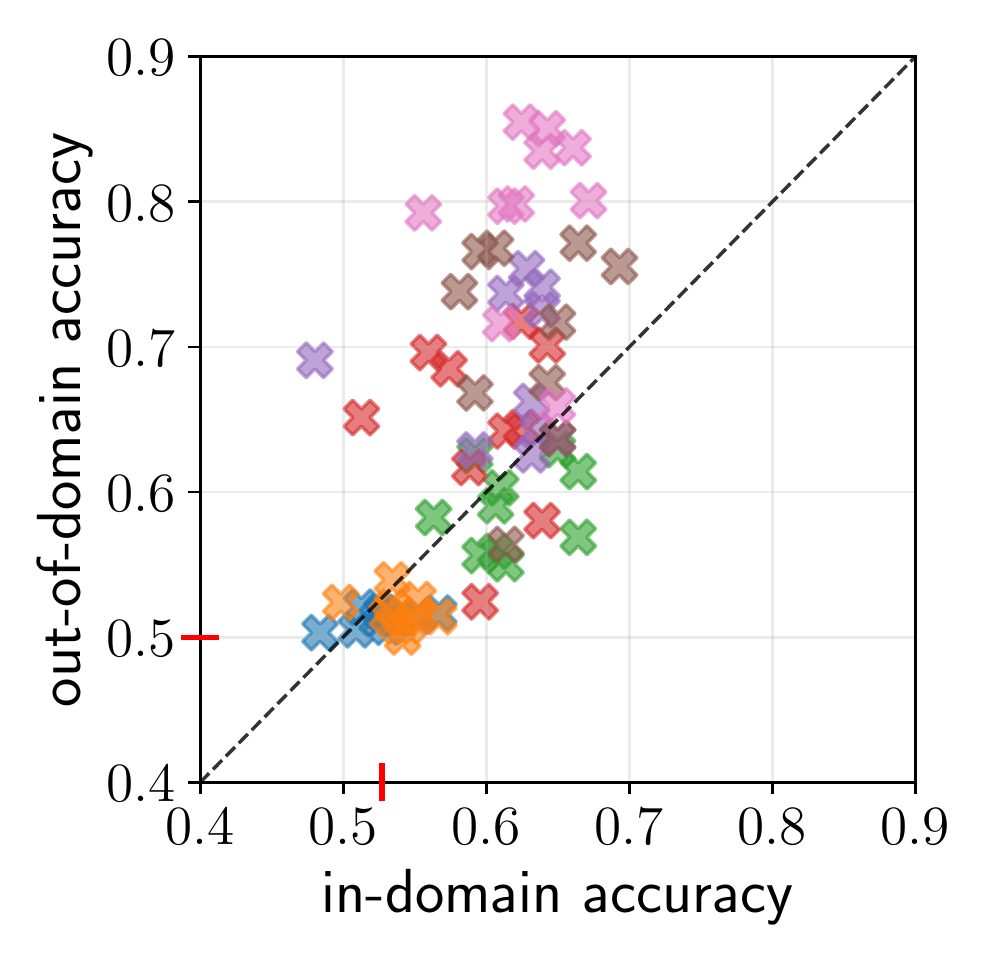}
        \caption{32 -- out-of-domain}
    \end{subfigure}
    \\
    \begin{subfigure}[b]{0.31\textwidth}
        \centering
        \includegraphics[width=\textwidth]{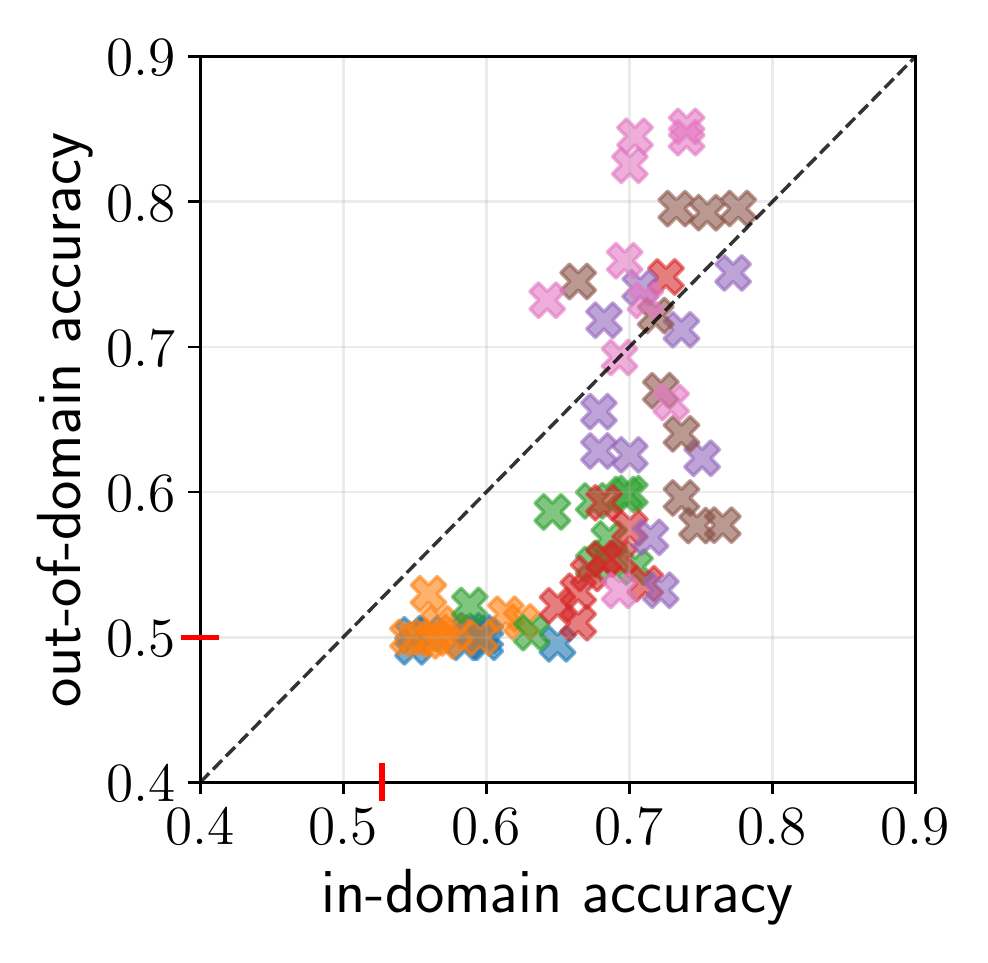}
        \caption{64 -- in-domain}
    \end{subfigure}
    ~
    \begin{subfigure}[b]{0.31\textwidth}
        \centering
        \includegraphics[width=\textwidth]{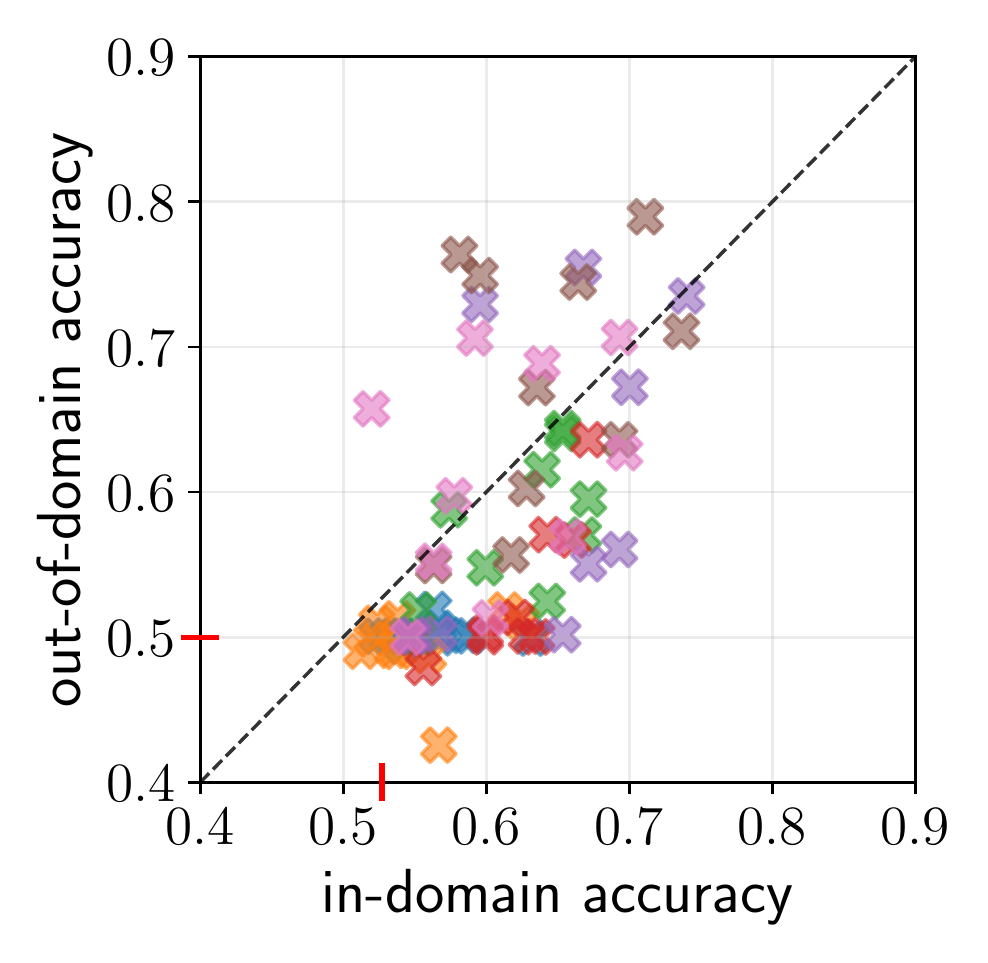}
        \caption{64 -- last checkpoint}
    \end{subfigure}
    ~
    \begin{subfigure}[b]{0.31\textwidth}
        \centering
        \includegraphics[width=\textwidth]{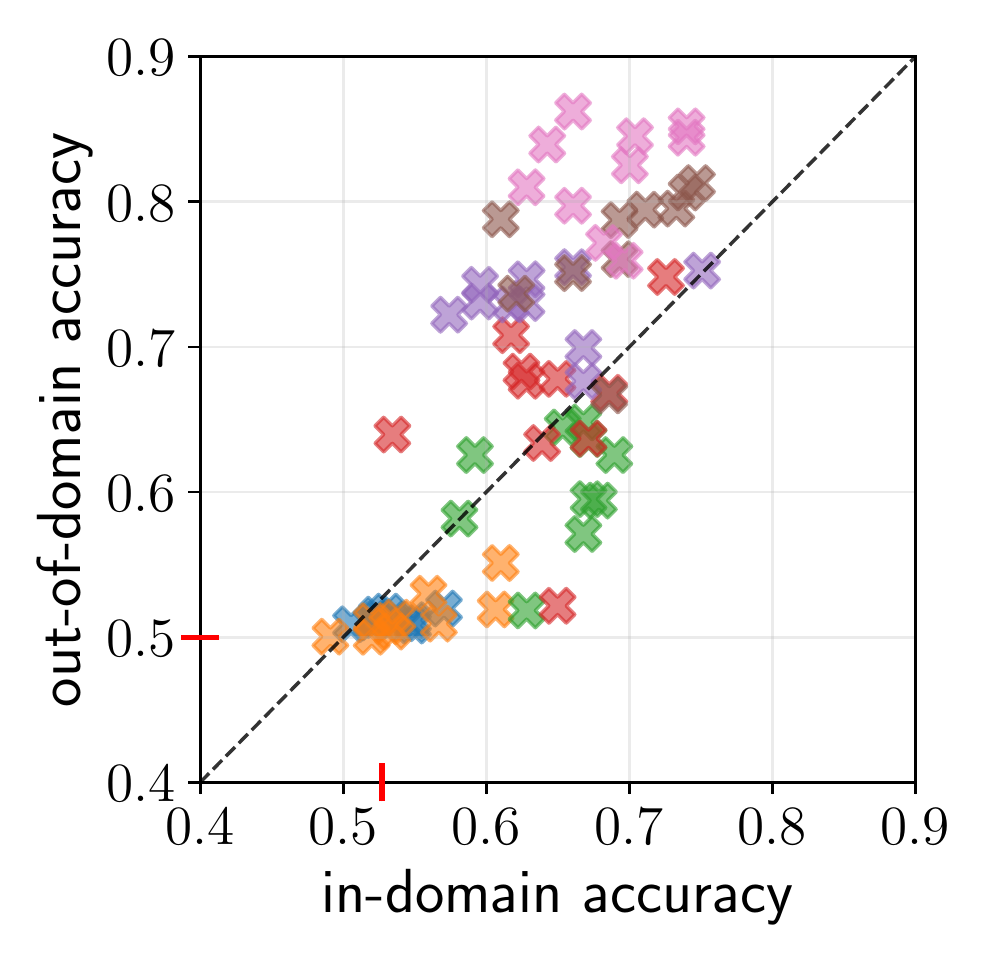}
        \caption{64 -- out-of-domain}
    \end{subfigure}
    \\
    \begin{subfigure}[b]{0.31\textwidth}
        \centering
        \includegraphics[width=\textwidth]{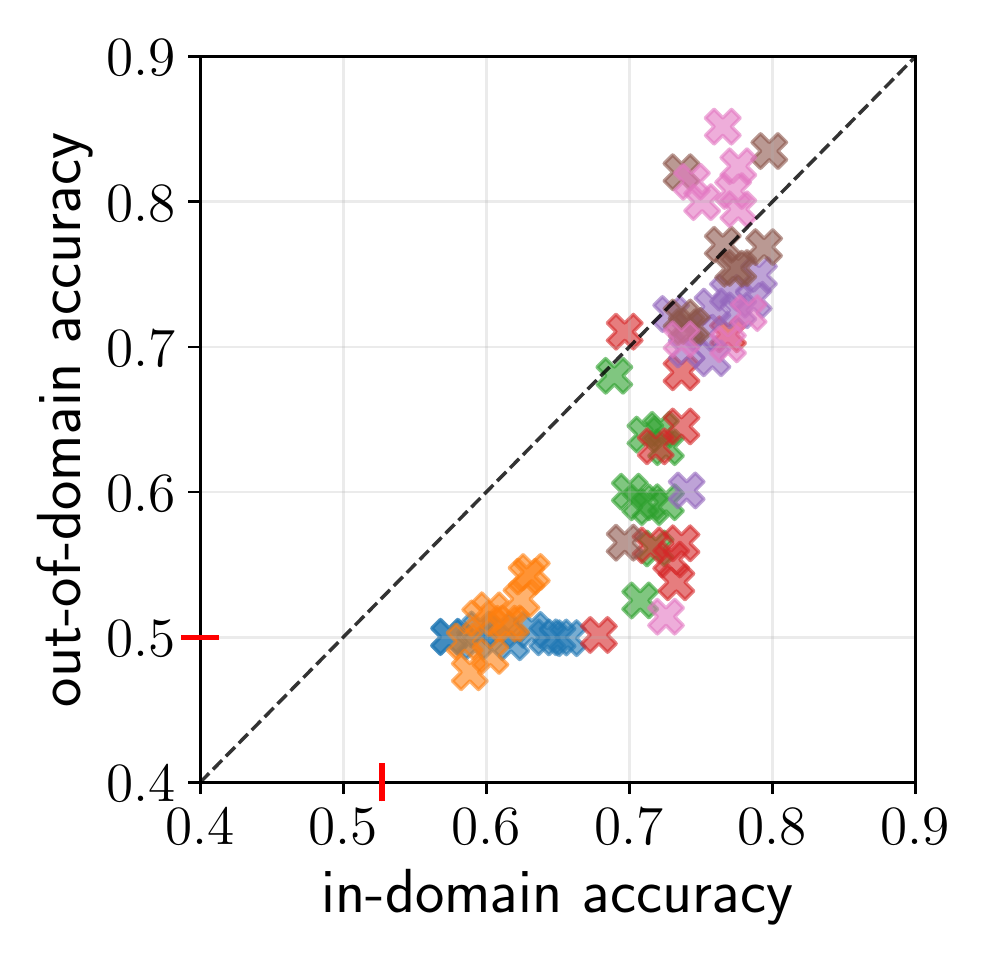}
        \caption{128 -- in-domain}
    \end{subfigure}
    ~
    \begin{subfigure}[b]{0.31\textwidth}
        \centering
        \includegraphics[width=\textwidth]{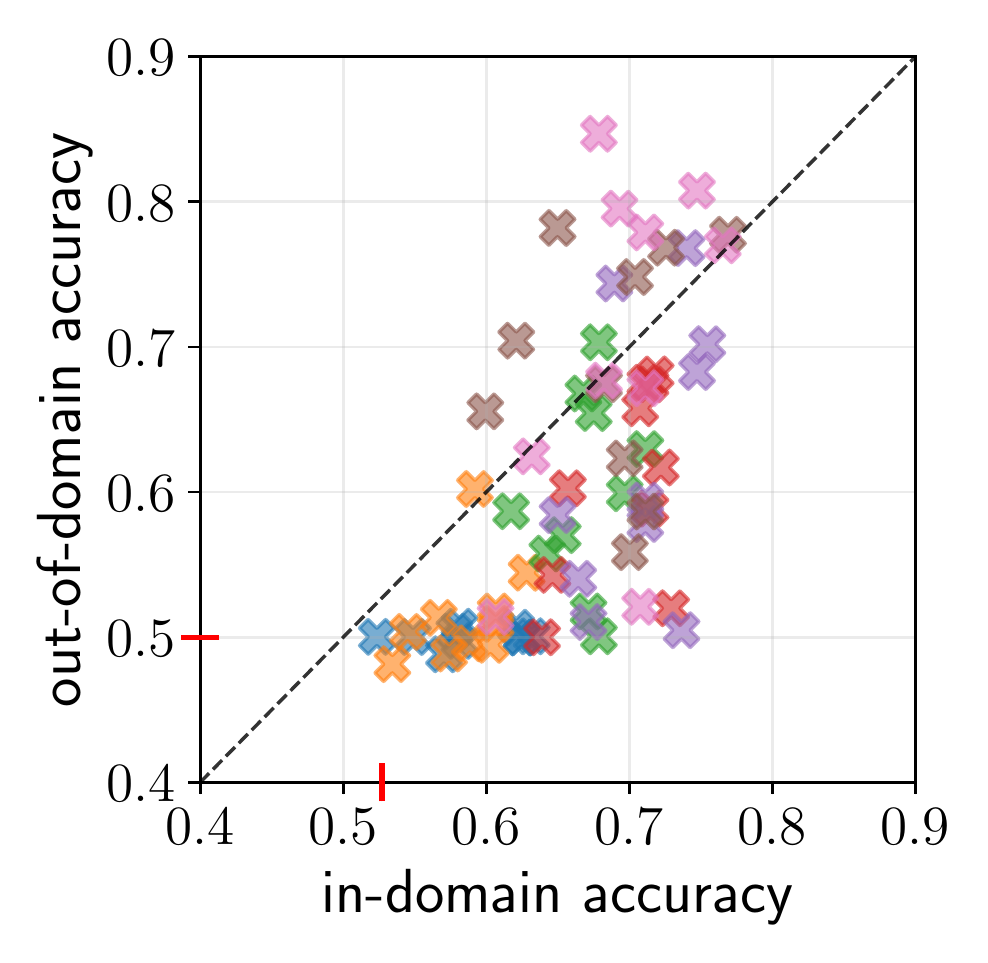}
        \caption{128 -- last checkpoint}
    \end{subfigure}
    ~
    \begin{subfigure}[b]{0.31\textwidth}
        \centering
        \includegraphics[width=\textwidth]{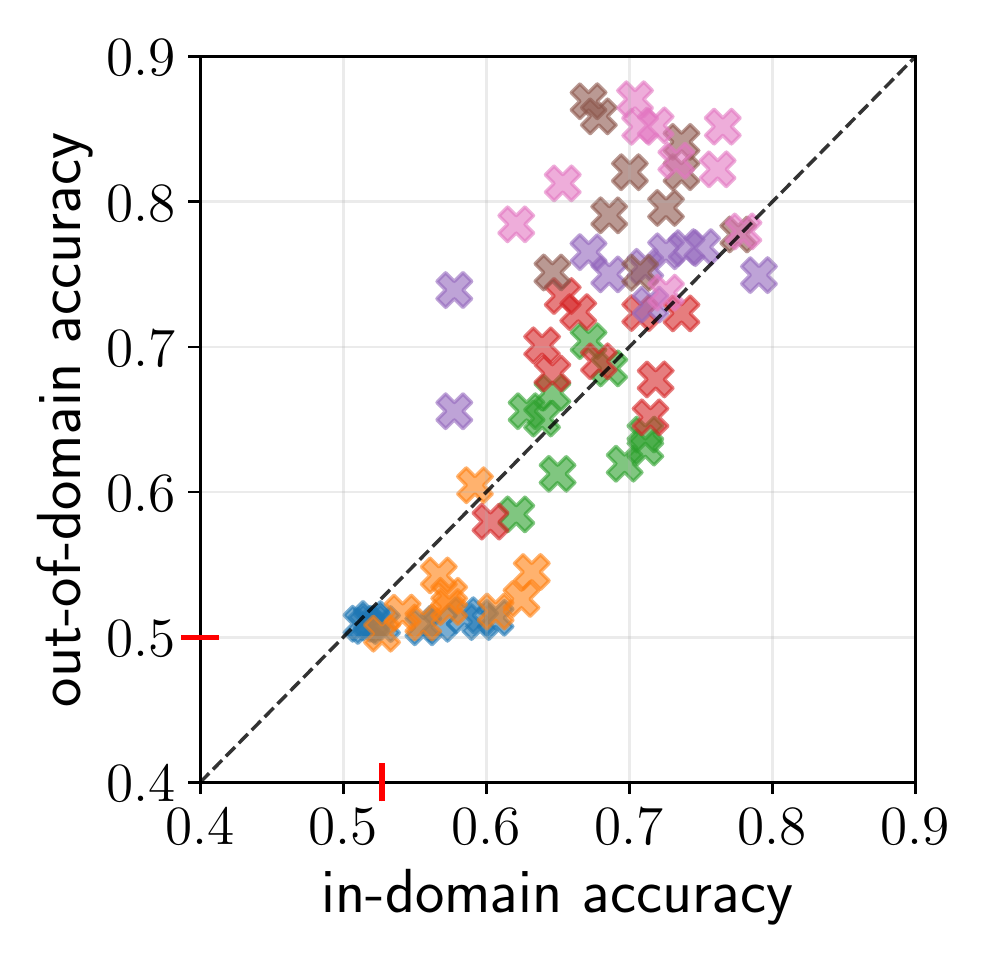}
        \caption{128 -- out-of-domain}
    \end{subfigure}
    
    \caption{\textbf{Relationship between in-domain and out-of-domain performance of PBFT on RTE} for OPT models of various sizes. Rows vary amount of training data. Columns vary model selection strategy.
    Colors indicate model size. We fine-tune 10 models per setting varying only the data seed.
    \textcolor{red}{$\boldsymbol{-}$} in the x- and y-axis indicates the performance of the majority class label.
    }
    \label{fig:appendix-ft-model-selection-rte}
\end{figure*}

\begin{figure*}[h]
    \centering
    \begin{subfigure}[b]{0.31\textwidth}
        \centering
        \includegraphics[width=\textwidth]{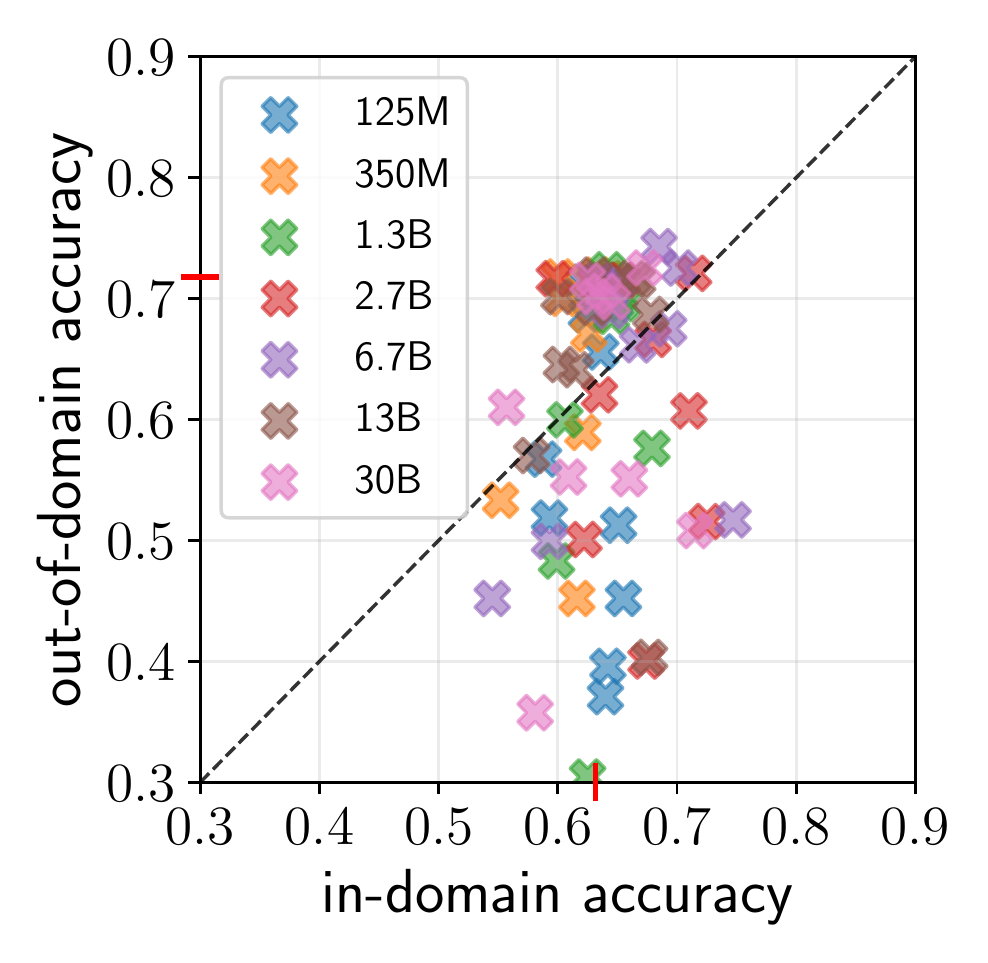}
        \caption{16 -- in-domain}
    \end{subfigure}
    ~
    \begin{subfigure}[b]{0.31\textwidth}
        \centering
        \includegraphics[width=\textwidth]{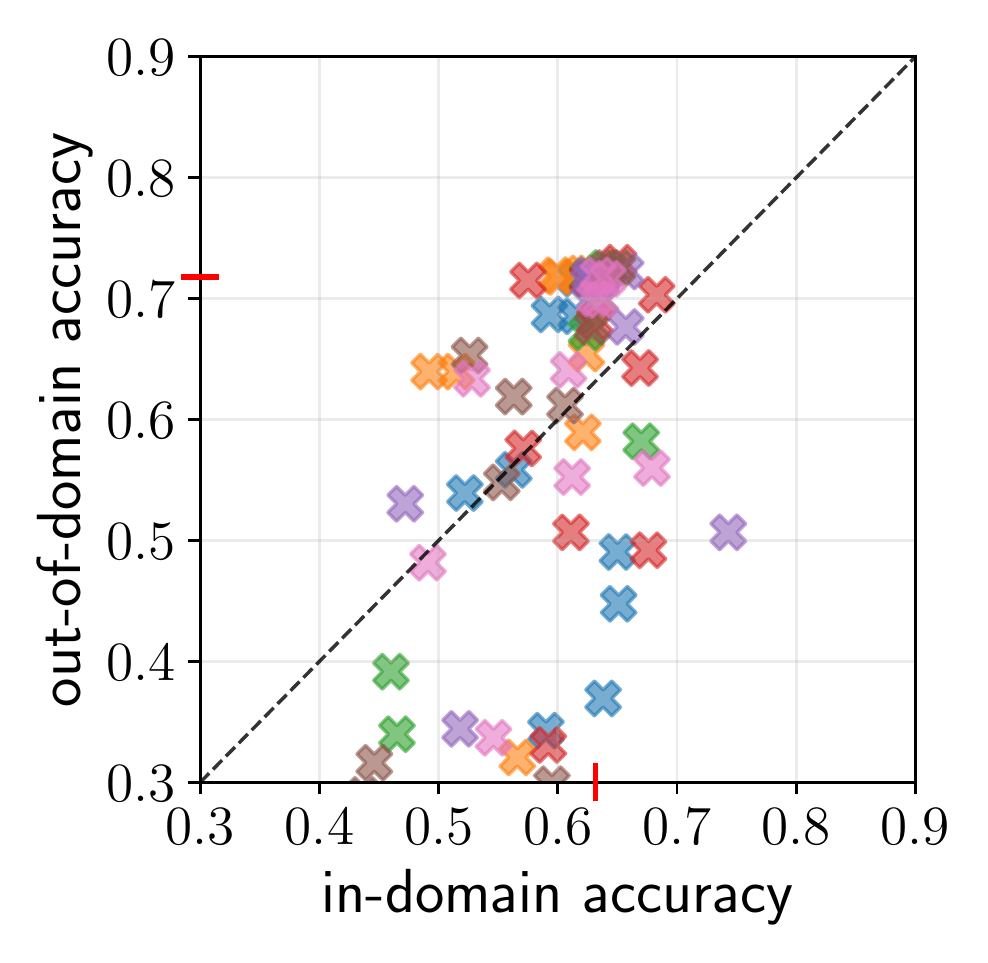}
        \caption{16 -- last checkpoint}
    \end{subfigure}
    ~
    \begin{subfigure}[b]{0.31\textwidth}
        \centering
        \includegraphics[width=\textwidth]{figures/ft/all-models_qqp_16_best_out-of-domain_pattern-verbalizer-ft.pdf}
        \caption{16 -- out-of-domain}
    \end{subfigure}
    \\
    \begin{subfigure}[b]{0.31\textwidth}
        \centering
        \includegraphics[width=\textwidth]{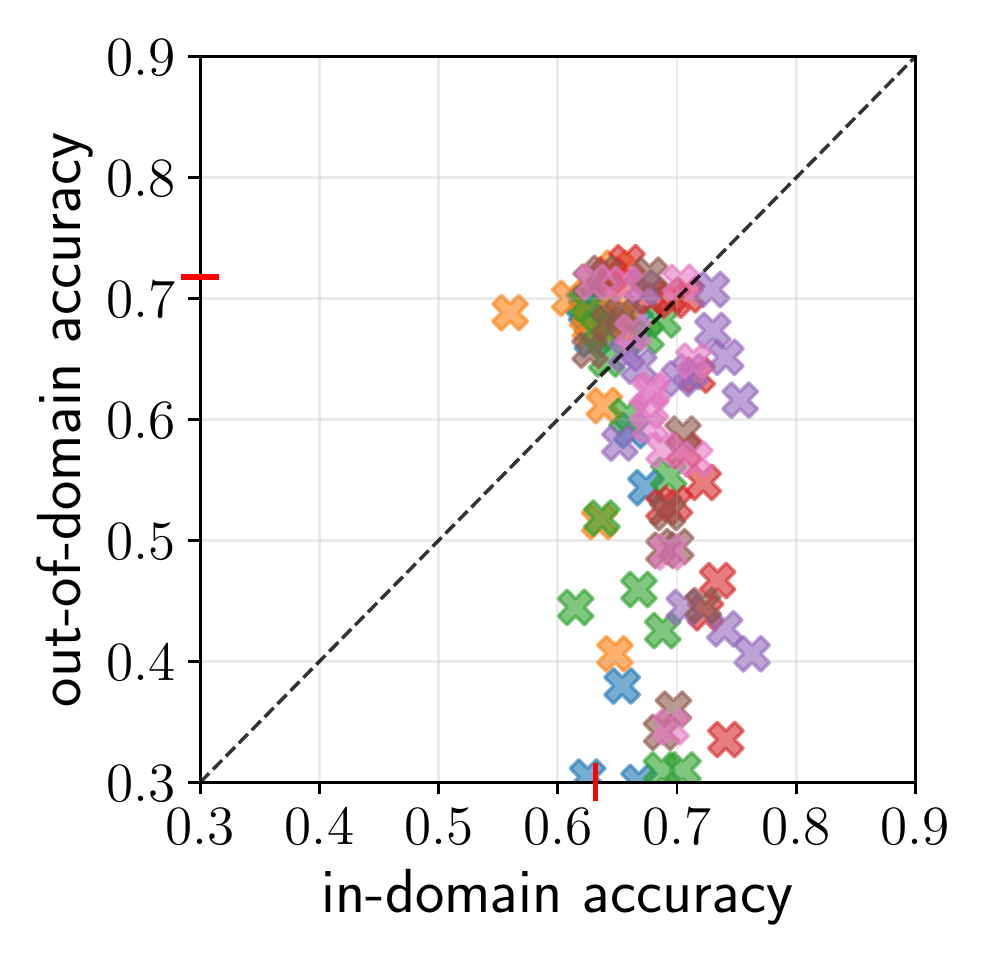}
        \caption{32 -- in-domain}
    \end{subfigure}
    ~
    \begin{subfigure}[b]{0.31\textwidth}
        \centering
        \includegraphics[width=\textwidth]{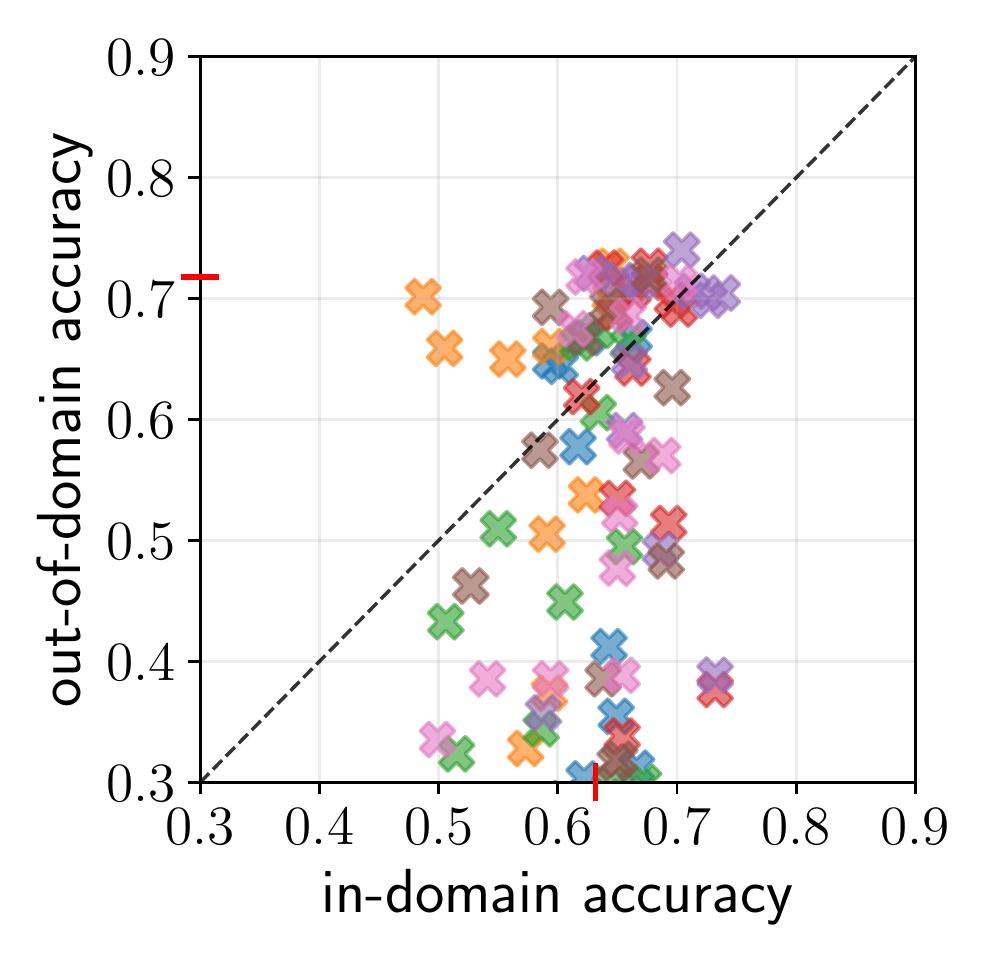}
        \caption{32 -- last checkpoint}
    \end{subfigure}
    ~
    \begin{subfigure}[b]{0.31\textwidth}
        \centering
        \includegraphics[width=\textwidth]{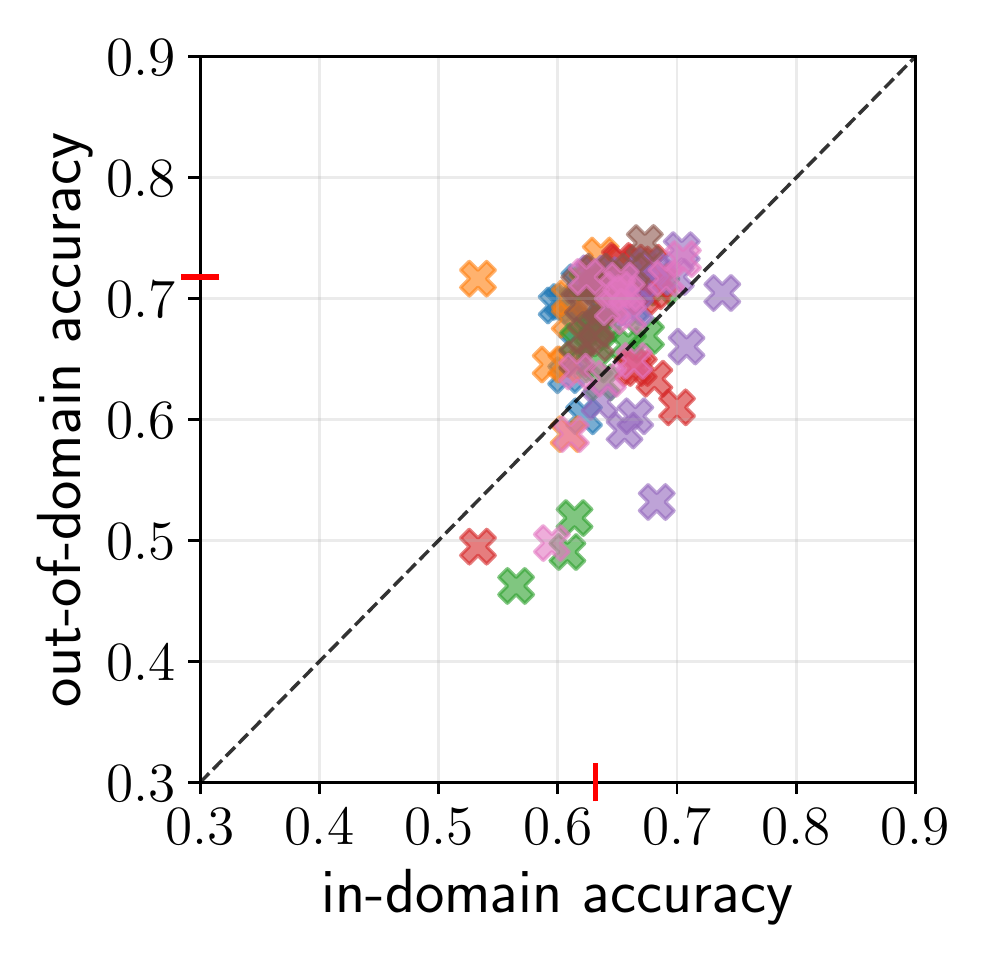}
        \caption{32 -- out-of-domain}
    \end{subfigure}
    \\
    \begin{subfigure}[b]{0.31\textwidth}
        \centering
        \includegraphics[width=\textwidth]{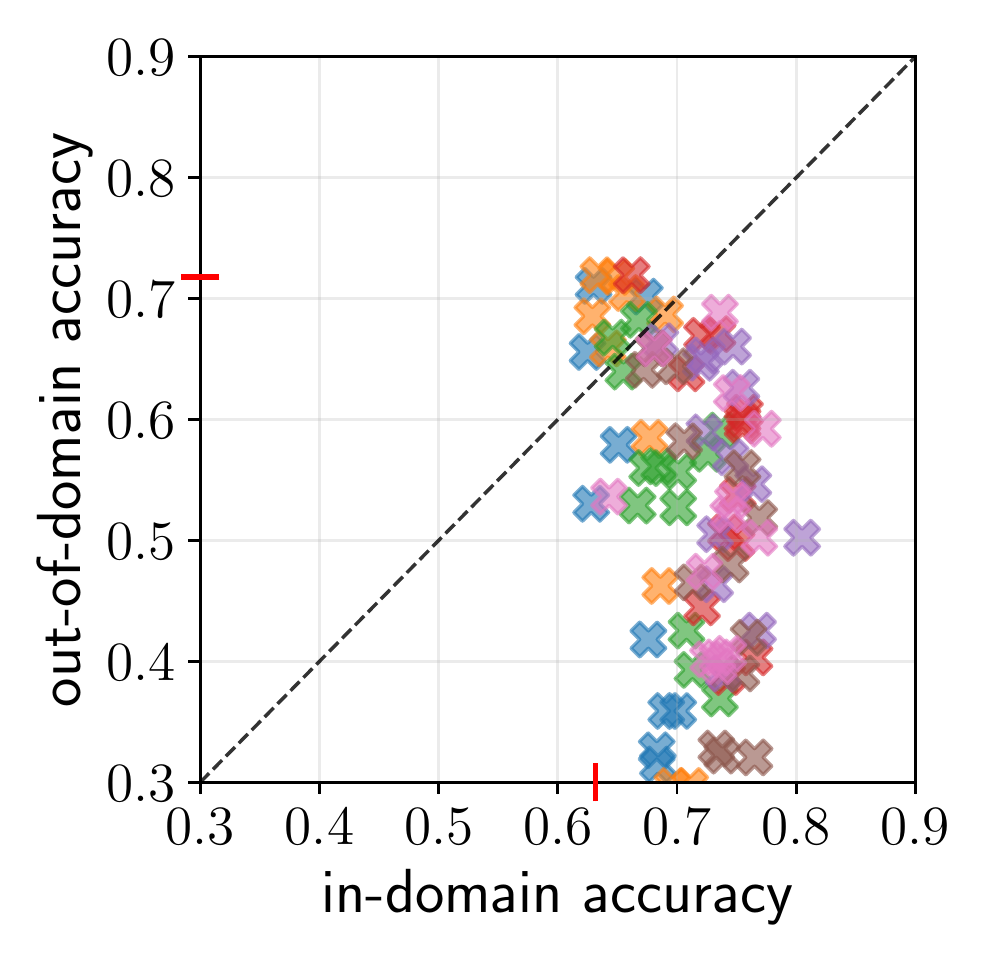}
        \caption{64 -- in-domain}
    \end{subfigure}
    ~
    \begin{subfigure}[b]{0.31\textwidth}
        \centering
        \includegraphics[width=\textwidth]{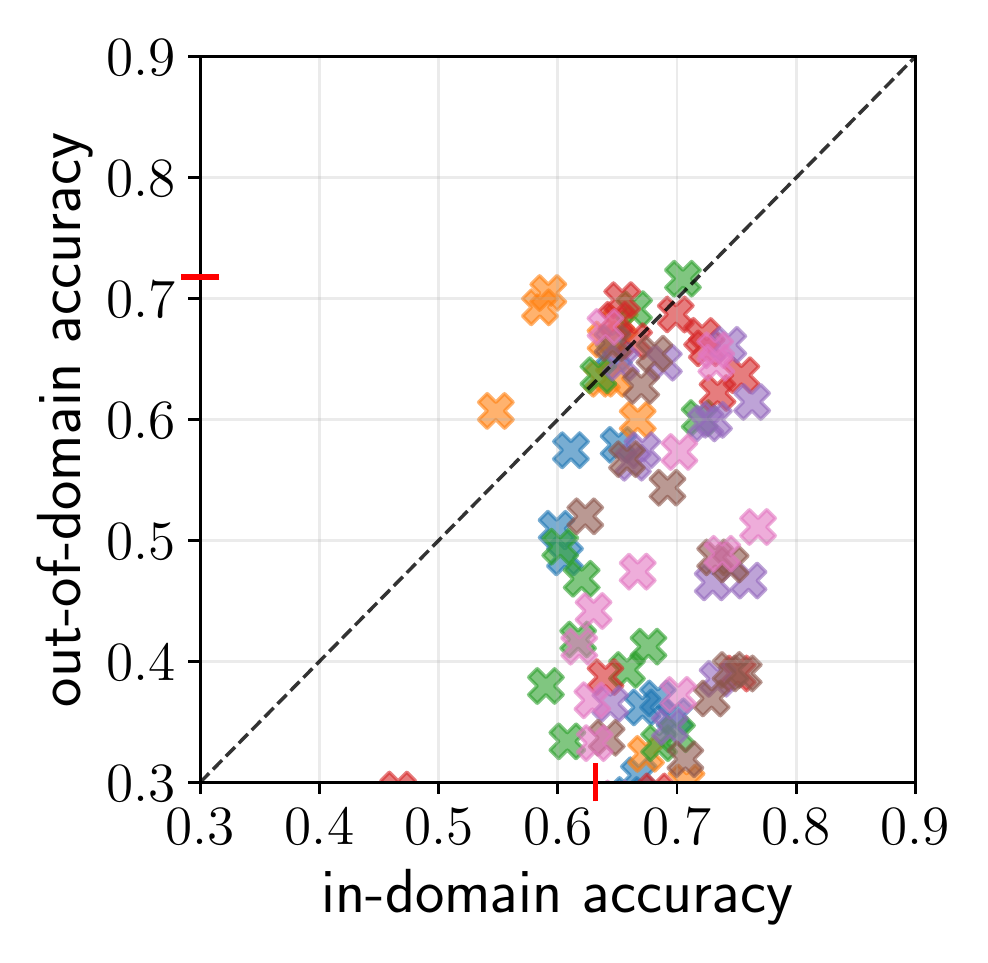}
        \caption{64 -- last checkpoint}
    \end{subfigure}
    ~
    \begin{subfigure}[b]{0.31\textwidth}
        \centering
        \includegraphics[width=\textwidth]{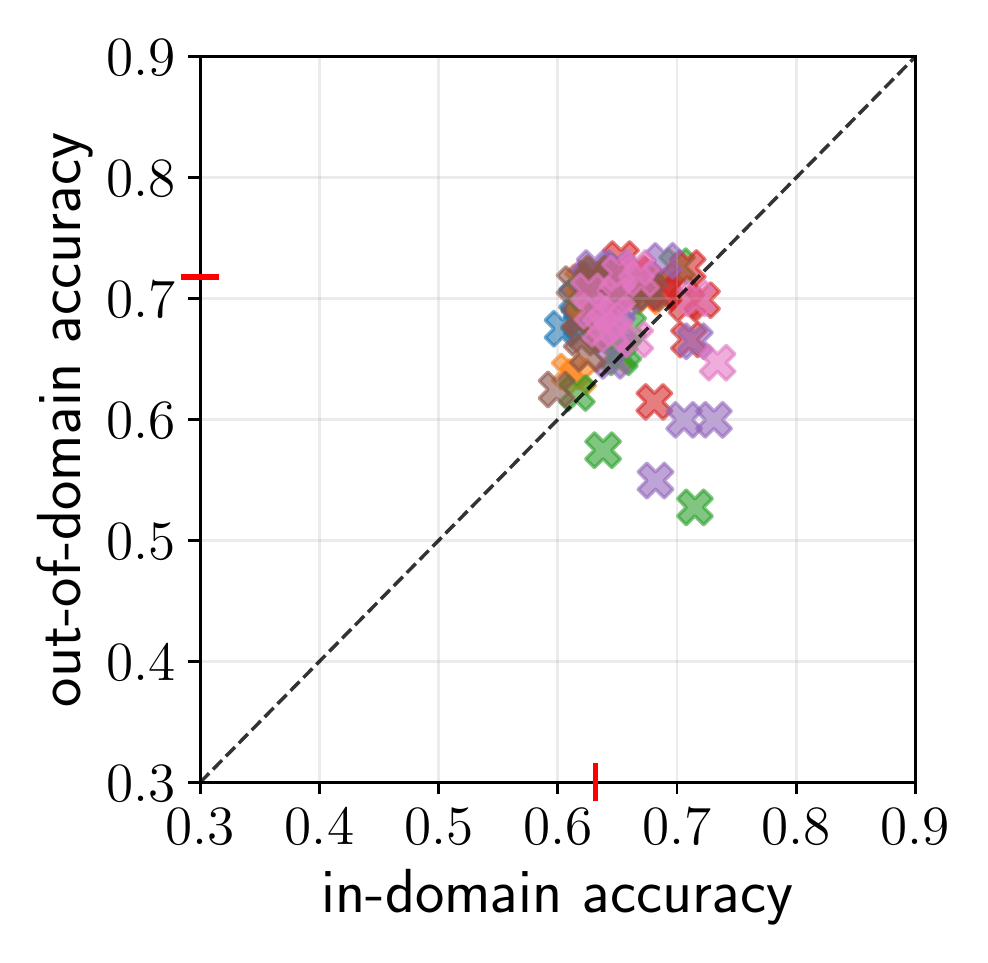}
        \caption{64 -- out-of-domain}
    \end{subfigure}
    \\
    \begin{subfigure}[b]{0.31\textwidth}
        \centering
        \includegraphics[width=\textwidth]{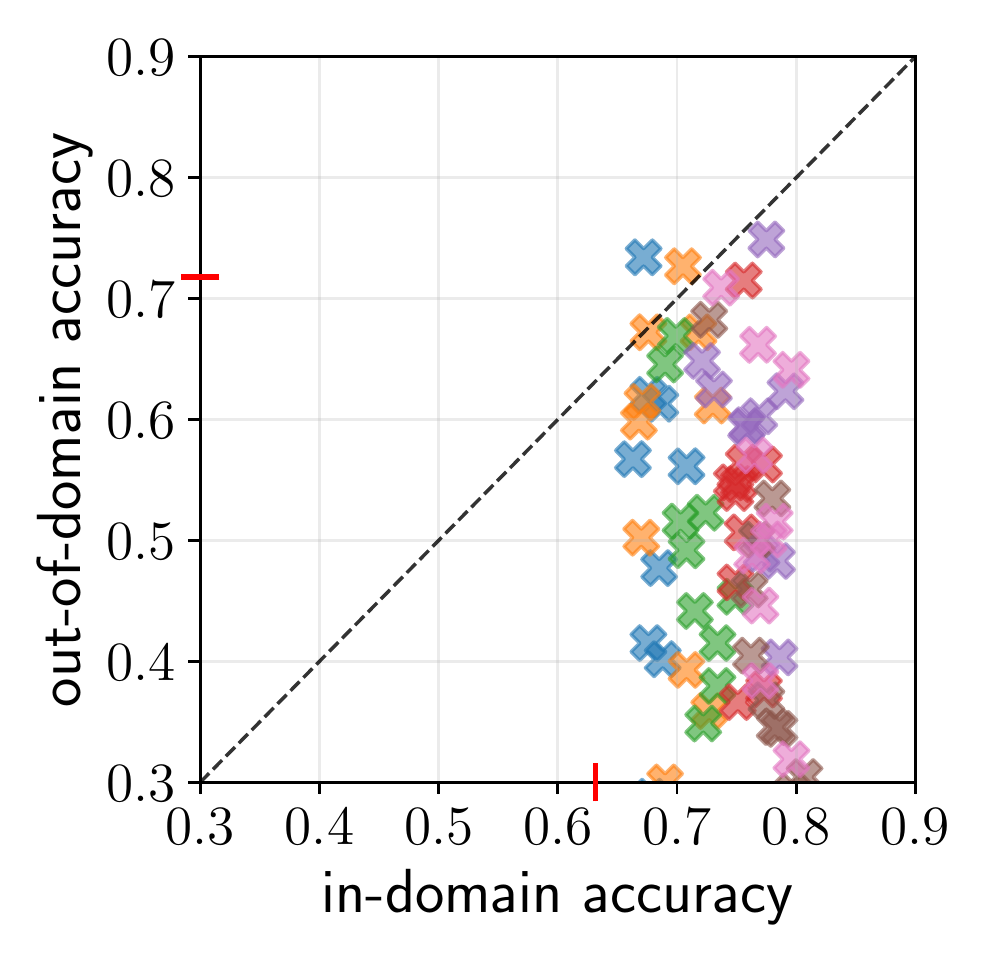}
        \caption{128 -- in-domain}
    \end{subfigure}
    ~
    \begin{subfigure}[b]{0.31\textwidth}
        \centering
        \includegraphics[width=\textwidth]{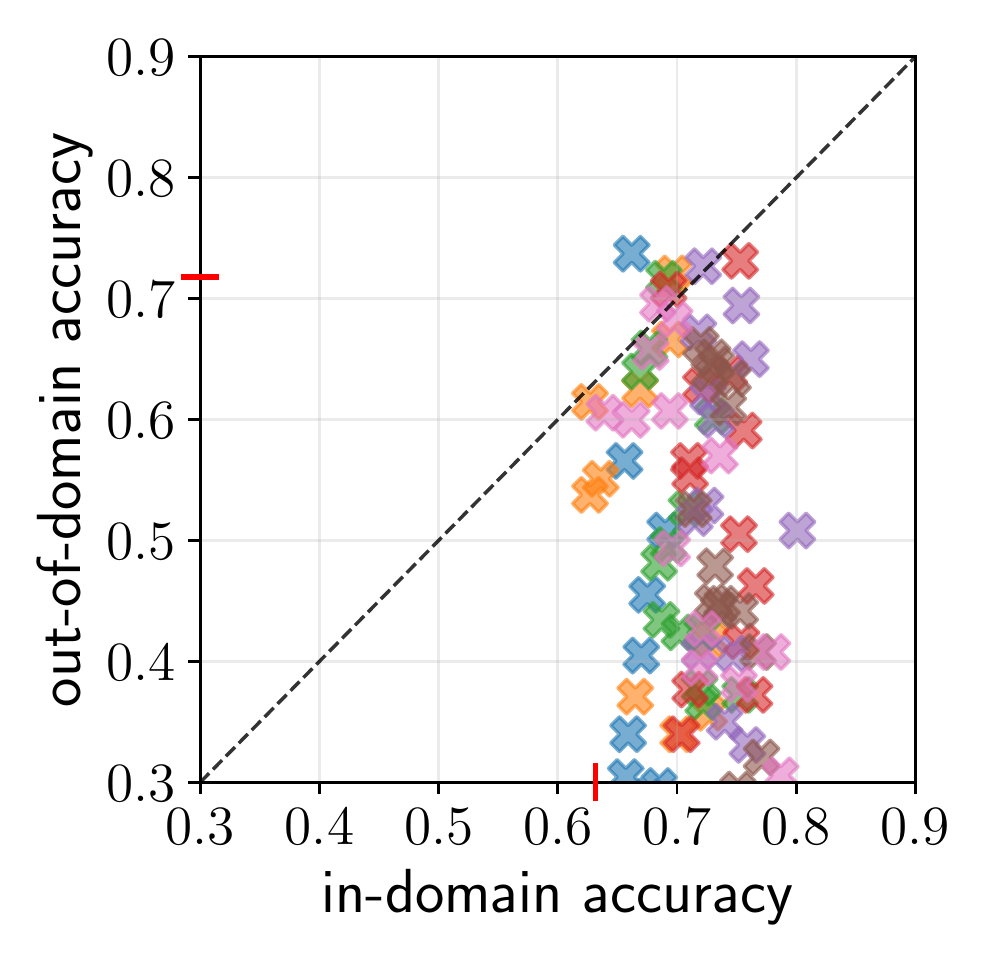}
        \caption{128 -- last checkpoint}
    \end{subfigure}
    ~
    \begin{subfigure}[b]{0.31\textwidth}
        \centering
        \includegraphics[width=\textwidth]{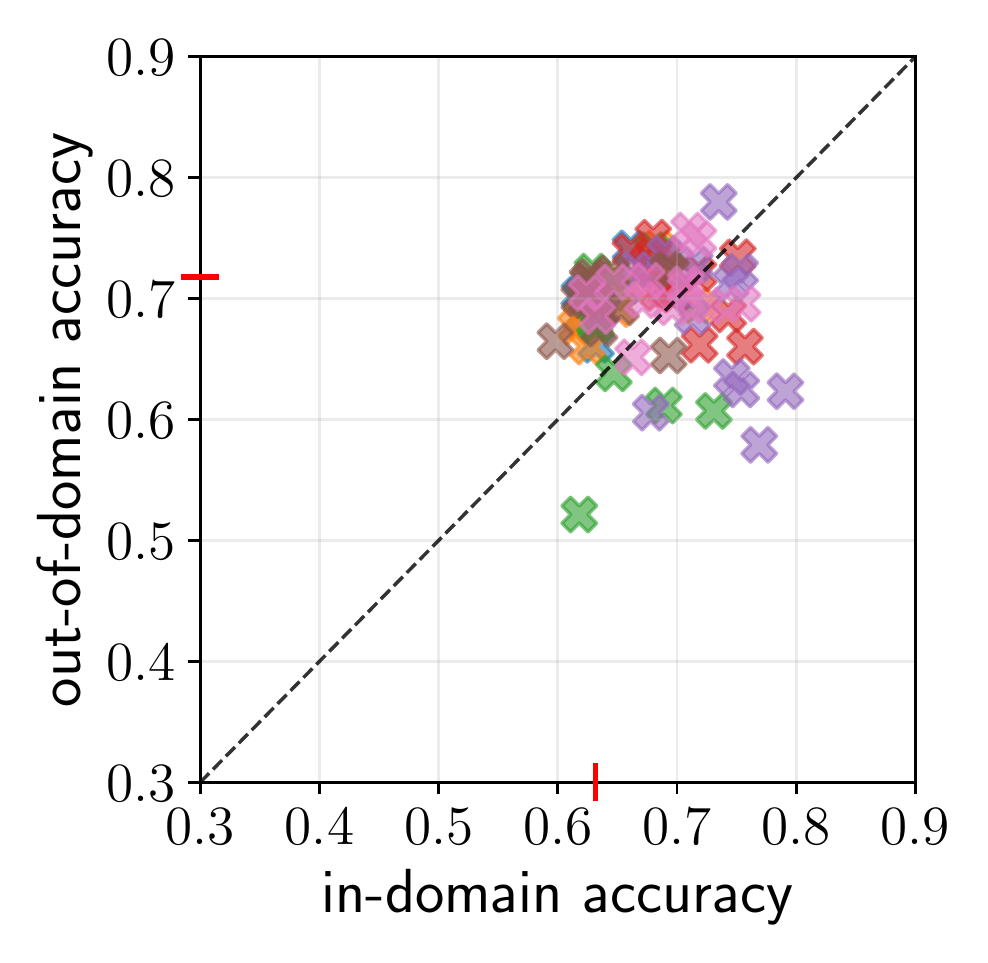}
        \caption{128 -- out-of-domain}
    \end{subfigure}
    
    \caption{\textbf{Relationship between in-domain and out-of-domain performance of PBFT on QQP} for OPT models of various sizes. Rows vary amount of training data. Columns vary model selection strategy.
    Colors indicate model size. We fine-tune 10 models per setting varying only the data seed.
    \textcolor{red}{$\boldsymbol{-}$} in the x- and y-axis indicates the performance of the majority class label.
    }
    \label{fig:appendix-ft-model-selection-qqp}
\end{figure*}

We provide all FT results in Figures \ref{fig:appendix-ft-model-selection-mnli}, \ref{fig:appendix-ft-model-selection-rte}, and \ref{fig:appendix-ft-model-selection-qqp}. When comparing results across rows, we see the impact of the number of training examples on the results. Comparing results across columns demonstrates the importance of model selection for in-domain and out-of-domain performance. 

\begin{figure*}[h]
    \centering
    \begin{subfigure}[b]{0.31\textwidth}
        \centering
        \includegraphics[width=\textwidth]{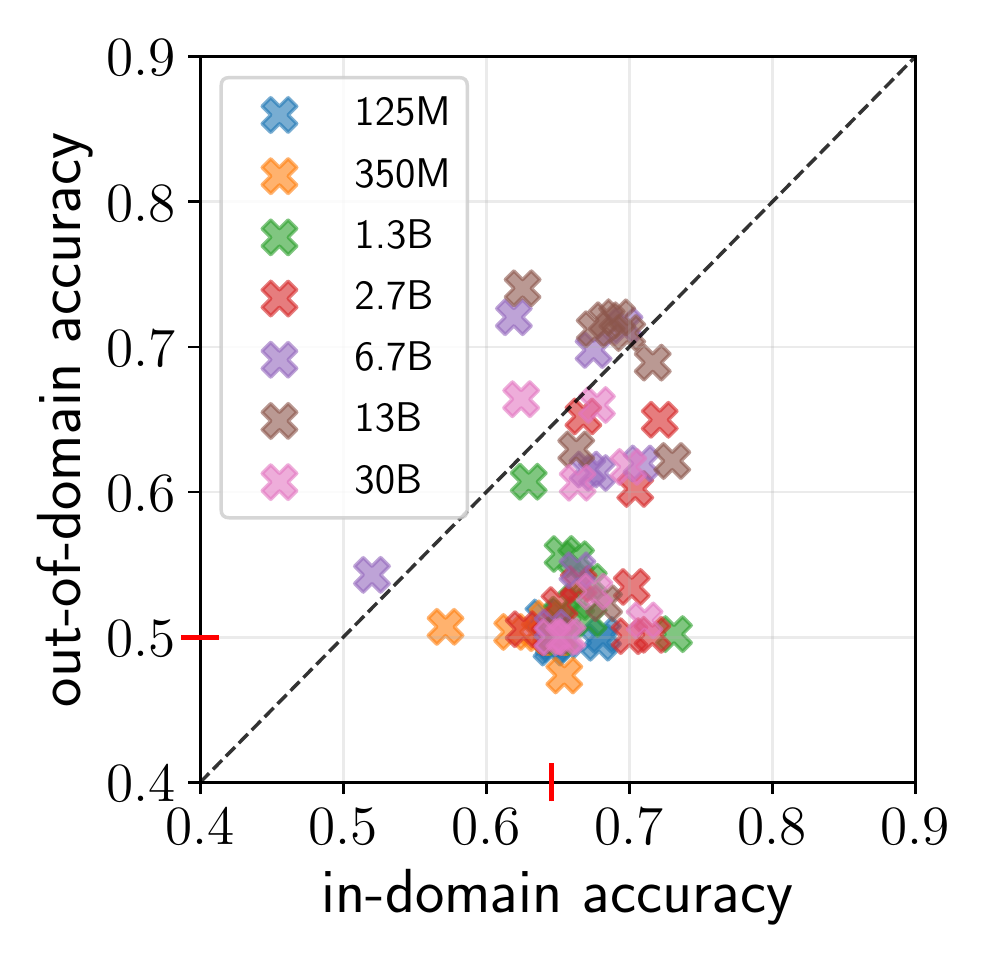}
        \caption{\textit{merge} -- in-domain}
    \end{subfigure}
    ~
    \begin{subfigure}[b]{0.31\textwidth}
        \centering
        \includegraphics[width=\textwidth]{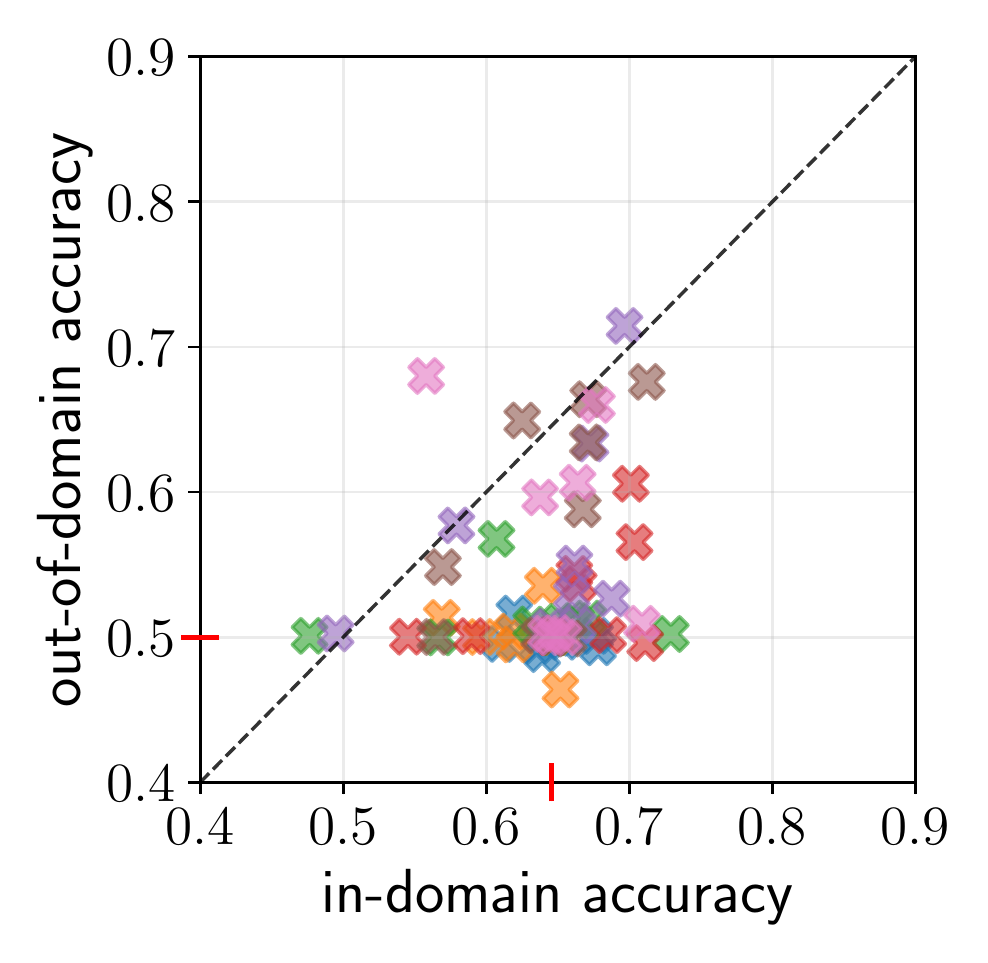}
        \caption{\textit{merge} -- last}
    \end{subfigure}
    ~
    \begin{subfigure}[b]{0.31\textwidth}
        \centering
        \includegraphics[width=\textwidth]{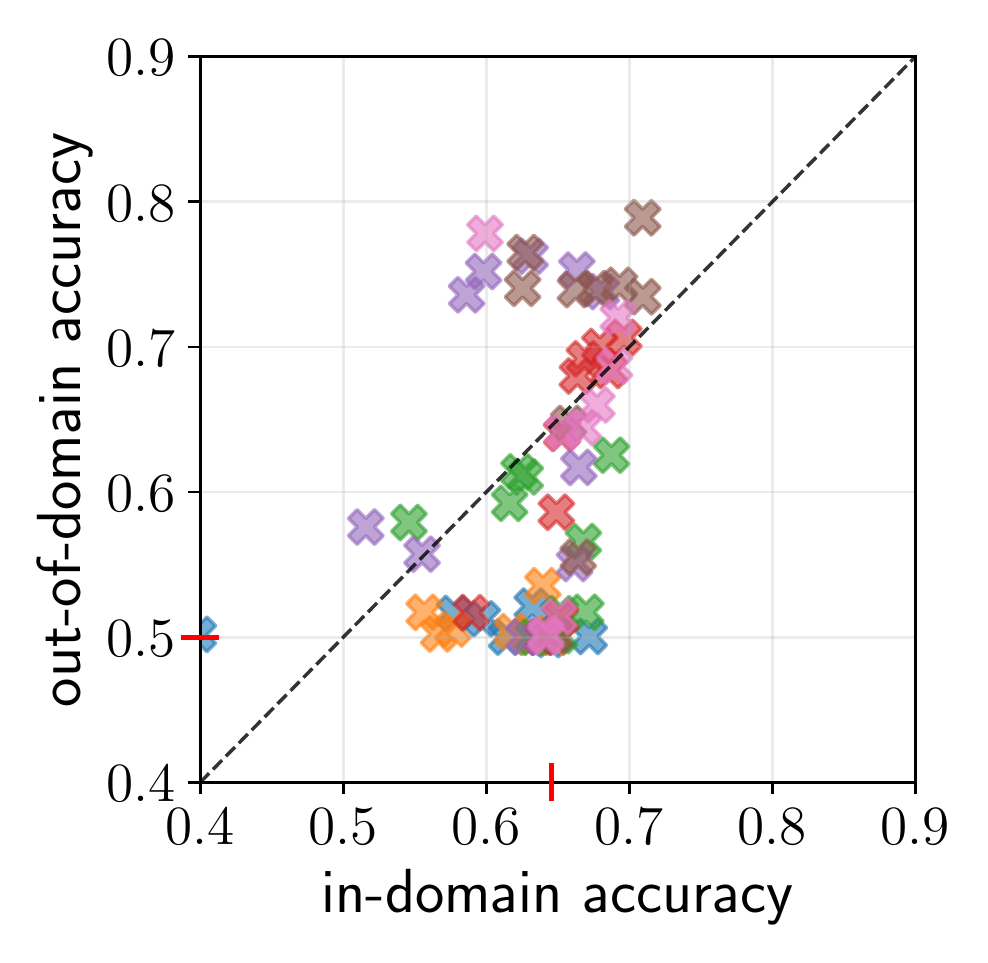}
        \caption{\textit{merge} -- out-of-domain}
    \end{subfigure}
    \\
    \begin{subfigure}[b]{0.31\textwidth}
        \centering
        \includegraphics[width=\textwidth]{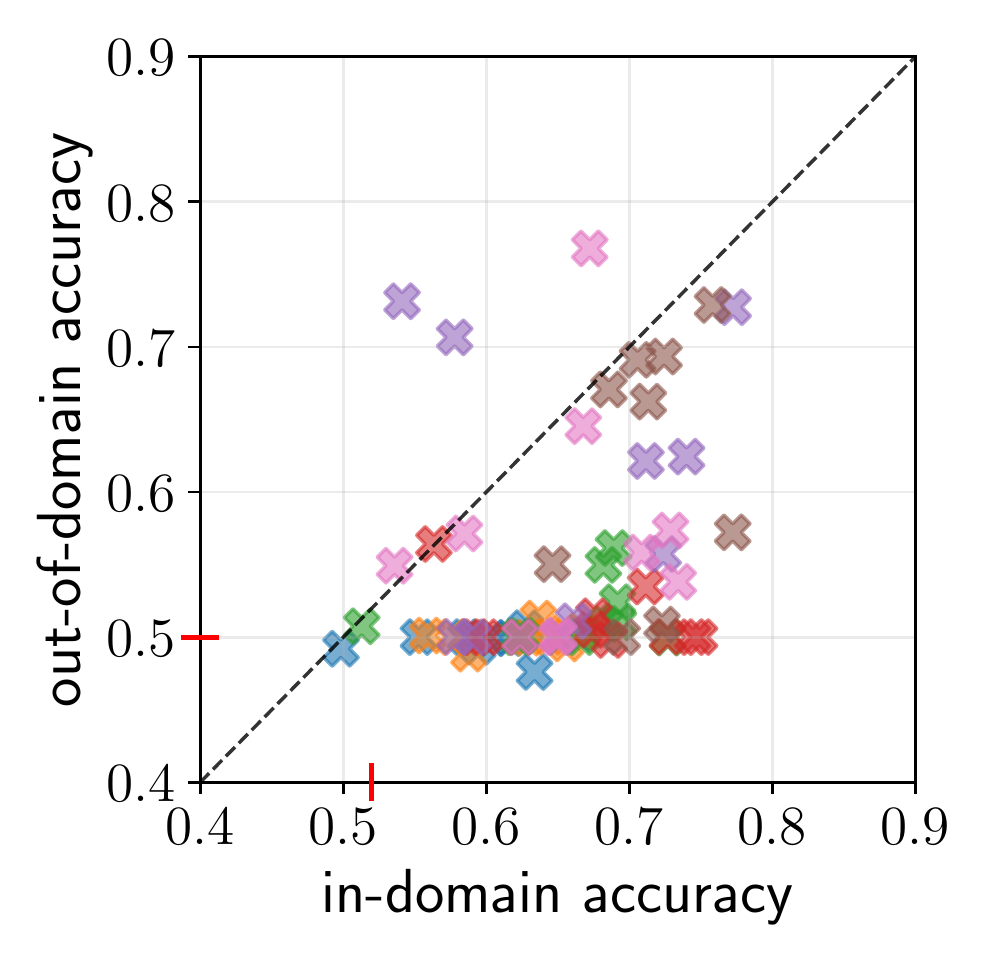}
        \caption{\textit{remove} -- in-domain}
    \end{subfigure}
    ~
    \begin{subfigure}[b]{0.31\textwidth}
        \centering
        
        \includegraphics[width=\textwidth]{figures/ft/all-models_mnli_16_last_pattern-verbalizer-ft}
        \caption{\textit{remove} -- last}
    \end{subfigure}
    ~
    \begin{subfigure}[b]{0.31\textwidth}
        \centering
        \includegraphics[width=\textwidth]{figures/ft/all-models_mnli_16_best_out-of-domain_pattern-verbalizer-ft}
        \caption{\textit{remove} -- out-of-domain}
    \end{subfigure}
    
    \caption{\textbf{Relationship between in-domain and out-of-domain performance of PBFT on MNLI} for OPT models of various sizes when \textbf{merging} the neutral and contradiction classes \textbf{vs. removing} the neutral examples altogether.
    We fine-tune on \textbf{16 examples} using 10 different seeds.
    \textcolor{red}{$\boldsymbol{-}$} in the x- and y-axis indicates the performance of the majority class label.
    }
    \label{fig:appendix-ft-mnli-original-16}
\end{figure*}

\begin{figure*}[h]
    \centering
    \begin{subfigure}[b]{0.31\textwidth}
        \centering
        \includegraphics[width=\textwidth]{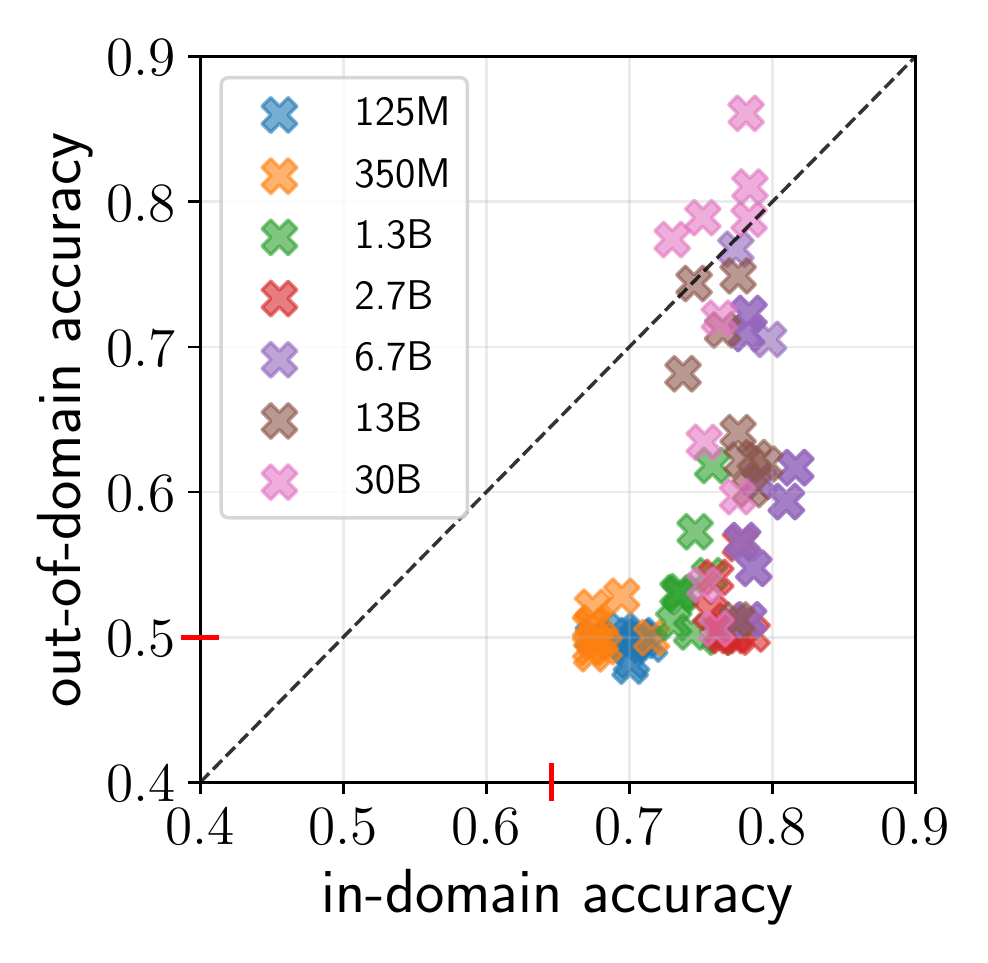}
        \caption{\textit{merge} -- in-domain}
    \end{subfigure}
    ~
    \begin{subfigure}[b]{0.31\textwidth}
        \centering
        \includegraphics[width=\textwidth]{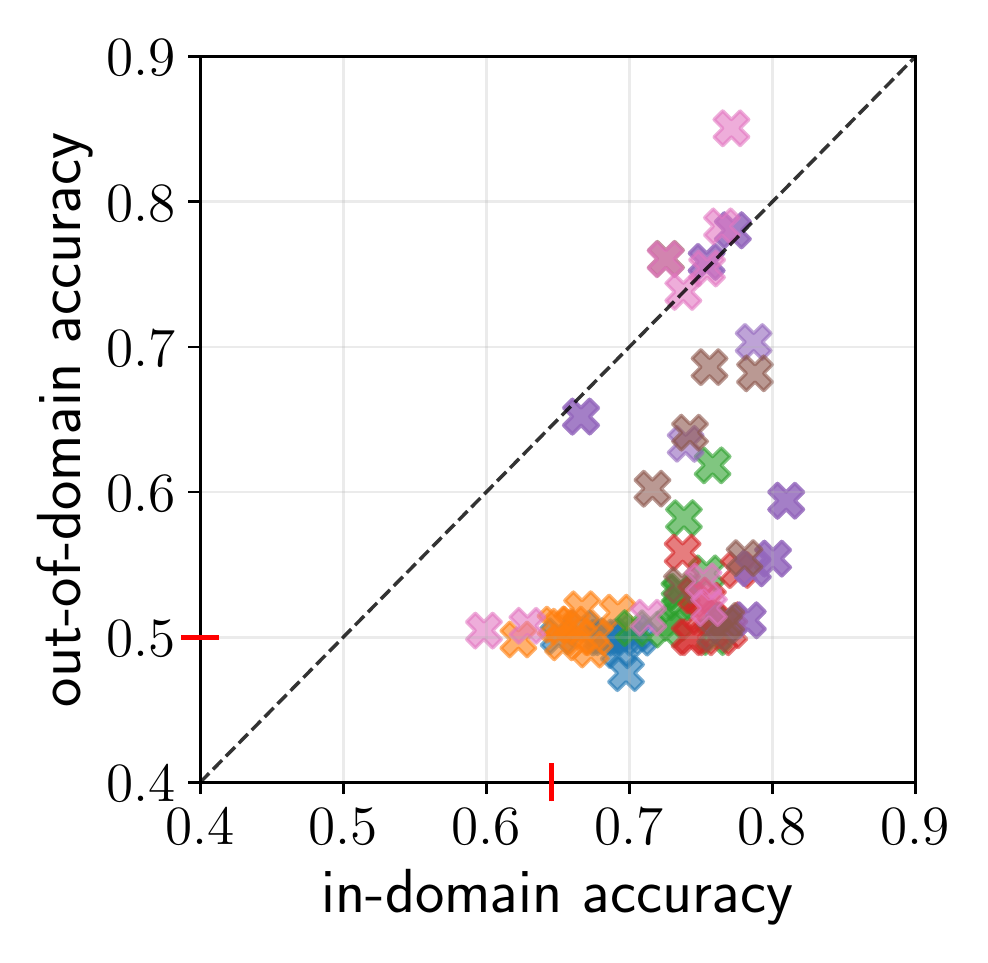}
        \caption{\textit{merge} -- last}
    \end{subfigure}
    ~
    \begin{subfigure}[b]{0.31\textwidth}
        \centering
        \includegraphics[width=\textwidth]{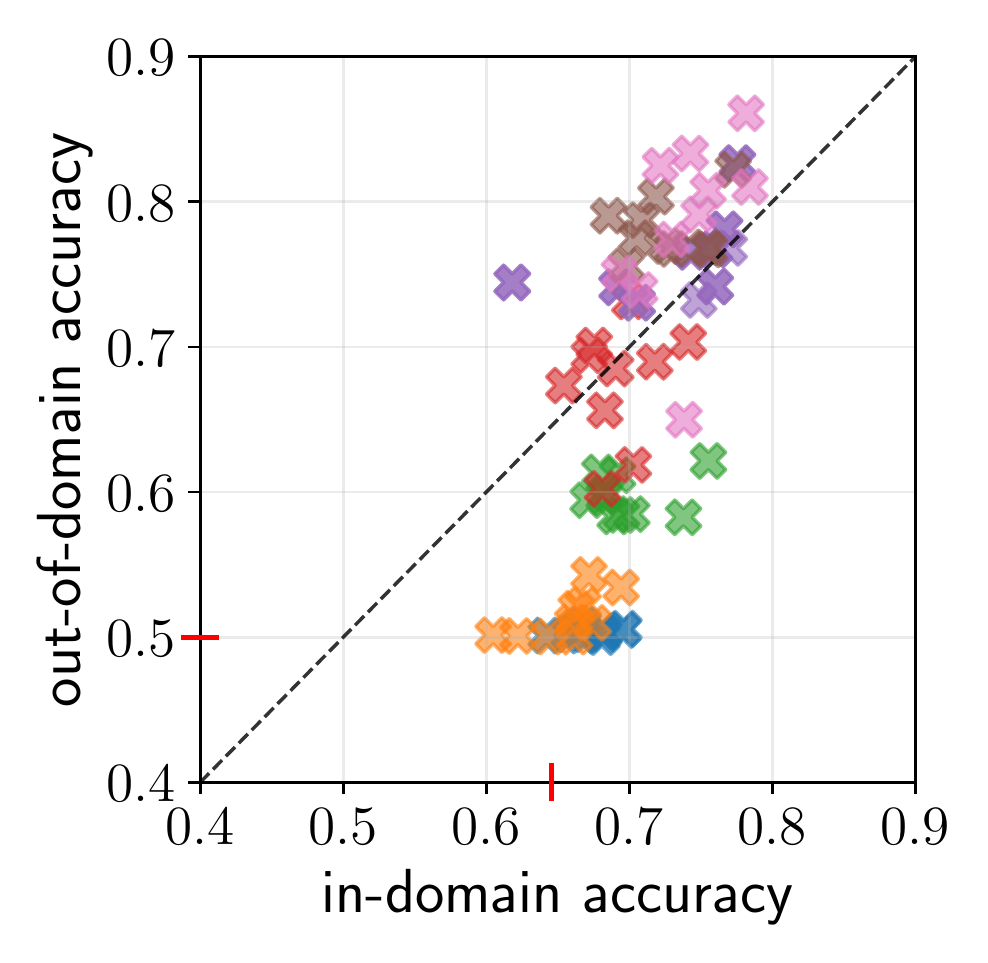}
        \caption{\textit{merge} -- out-of-domain}
    \end{subfigure}
    \\
    \begin{subfigure}[b]{0.31\textwidth}
        \centering
        \includegraphics[width=\textwidth]{figures/ft/all-models_mnli_128_best_in-domain_pattern-verbalizer-ft}
        \caption{\textit{remove} -- in-domain}
    \end{subfigure}
    ~
    \begin{subfigure}[b]{0.31\textwidth}
        \centering
        
        \includegraphics[width=\textwidth]{figures/ft/all-models_mnli_128_last_pattern-verbalizer-ft}
        \caption{\textit{remove} -- last}
    \end{subfigure}
    ~
    \begin{subfigure}[b]{0.31\textwidth}
        \centering
        \includegraphics[width=\textwidth]{figures/ft/all-models_mnli_128_best_out-of-domain_pattern-verbalizer-ft}
        \caption{\textit{remove} -- out-of-domain}
    \end{subfigure}
    
    \caption{\textbf{Relationship between in-domain and out-of-domain performance of PBFT on MNLI} for OPT models of various sizes when \textbf{merging} the neutral and contradiction classes \textbf{vs. removing} the neutral examples altogether.
    We fine-tune on \textbf{128 examples} using 10 different seeds.
    \textcolor{red}{$\boldsymbol{-}$} in the x- and y-axis indicates the performance of the majority class label.
    }
    \label{fig:appendix-ft-mnli-original}
\end{figure*}

\Cref{fig:appendix-ft-mnli-original-16,fig:appendix-ft-mnli-original} show a comparison between two different ways of binarizing MNLI. For our main experiments, we remove the neutral class entirely. Merging it with the contradiction class instead leads to an even better relationship between in-domain and OOD performance.


\section{Additional results for Pythia models}
\label{sec:appendix-pythia}

\begin{figure*}[h]
    \centering
    \begin{subfigure}[b]{0.31\textwidth}
        \centering
        \includegraphics[width=\textwidth]{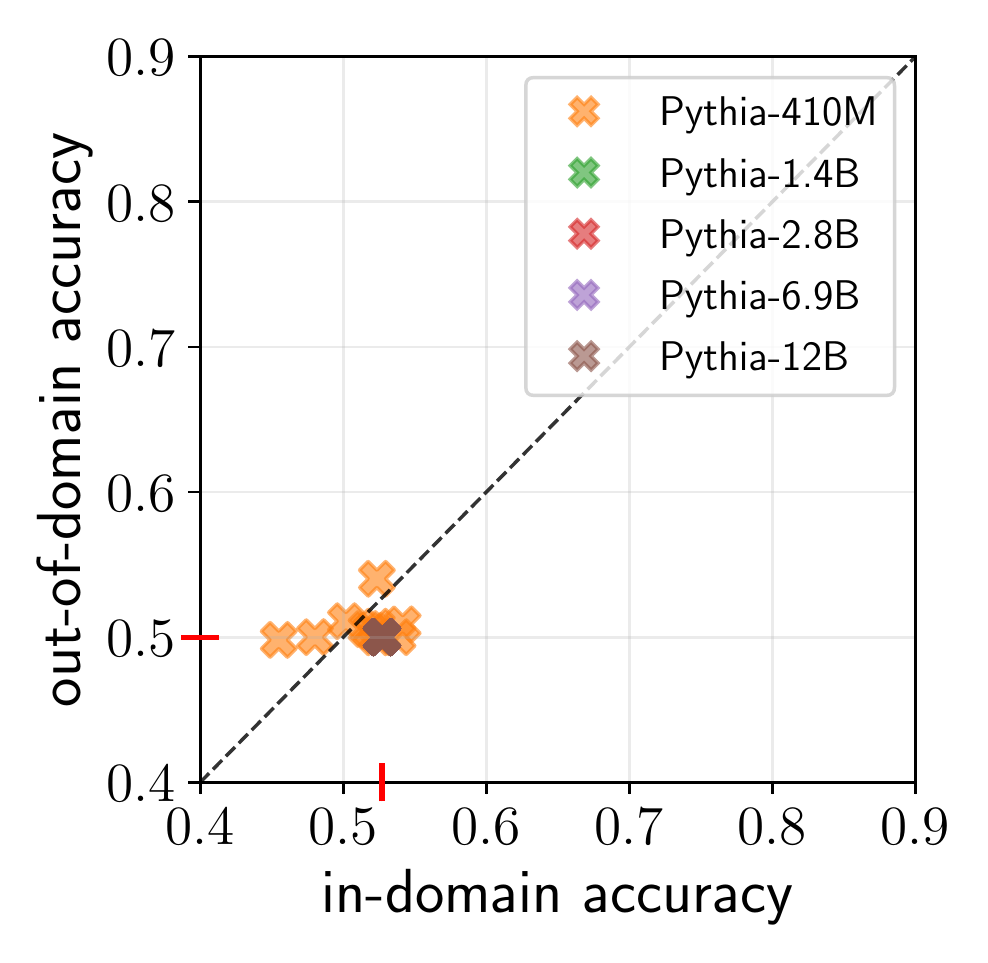}
        \caption{ICL 16 samples -- \texttt{minimal}}
    \end{subfigure}
    ~
    \begin{subfigure}[b]{0.31\textwidth}
        \centering
        \includegraphics[width=\textwidth]{figures/in-context/pythia_16-shots_rte_gpt-3.pdf}
        \caption{ICL 16 samples -- \texttt{gpt-3}}
    \end{subfigure}
    ~
    \begin{subfigure}[b]{0.31\textwidth}
        \centering
        \includegraphics[width=\textwidth]{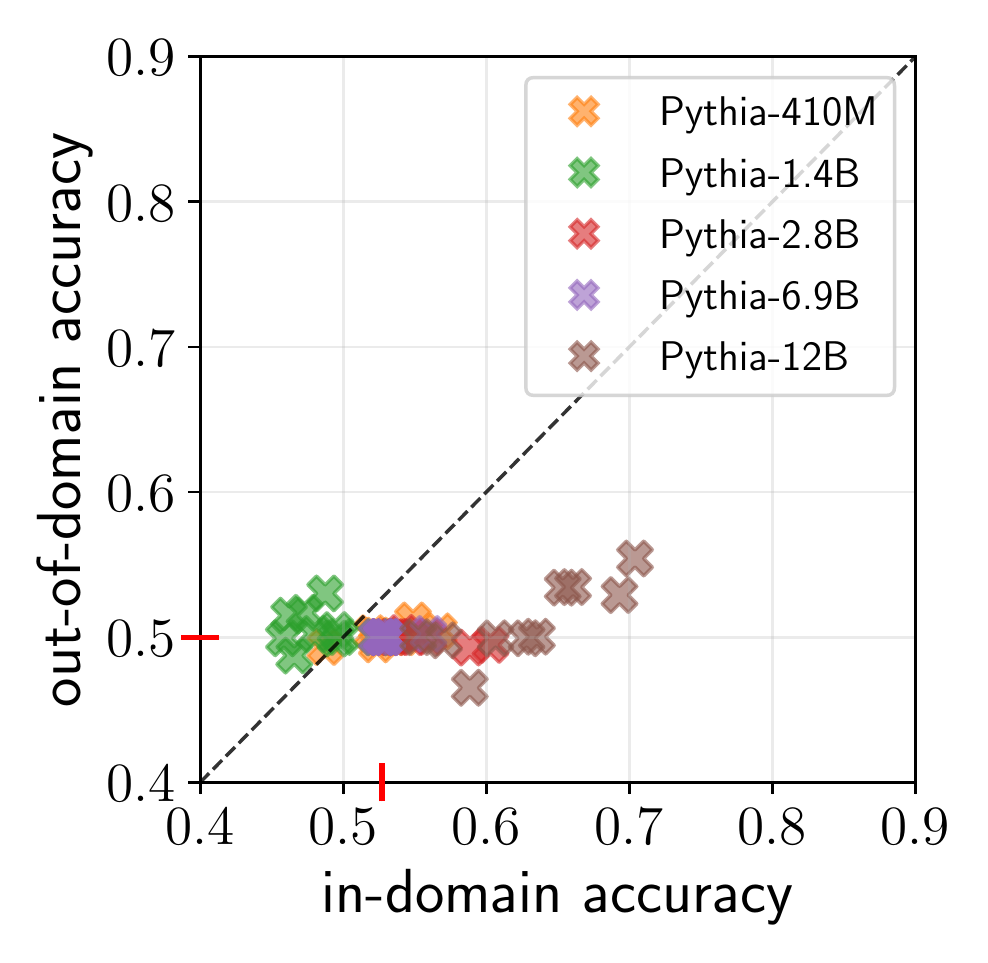}
        \caption{ICL 16 samples -- \texttt{eval-harness}}
    \end{subfigure}
    \\
    \begin{subfigure}[b]{0.31\textwidth}
        \centering
        \includegraphics[width=\textwidth]{figures/ft/pythia_rte_16_best_in-domain_pattern-verbalizer-ft.pdf}
        \caption{FT 16 samples -- best in-domain}
    \end{subfigure}
    ~
    \begin{subfigure}[b]{0.31\textwidth}
        \centering
        
        \includegraphics[width=\textwidth]{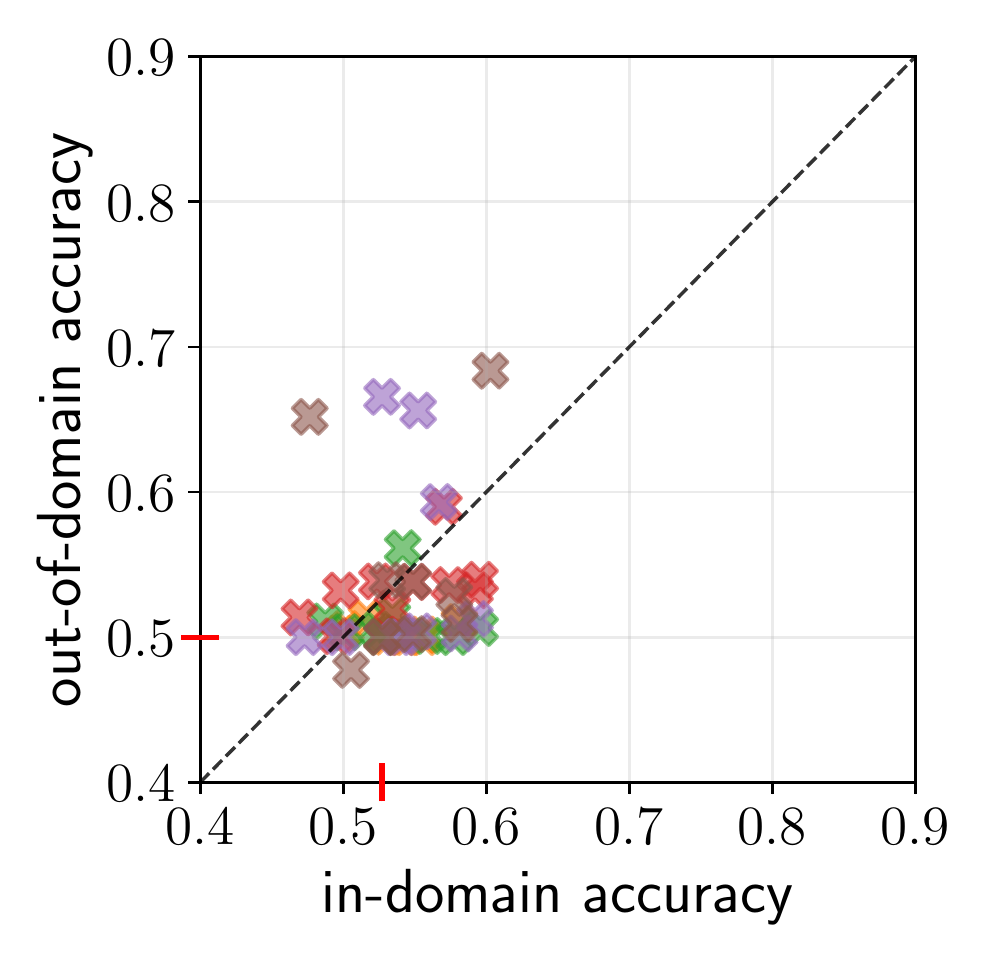}
        \caption{FT 16 samples -- last}
    \end{subfigure}
    ~
    \begin{subfigure}[b]{0.31\textwidth}
        \centering
        \includegraphics[width=\textwidth]{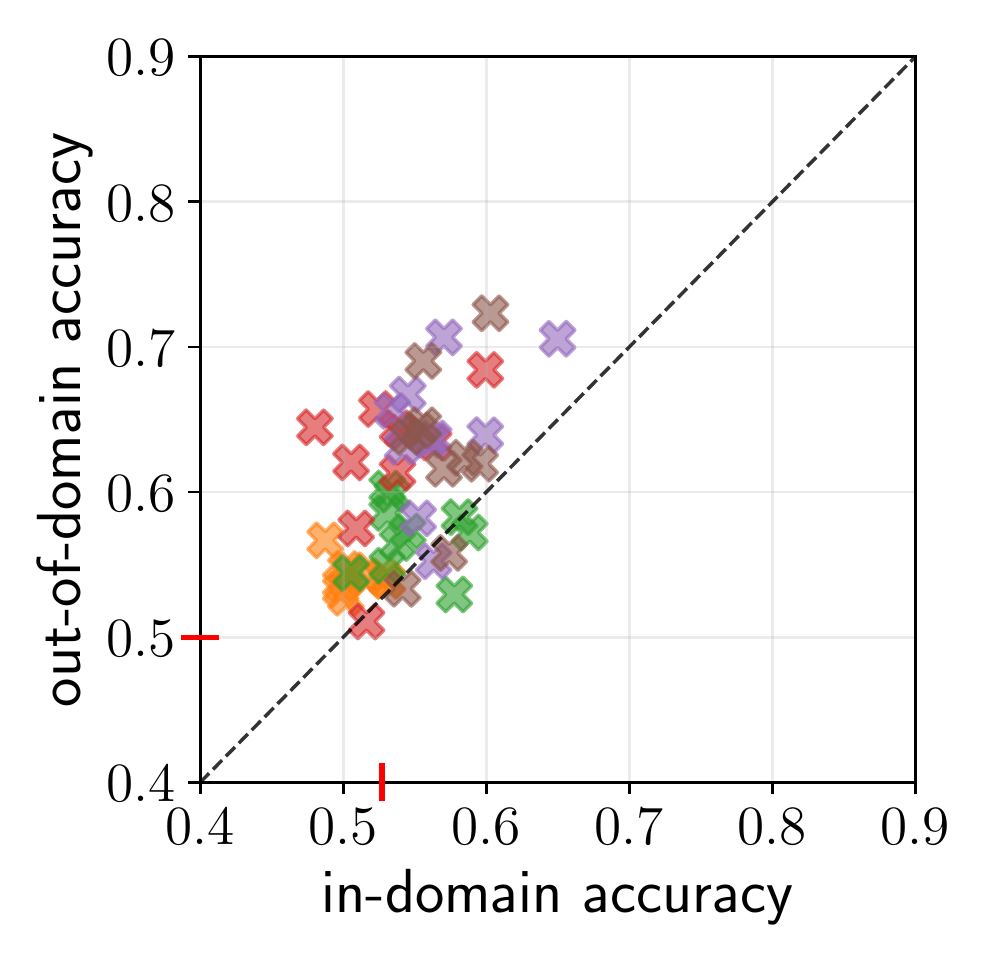}
        \caption{FT 16 samples -- best out-of-domain}
    \end{subfigure}
    \\
    \begin{subfigure}[b]{0.31\textwidth}
        \centering
        \includegraphics[width=\textwidth]{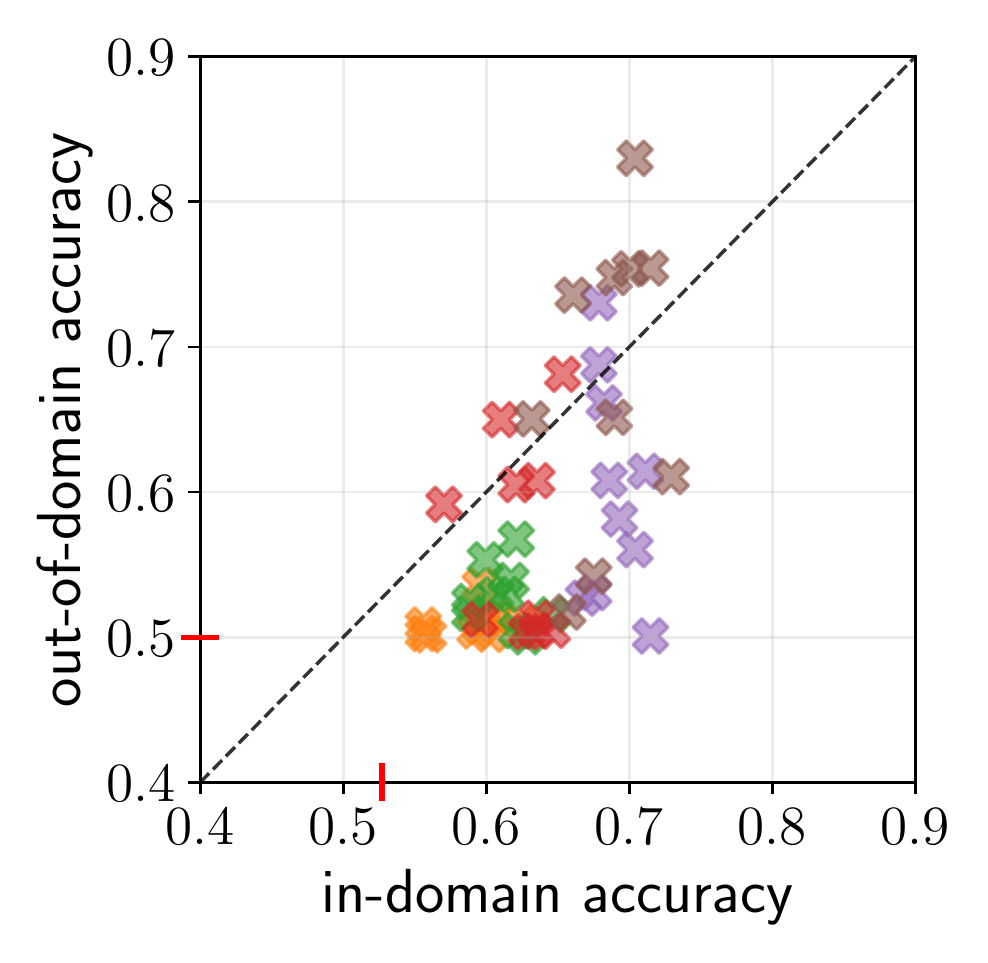}
        \caption{FT 128 samples -- best in-domain}
    \end{subfigure}
    ~
    \begin{subfigure}[b]{0.31\textwidth}
        \centering
        
        \includegraphics[width=\textwidth]{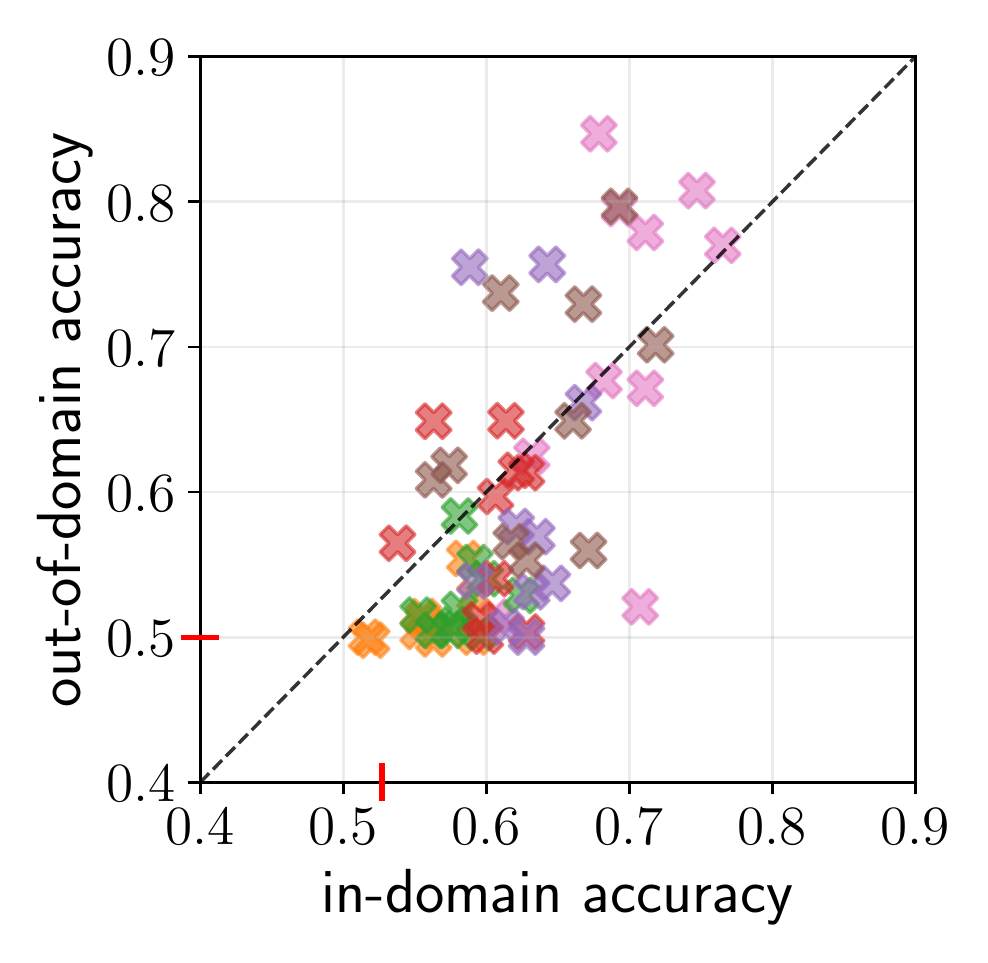}
        \caption{FT 128 samples -- last}
    \end{subfigure}
    ~
    \begin{subfigure}[b]{0.31\textwidth}
        \centering
        \includegraphics[width=\textwidth]{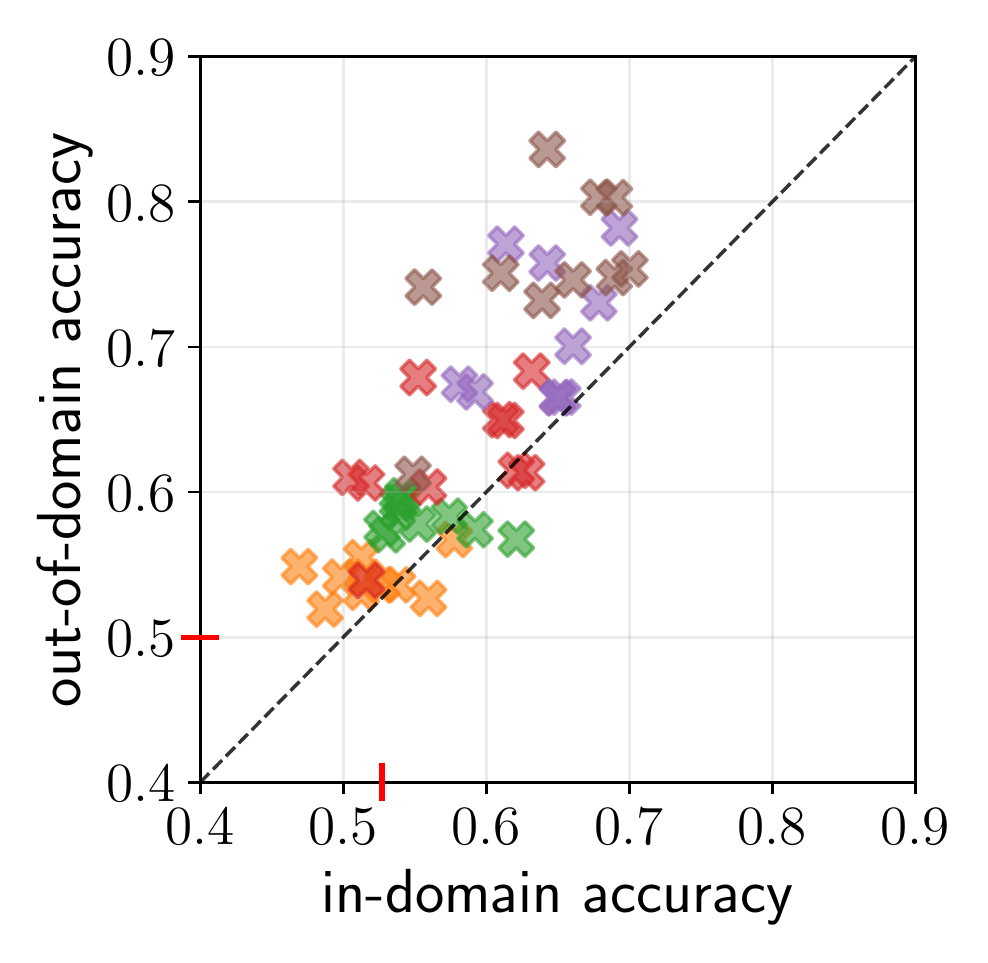}
        \caption{FT 128 samples -- best out-of-domain}
    \end{subfigure}
    
    \caption{\textbf{ICL and FT results for Pythia models of different size}. For ICL, we report results using 16 examples and three different patterns (\texttt{minimal}, \texttt{gpt-3}, \texttt{eval-harness}). For FT, we report results using 16 and 128 examples using three different model selection strategies (best in-domain, last checkpoint, best out-of-domain). In all cases, we show results for 10 different random seeds.
    \textcolor{red}{$\boldsymbol{-}$} in the x- and y-axis indicates the performance of the majority class label.
    }
    \label{fig:appendix-pythia}
\end{figure*}

\cref{fig:appendix-pythia} compares FT and ICL of 
Pythia models ranging from 410M to 12B parameters \citep{biderman2023pythia}. Similar to OPT, the Pythia models differ only in their size and have all been trained on exactly the same data (even in the exact same order). We focus on RTE and report results using 16 examples. For ICL, we use three different patterns (\texttt{minimal}, \texttt{gpt-3}, \texttt{eval-harness}). For FT, we report results using 16 and 128 examples and three different model selection strategies (best in-domain, last checkpoint, best out-of-domain). Significance tests are provided in \Cref{tab:appendix-statistical-tests-pythia-ood-ood,tab:appendix-statistical-tests-pythia-in-domain-in-domain,tab:appendix-statistical-tests-pythia-in-domain-ood,tab:appendix-statistical-tests-pythia-ood-in-domain}.

For ICL, all models perform poorly when using the \texttt{minimal} pattern. With the \texttt{gpt-3} pattern, we can observe a clear impact of model size on in-domain and out-of-domain performance. On the other hand, with the \texttt{eval-harness} pattern, for Pythia models, only in-domain performance improves with model size. 

For FT, when using 16 samples and selecting checkpoints according to out-of-domain performance, almost all checkpoints lead to better out-of-domain than in-domain performance. Moreover, almost all fine-tuned models perform significantly better OOD than models adapted via ICL. When fine-tuning with 128 examples, we can see a very clear effect of model size on both in-domain and out-of-domain performance. In particular, when selecting checkpoints according to out-of-domain performance, almost all models perform better out-of-domain than in-domain.

\begin{table*}[hp]
    \centering
    \resizebox{0.55\textwidth}{!}{%
    \begin{tabular}{llccccc}
        \toprule
        & 
        & \multicolumn{5}{c}{\textbf{FT}} \\ \cmidrule{3-7} 
        & & \textbf{410M} & \textbf{1.4B} & \textbf{2.8B} & \textbf{6.9B} & \textbf{12B} 
        \\ \midrule
        \multirow{5}{*}{\rotatebox[origin=c]{90}{\textbf{ICL}}} & \textbf{410M}  & \cellcolor{blue!35}$0.05$ & \cellcolor{blue!35}$0.06$ & \cellcolor{blue!35}$0.06$ & \cellcolor{blue!35}$0.09$ & \cellcolor{blue!35}$0.07$ \\
         & \textbf{1.4B}  & \cellcolor{blue!35}$0.03$ & \cellcolor{blue!35}$0.04$ & \cellcolor{blue!35}$0.04$ & \cellcolor{blue!35}$0.07$ & \cellcolor{blue!35}$0.05$ \\
         & \textbf{2.8B} & $-0.02$ & $-0.00$ & $-0.01$ & $0.02$ & $0.01$ \\
          & \textbf{6.9B} & \cellcolor{red!15}$-0.03$ & \cellcolor{red!15}$-0.02$ & \cellcolor{red!15}$-0.02$ & $0.01$ & $-0.01$  \\
         & \textbf{12B} & \cellcolor{red!15}$-0.04$ & \cellcolor{red!15}$-0.03$ & \cellcolor{red!15}$-0.03$ & $-0.00$ & $-0.02$ \\
        \bottomrule
    \end{tabular}%
    }
    \caption{Difference between average \textbf{in-domain performance} of ICL and FT with Pythia models on RTE. We use 16 examples and 10 random seeds for both approaches. For ICL, we use the \texttt{gpt-3} pattern. For FT, we use pattern-based fine-tuning (PBFT) and select checkpoints according to \underline{in-domain performance}. We perform a Welch's t-test and color cells according to whether: \textcolor{red!45}{ICL performs significantly better than FT}, \textcolor{blue!75}{FT performs significantly better than ICL}. For cells without color, there is no significant difference between ICL and FT.\looseness-1
    } \label{tab:appendix-statistical-tests-pythia-in-domain-in-domain}
\end{table*}
\begin{table*}[hp]
    \centering
    \resizebox{0.55\textwidth}{!}{%
    \begin{tabular}{llccccc}
        \toprule
        & 
        & \multicolumn{5}{c}{\textbf{FT}} \\ \cmidrule{3-7} 
        & & \textbf{410M} & \textbf{1.4B} & \textbf{2.8B} & \textbf{6.9B} & \textbf{12B} 
        \\ \midrule
        \multirow{5}{*}{\rotatebox[origin=c]{90}{\textbf{ICL}}} & \textbf{410M}  & $-0.00$ & \cellcolor{blue!35}$0.04$ & $0.02$ & \cellcolor{blue!35}$0.06$ & \cellcolor{blue!35}$0.06$ \\
         & \textbf{1.4B}  & $-0.02$ & $0.02$ & $0.00$ & \cellcolor{blue!35}$0.04$ & \cellcolor{blue!35}$0.04$  \\
         & \textbf{2.8B} & \cellcolor{red!15}$-0.06$ & \cellcolor{red!15}$-0.03$ & \cellcolor{red!15}$-0.04$ & $-0.01$ & $-0.01$ \\
          & \textbf{6.9B} & \cellcolor{red!15}$-0.08$ & \cellcolor{red!15}$-0.04$ & \cellcolor{red!15}$-0.06$ & \cellcolor{red!15}$-0.02$ & \cellcolor{red!15}$-0.02$ \\
         & \textbf{12B} & \cellcolor{red!15}$-0.09$ & \cellcolor{red!15}$-0.05$ & \cellcolor{red!15}$-0.07$ & \cellcolor{red!15}$-0.03$ & \cellcolor{red!15}$-0.03$ \\
        \bottomrule
    \end{tabular}%
    }
    \caption{Difference between average \textbf{in-domain performance} of ICL and FT with Pythia models on RTE. We use 16 examples and 10 random seeds for both approaches. For ICL, we use the \texttt{gpt-3} pattern. For FT, we use pattern-based fine-tuning (PBFT) and select checkpoints according to \underline{out-of-domain performance}. We perform a Welch's t-test and color cells according to whether: \textcolor{red!45}{ICL performs significantly better than FT}, \textcolor{blue!75}{FT performs significantly better than ICL}. For cells without color, there is no significant difference between ICL and FT.\looseness-1
    } \label{tab:appendix-statistical-tests-pythia-in-domain-ood}
\end{table*}
\begin{table*}[hp]
    \centering
    \resizebox{0.55\textwidth}{!}{%
        \begin{tabular}{llccccc}
        \toprule
        & 
        & \multicolumn{5}{c}{\textbf{FT}} \\ \cmidrule{3-7} 
        & & \textbf{410M} & \textbf{1.4B} & \textbf{2.8B} & \textbf{6.9B} & \textbf{12B} 
        \\ \midrule
        \multirow{5}{*}{\rotatebox[origin=c]{90}{\textbf{ICL}}} & \textbf{410M}  & \cellcolor{blue!35}$0.05$ & \cellcolor{blue!35}$0.08$ & \cellcolor{blue!35}$0.13$ & \cellcolor{blue!35}$0.15$ & \cellcolor{blue!35}$0.14$ \\
         & \textbf{1.4B}  & \cellcolor{blue!35}$0.04$ & \cellcolor{blue!35}$0.07$ & \cellcolor{blue!35}$0.12$ & \cellcolor{blue!35}$0.14$ & \cellcolor{blue!35}$0.13$ \\
         & \textbf{2.8B} & $-0.00$ & \cellcolor{blue!35}$0.03$ & \cellcolor{blue!35}$0.08$ & \cellcolor{blue!35}$0.10$ & \cellcolor{blue!35}$0.09$ \\
          & \textbf{6.9B} & \cellcolor{blue!35}$0.04$ & \cellcolor{blue!35}$0.07$ & \cellcolor{blue!35}$0.12$ & \cellcolor{blue!35}$0.14$ & \cellcolor{blue!35}$0.13$ \\
         & \textbf{12B} & $0.00$ & \cellcolor{blue!35}$0.03$ & \cellcolor{blue!35}$0.08$ & \cellcolor{blue!35}$0.10$ & \cellcolor{blue!35}$0.09$ \\
        \bottomrule
    \end{tabular}%
    }
    \caption{Difference between average \textbf{out-of-domain performance} of ICL and FT with Pythia models on RTE. We use 16 examples and 10 random seeds for both approaches. For ICL, we use the \texttt{gpt-3} pattern. For FT, we use pattern-based fine-tuning (PBFT) and select checkpoints according to \underline{out-of-domain performance}. We perform a Welch's t-test and color cells according to whether: \textcolor{red!45}{ICL performs significantly better than FT}, \textcolor{blue!75}{FT performs significantly better than ICL}. For cells without color, there is no significant difference between ICL and FT.\looseness-1
    } \label{tab:appendix-statistical-tests-pythia-ood-in-domain}
\end{table*}
\section{Analyzing individual OPT fine-tuning runs}
\label{sec:appendix-individual-ft-runs}

\begin{figure*}[h]
    \centering
    \begin{subfigure}[b]{0.35\textwidth}
        \centering
        \includegraphics[width=\textwidth]{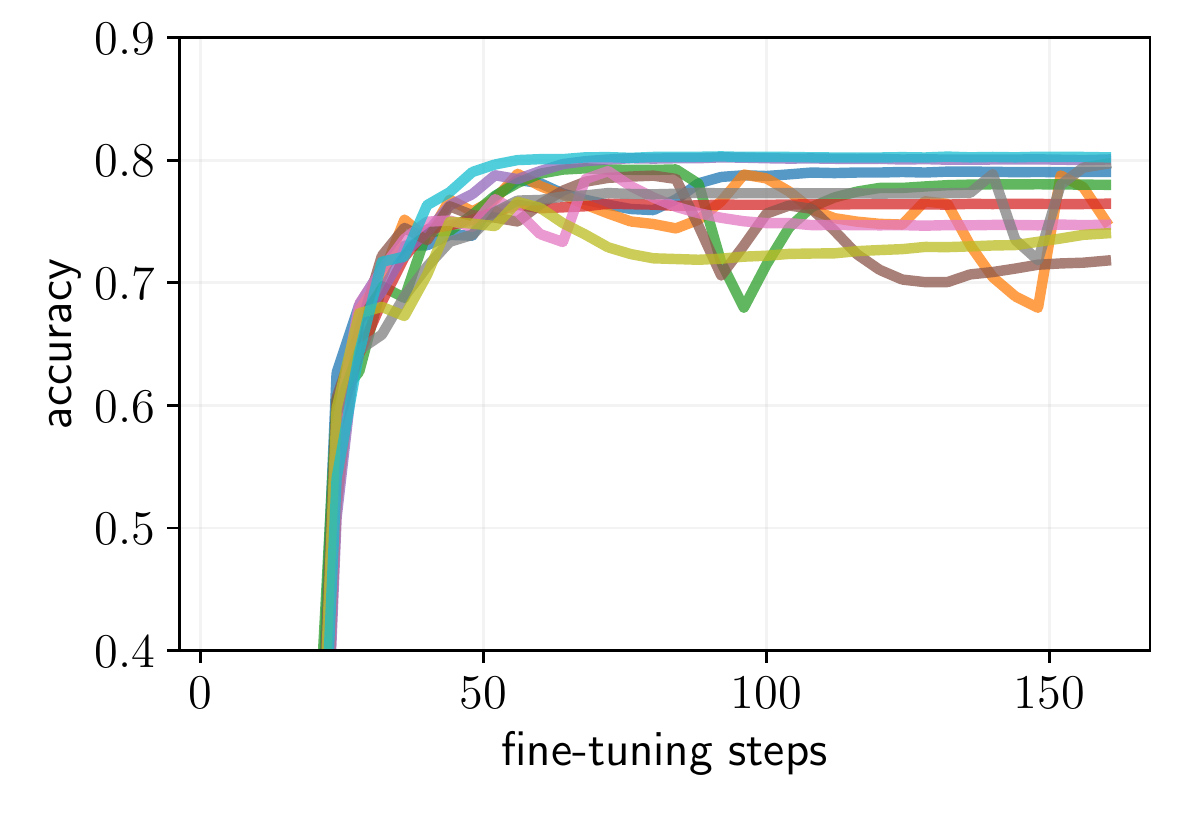}
        \caption{1.3B -- in-domain}
    \end{subfigure}
    ~
    \begin{subfigure}[b]{0.35\textwidth}
        \centering
        \includegraphics[width=\textwidth]{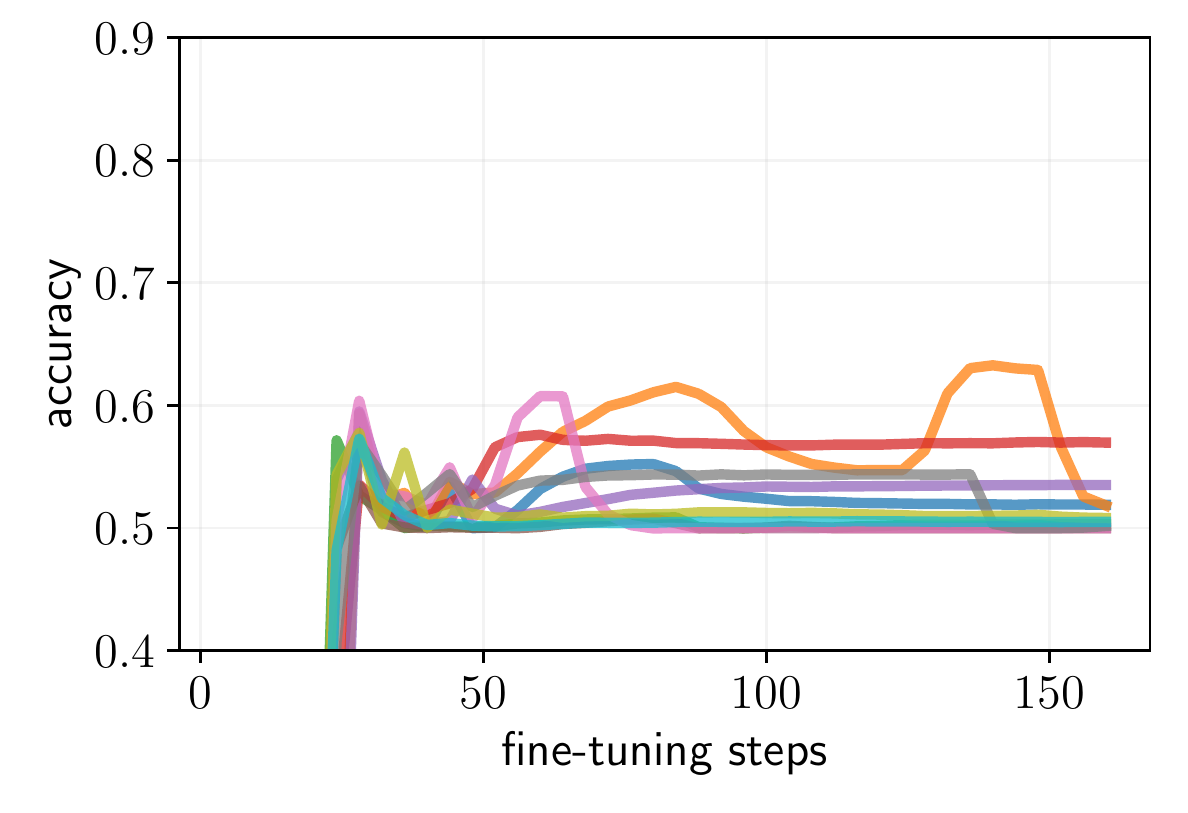}
        \caption{1.3B -- out-of-domain}
    \end{subfigure}
    \\
    \begin{subfigure}[b]{0.35\textwidth}
        \centering
        \includegraphics[width=\textwidth]{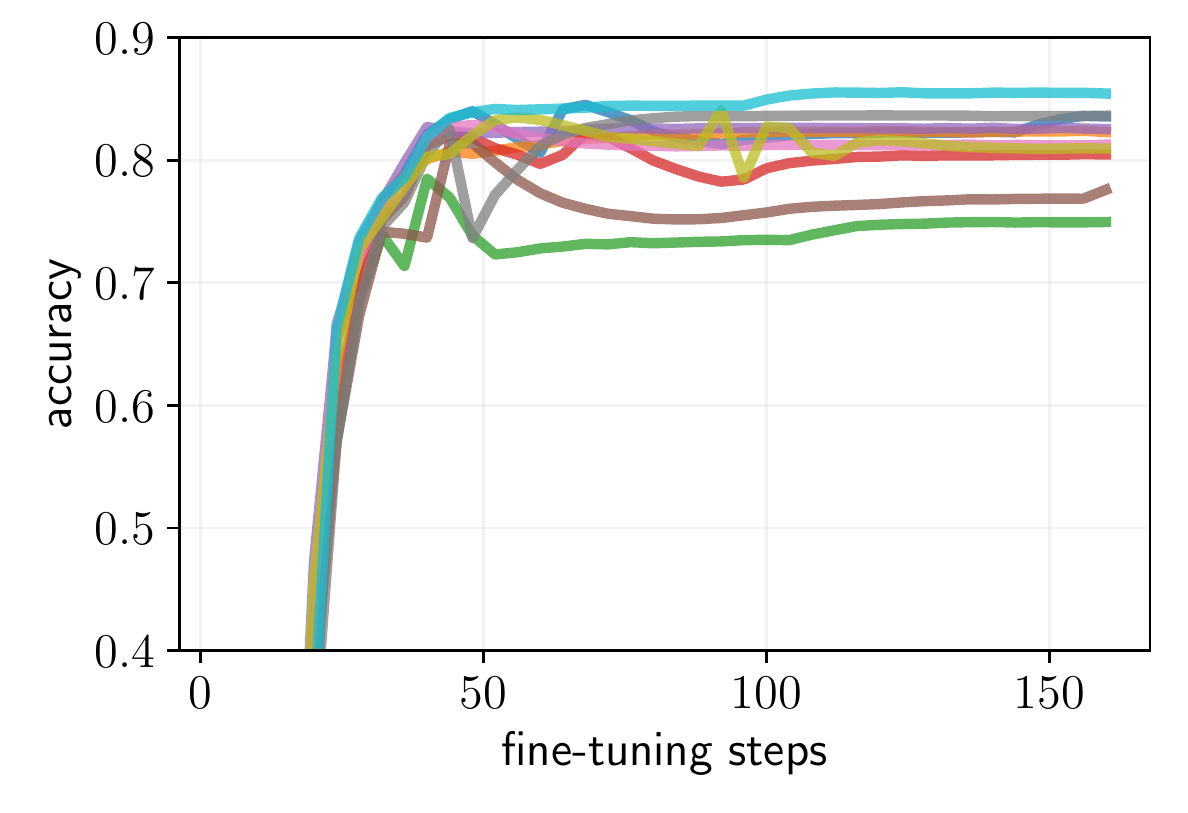}
        \caption{2.7B -- in-domain}
    \end{subfigure}
    ~
    \begin{subfigure}[b]{0.35\textwidth}
        \centering
        \includegraphics[width=\textwidth]{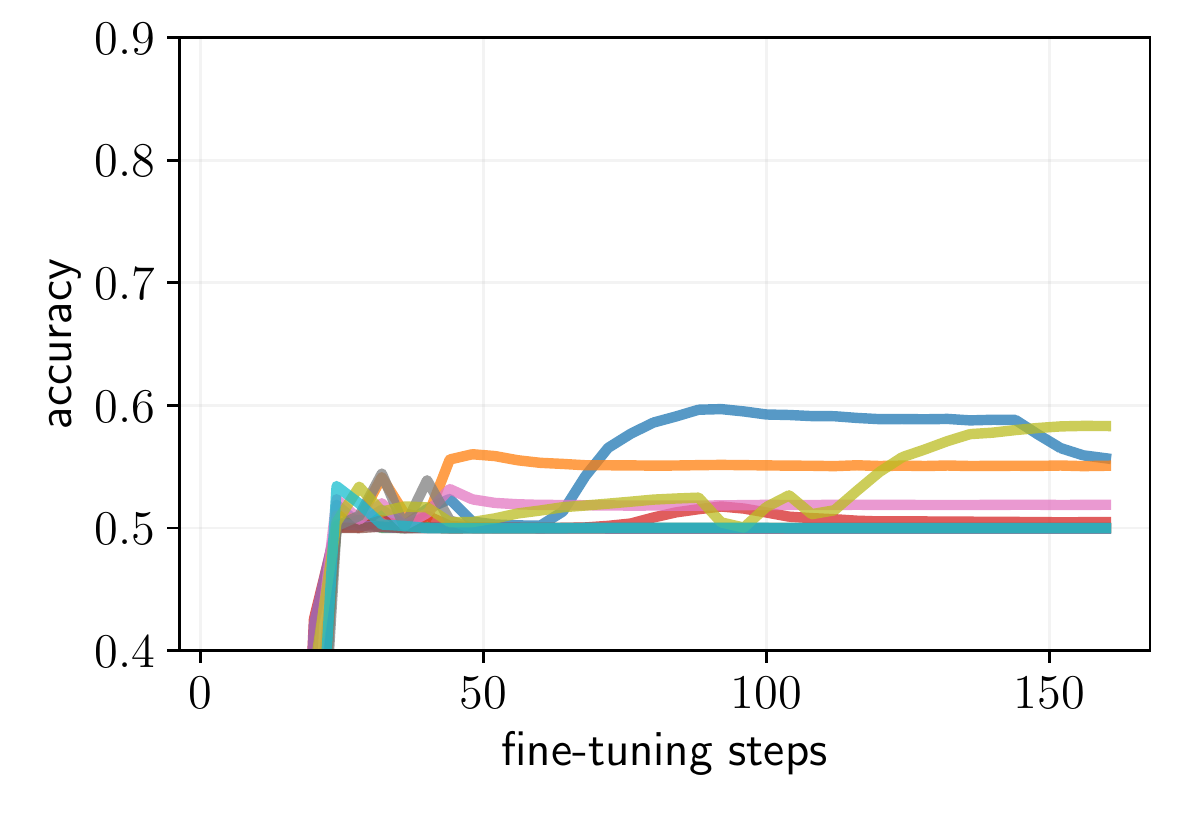}
        \caption{2.7B -- out-of-domain}
    \end{subfigure}
    \\
    \begin{subfigure}[b]{0.35\textwidth}
        \centering
        \includegraphics[width=\textwidth]{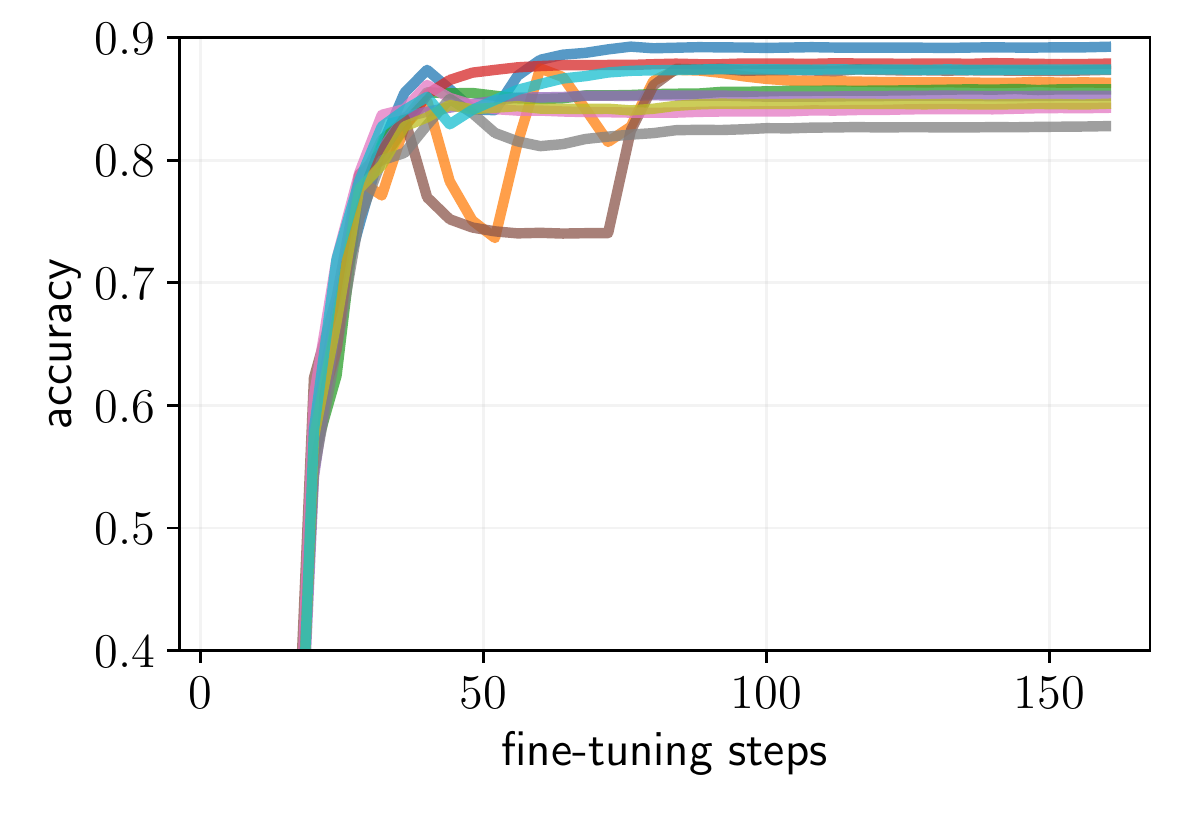}
        \caption{6.7B -- in-domain}
    \end{subfigure}
    ~
    \begin{subfigure}[b]{0.35\textwidth}
        \centering
        \includegraphics[width=\textwidth]{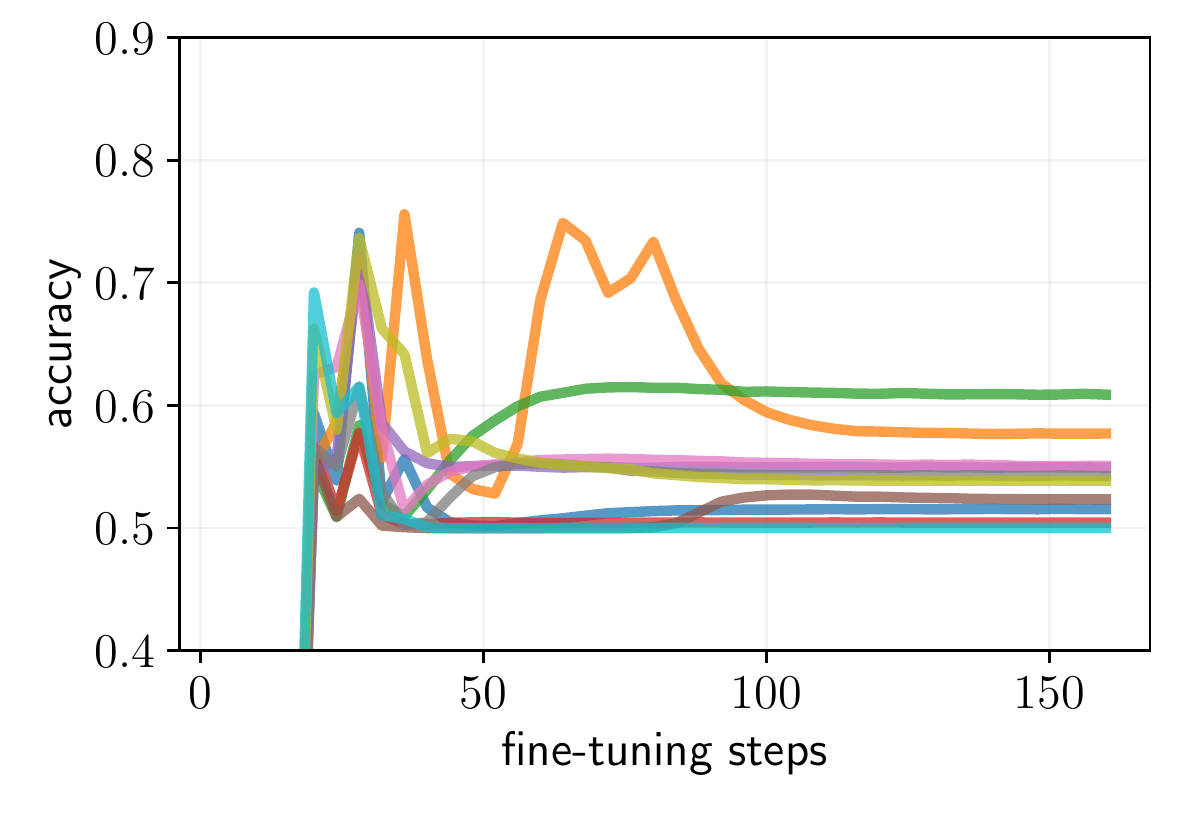}
        \caption{6.7B -- out-of-domain}
    \end{subfigure}
    \\
    \begin{subfigure}[b]{0.35\textwidth}
        \centering
        \includegraphics[width=\textwidth]{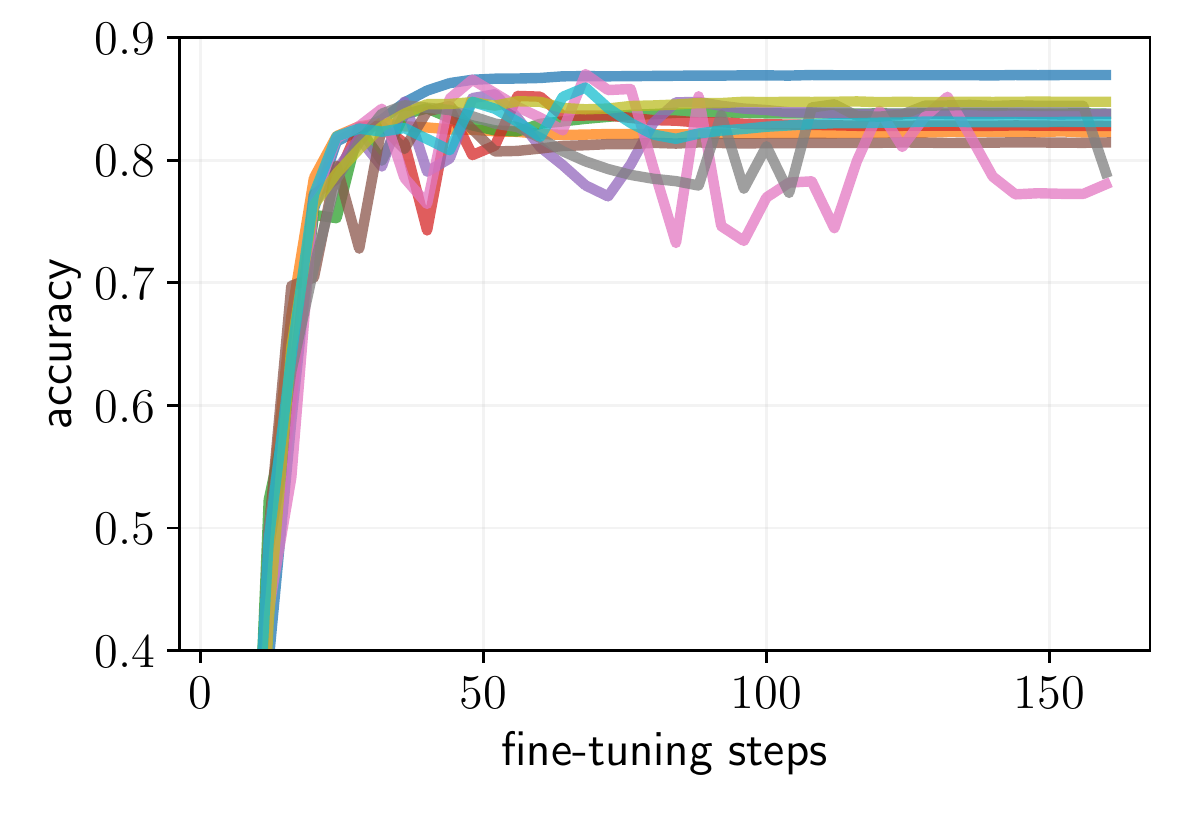}
        \caption{13B -- in-domain}
    \end{subfigure}
    ~
    \begin{subfigure}[b]{0.35\textwidth}
        \centering
        \includegraphics[width=\textwidth]{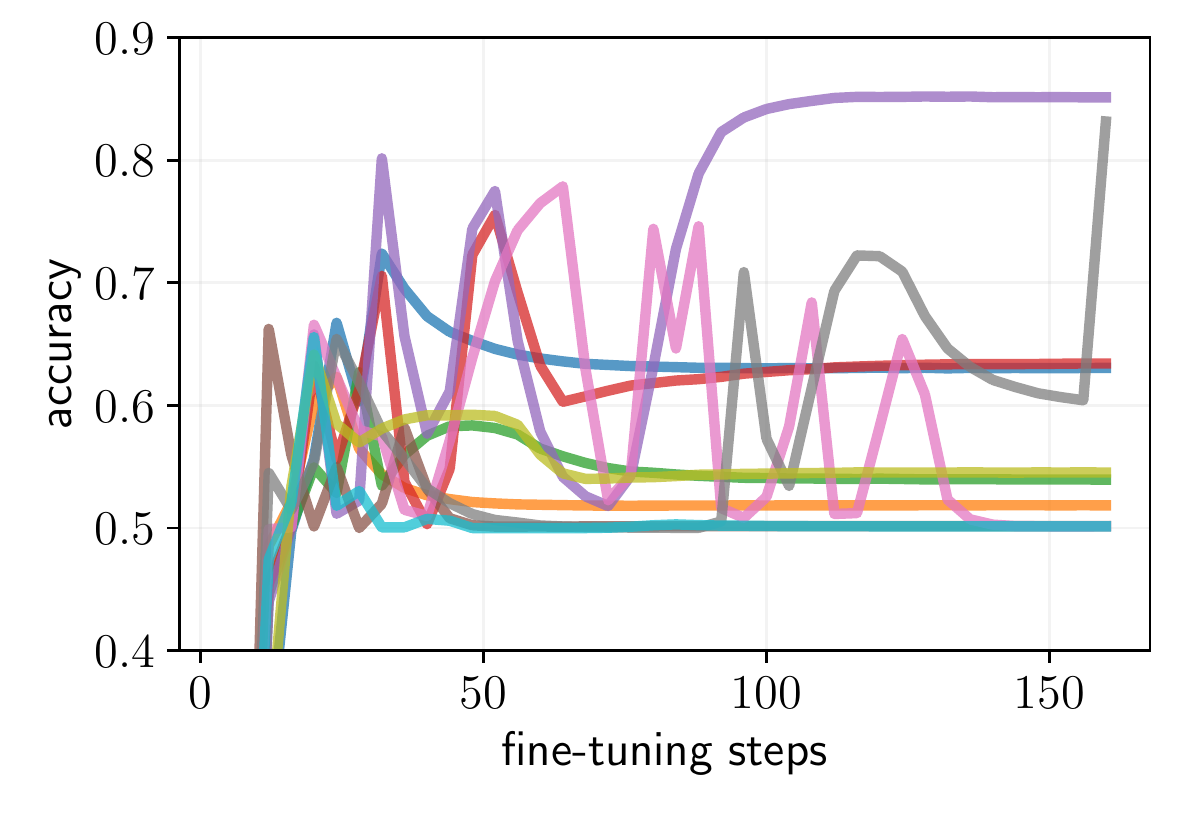}
        \caption{13B -- out-of-domain}
    \end{subfigure}
    \\
    \begin{subfigure}[b]{0.35\textwidth}
        \centering
        \includegraphics[width=\textwidth]{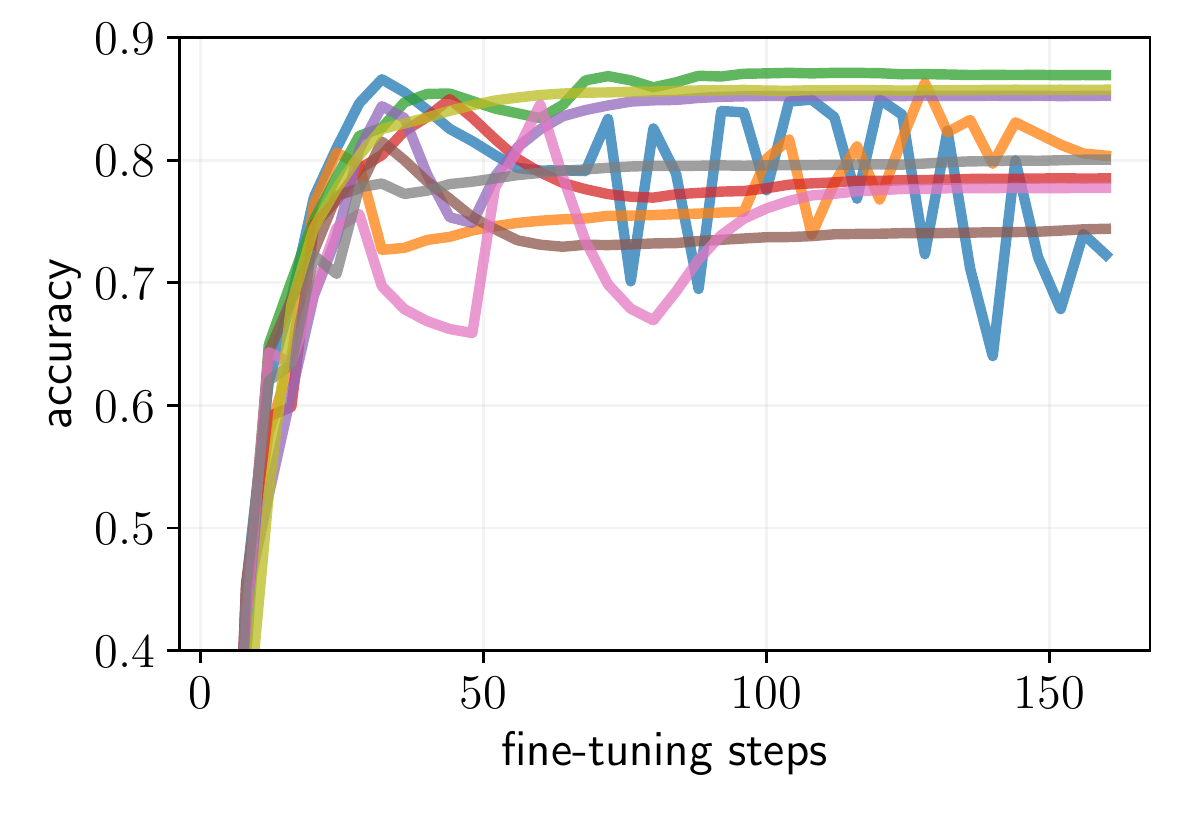}
        \caption{30B -- in-domain}
    \end{subfigure}
    ~
    \begin{subfigure}[b]{0.35\textwidth}
        \centering
        \includegraphics[width=\textwidth]{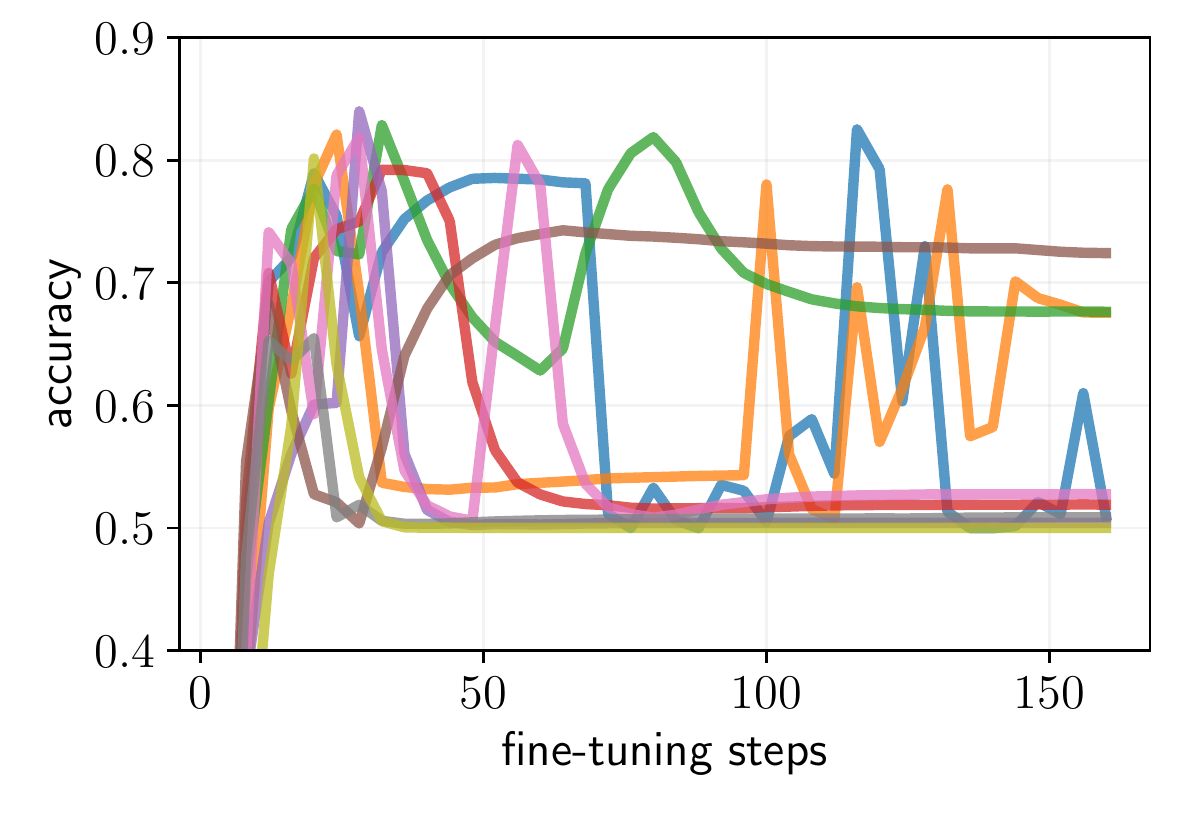}
        \caption{30B -- out-of-domain}
    \end{subfigure}
    
    \caption{\textbf{Generalization throughout PBFT on MNLI for OPT models of various sizes.} We train on 128 examples. Colors denote different data seeds. First column shows in-domain, second column out-of-domain performance.
    }
    \label{fig:appendix-individual-runs-mnli}
\end{figure*}

\begin{figure*}[h]
    \centering
    \begin{subfigure}[b]{0.35\textwidth}
        \centering
        \includegraphics[width=\textwidth]{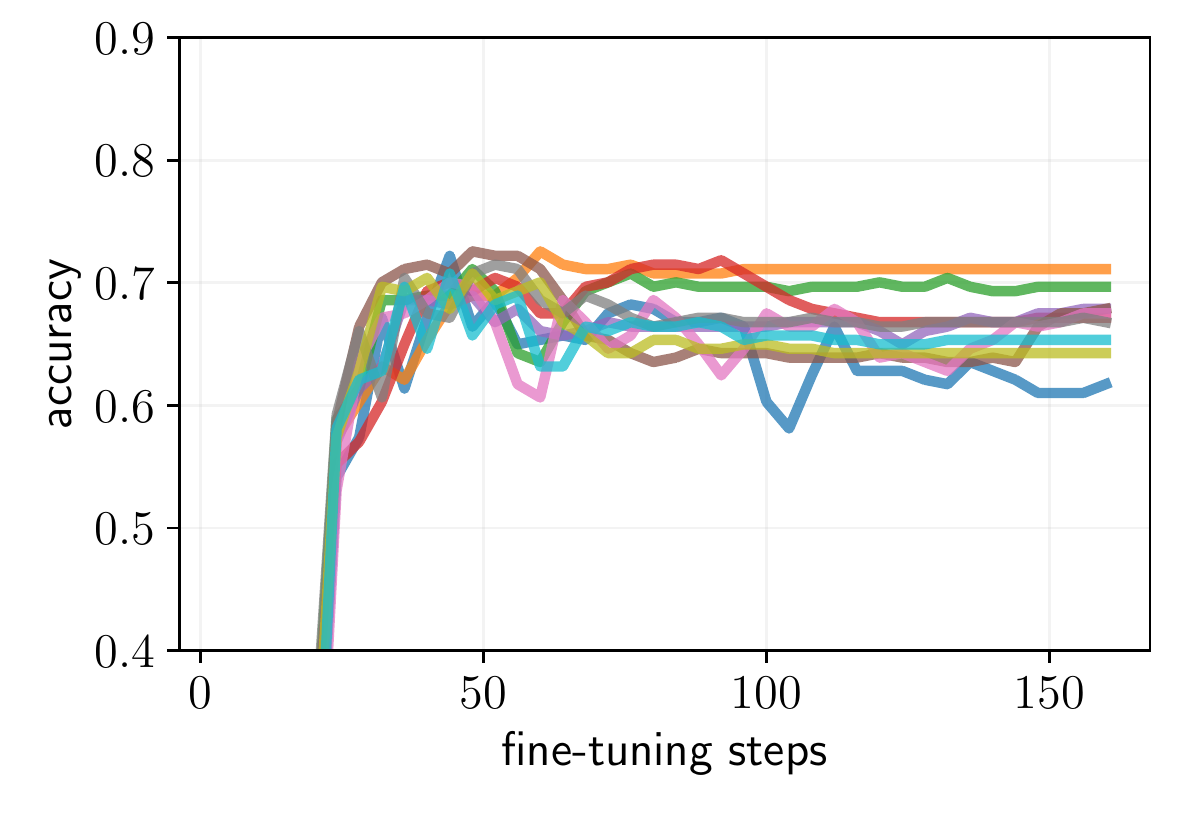}
        \caption{1.3B -- in-domain}
    \end{subfigure}
    ~
    \begin{subfigure}[b]{0.35\textwidth}
        \centering
        \includegraphics[width=\textwidth]{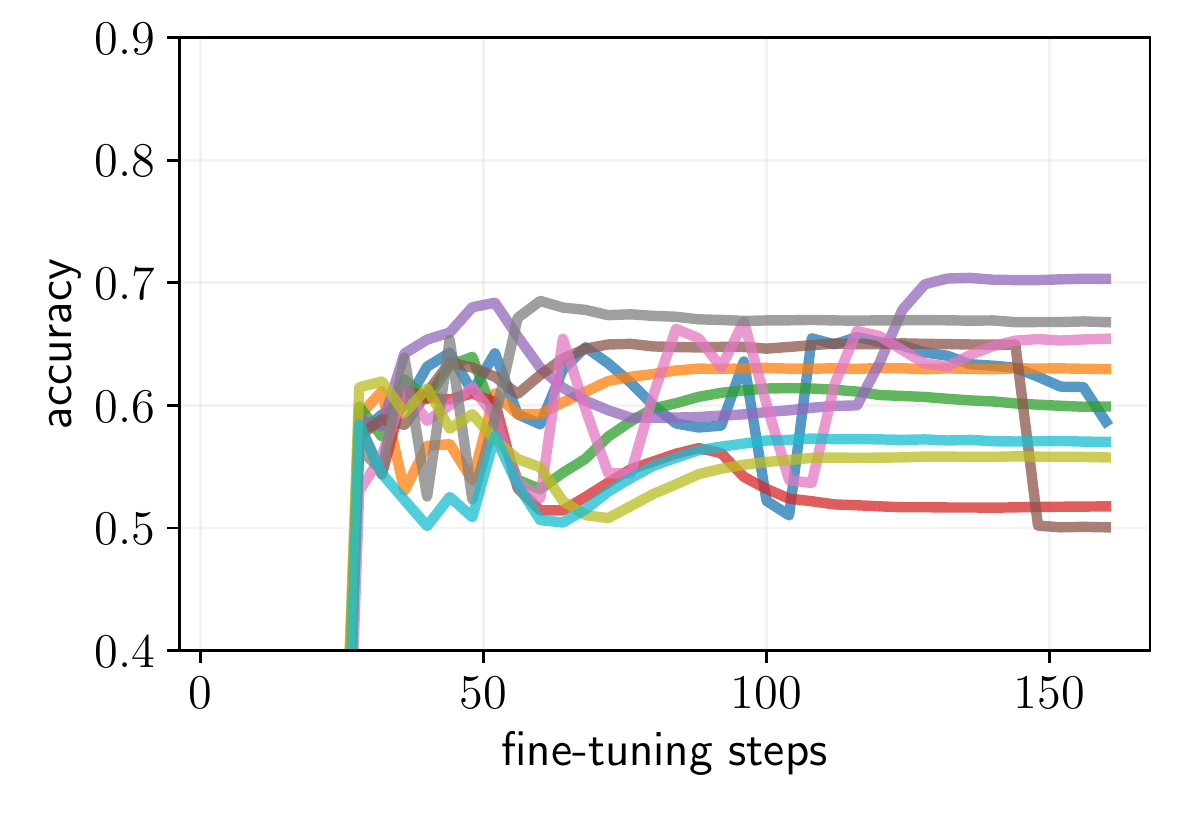}
        \caption{1.3B -- out-of-domain}
    \end{subfigure}
    \\
    \begin{subfigure}[b]{0.35\textwidth}
        \centering
        \includegraphics[width=\textwidth]{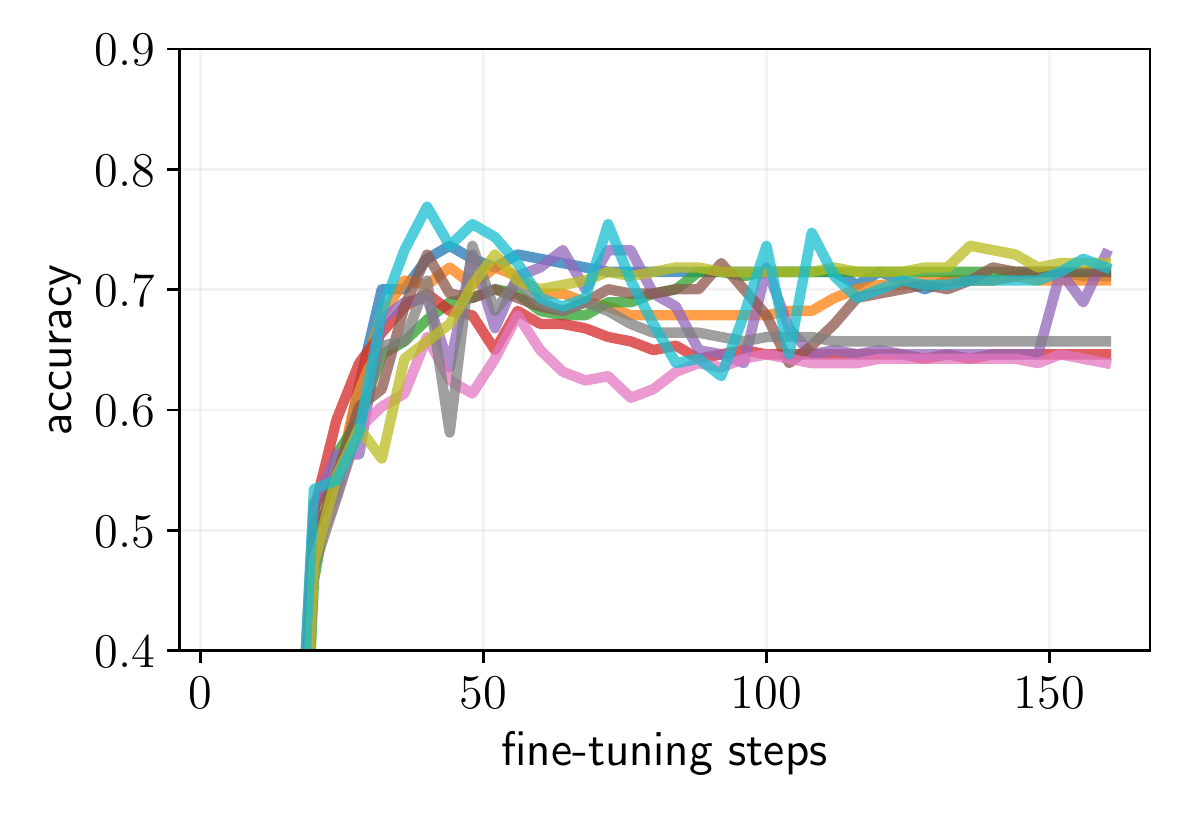}
        \caption{2.7B -- in-domain}
    \end{subfigure}
    ~
    \begin{subfigure}[b]{0.35\textwidth}
        \centering
        \includegraphics[width=\textwidth]{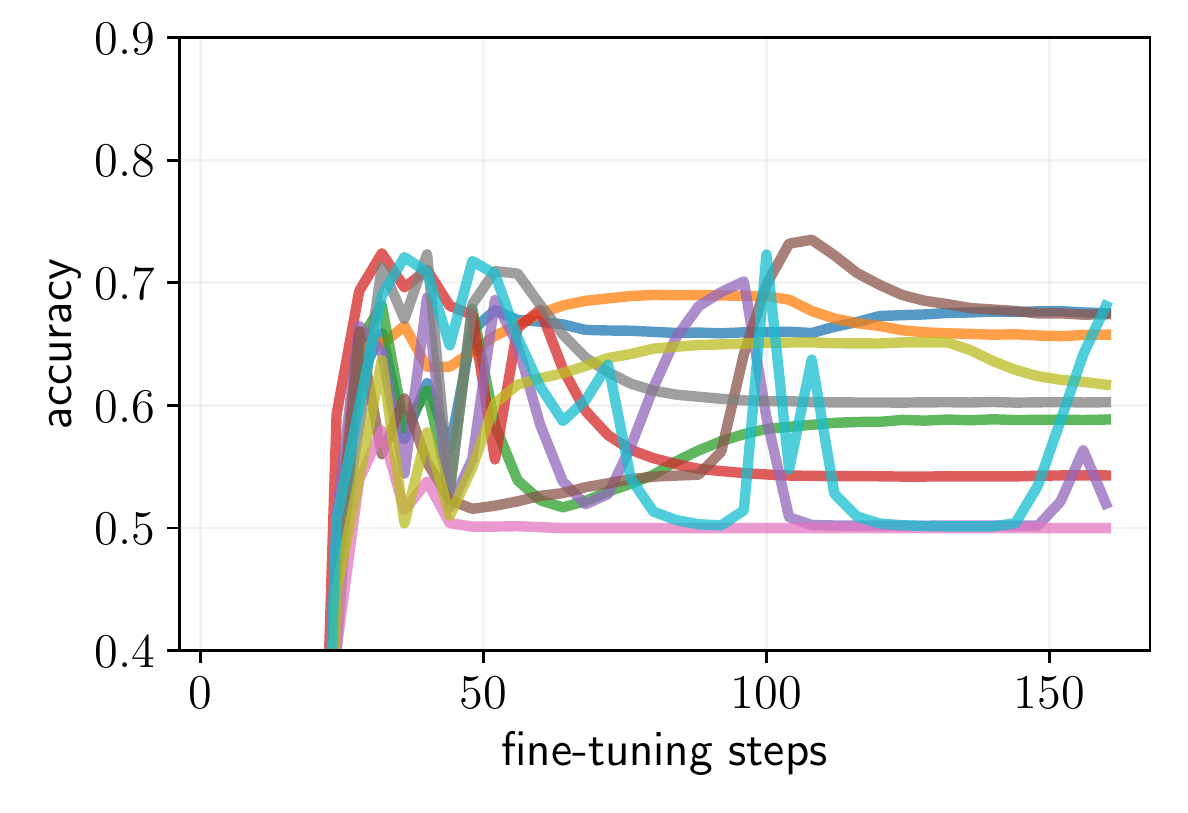}
        \caption{2.7B -- out-of-domain}
    \end{subfigure}
    \\
    \begin{subfigure}[b]{0.35\textwidth}
        \centering
        \includegraphics[width=\textwidth]{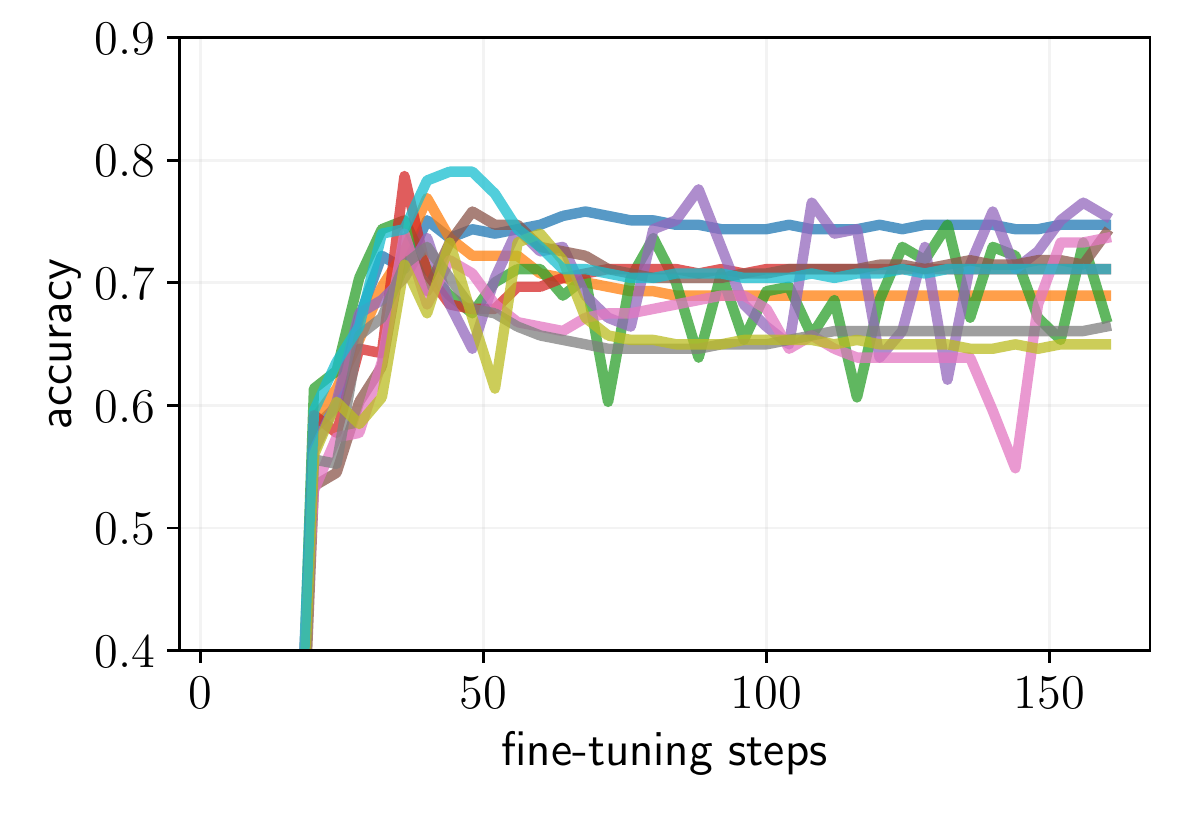}
        \caption{6.7B -- in-domain}
    \end{subfigure}
    ~
    \begin{subfigure}[b]{0.35\textwidth}
        \centering
        \includegraphics[width=\textwidth]{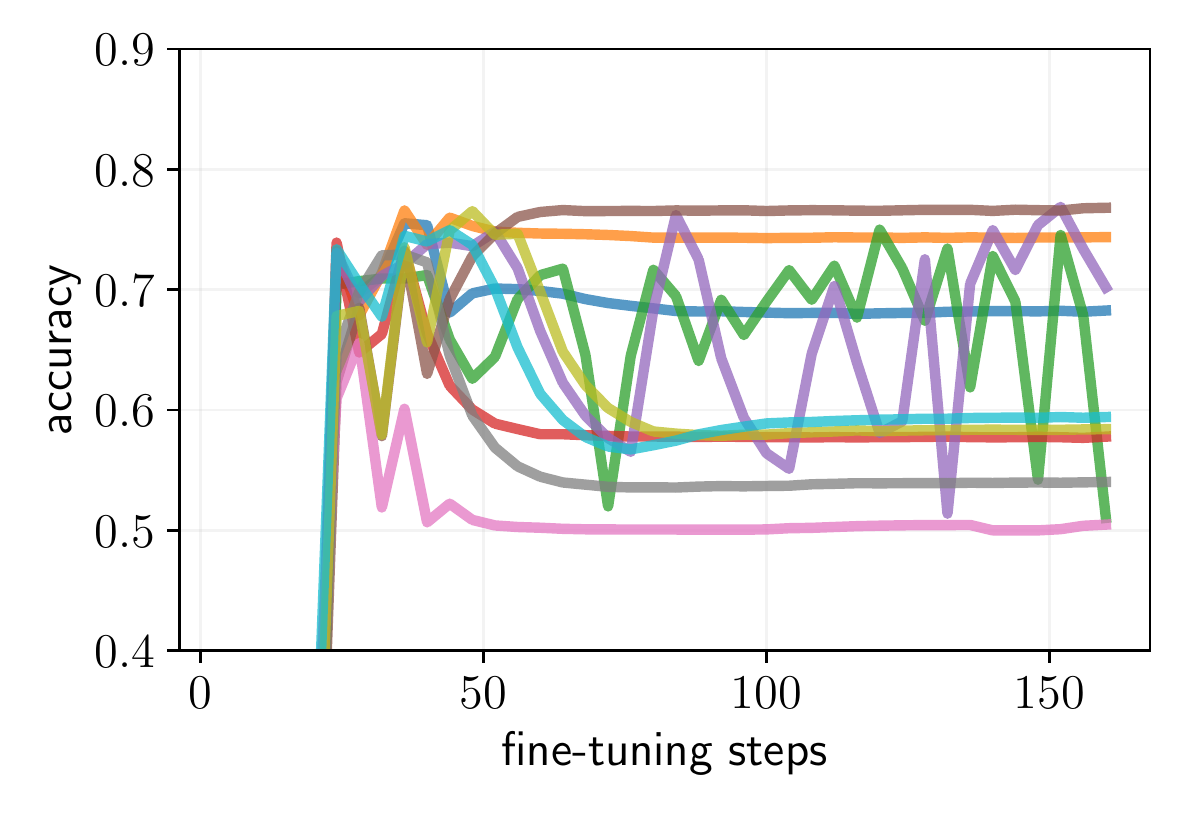}
        \caption{6.7B -- out-of-domain}
    \end{subfigure}
    \\
    \begin{subfigure}[b]{0.35\textwidth}
        \centering
        \includegraphics[width=\textwidth]{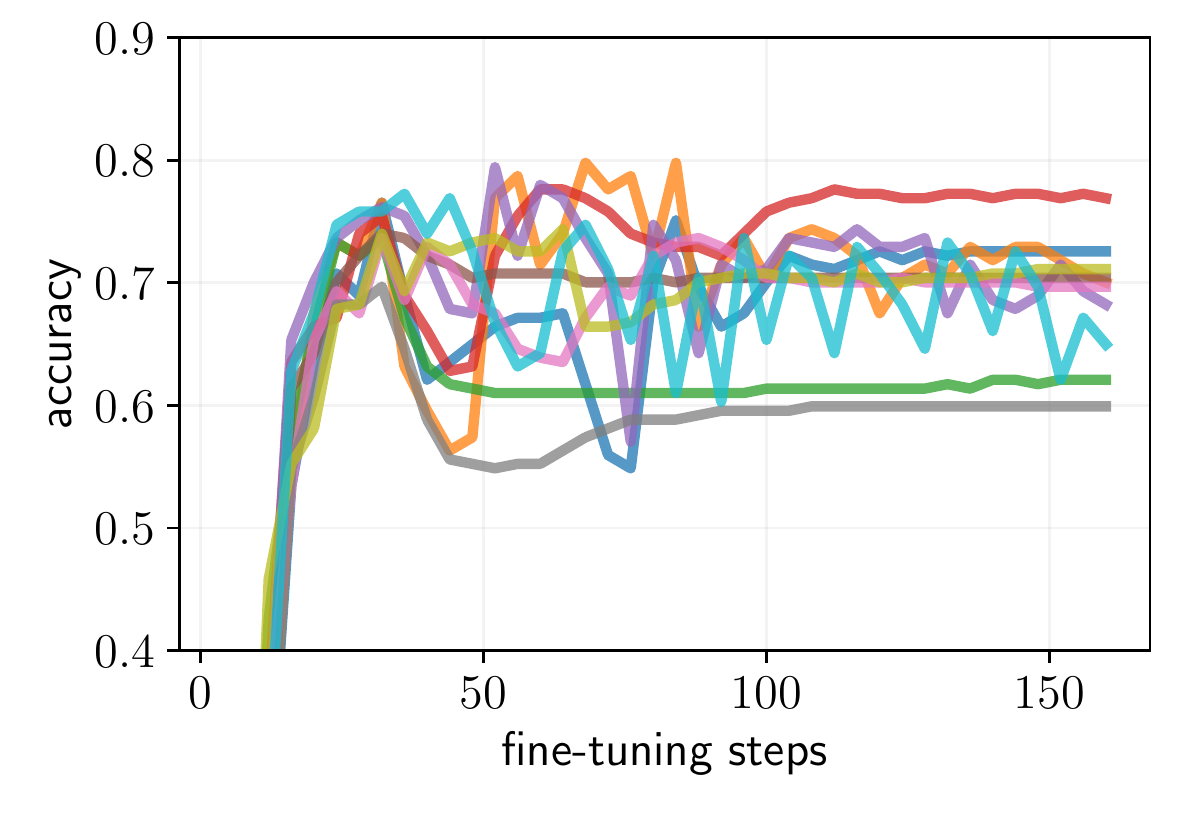}
        \caption{13B -- in-domain}
    \end{subfigure}
    ~
    \begin{subfigure}[b]{0.35\textwidth}
        \centering
        \includegraphics[width=\textwidth]{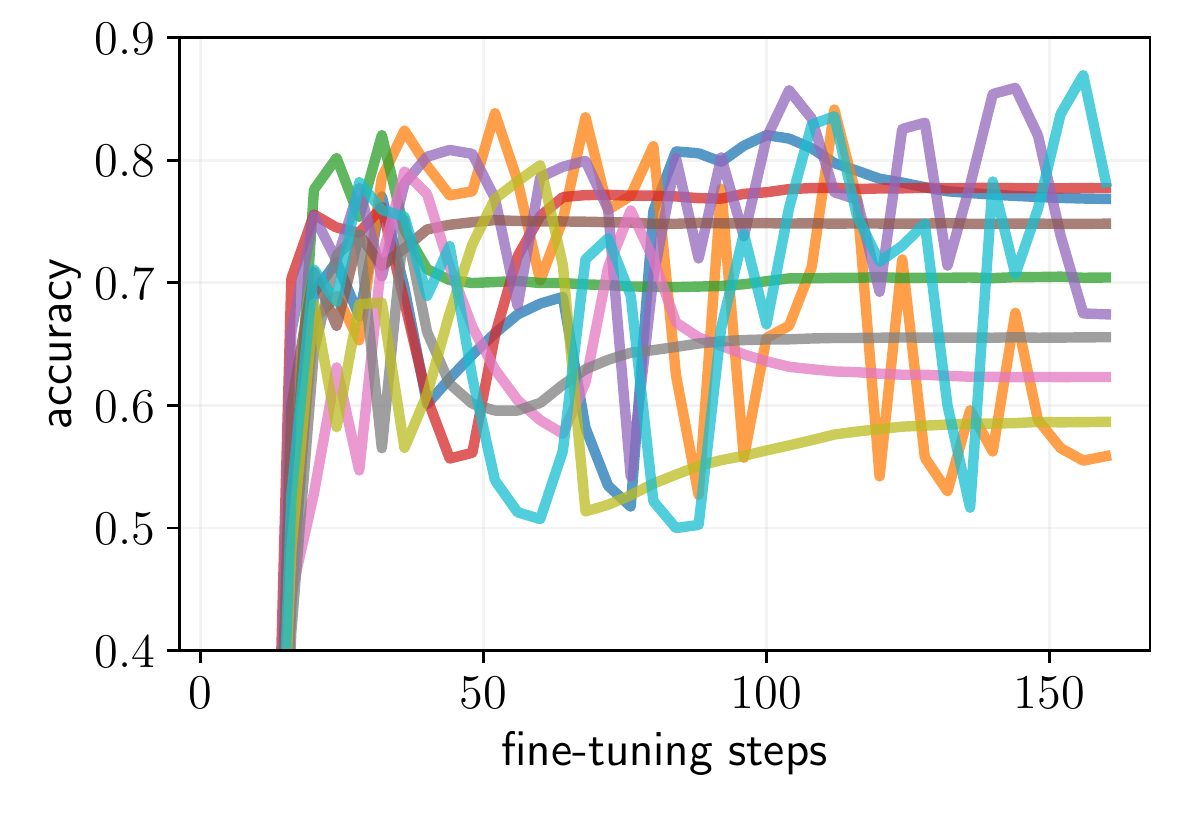}
        \caption{13B -- out-of-domain}
    \end{subfigure}
    \\
    \begin{subfigure}[b]{0.35\textwidth}
        \centering
        \includegraphics[width=\textwidth]{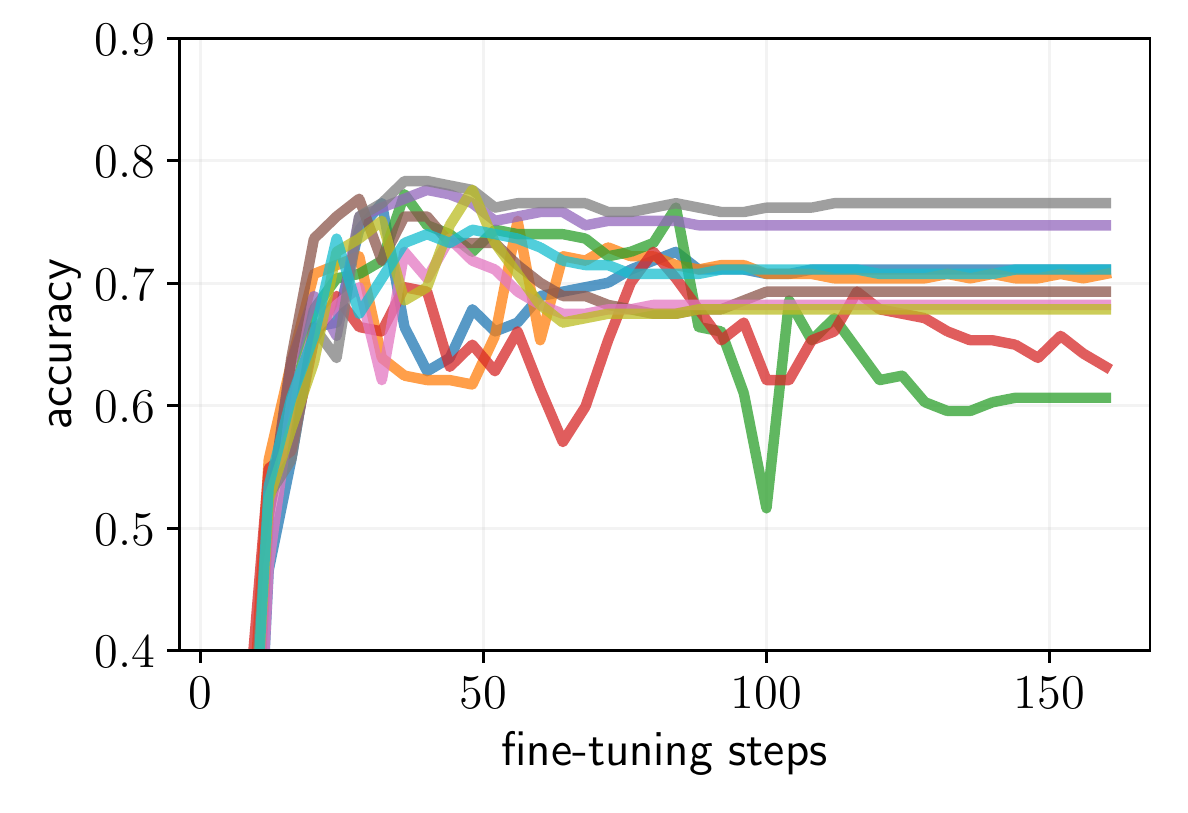}
        \caption{30B -- in-domain}
    \end{subfigure}
    ~
    \begin{subfigure}[b]{0.35\textwidth}
        \centering
        \includegraphics[width=\textwidth]{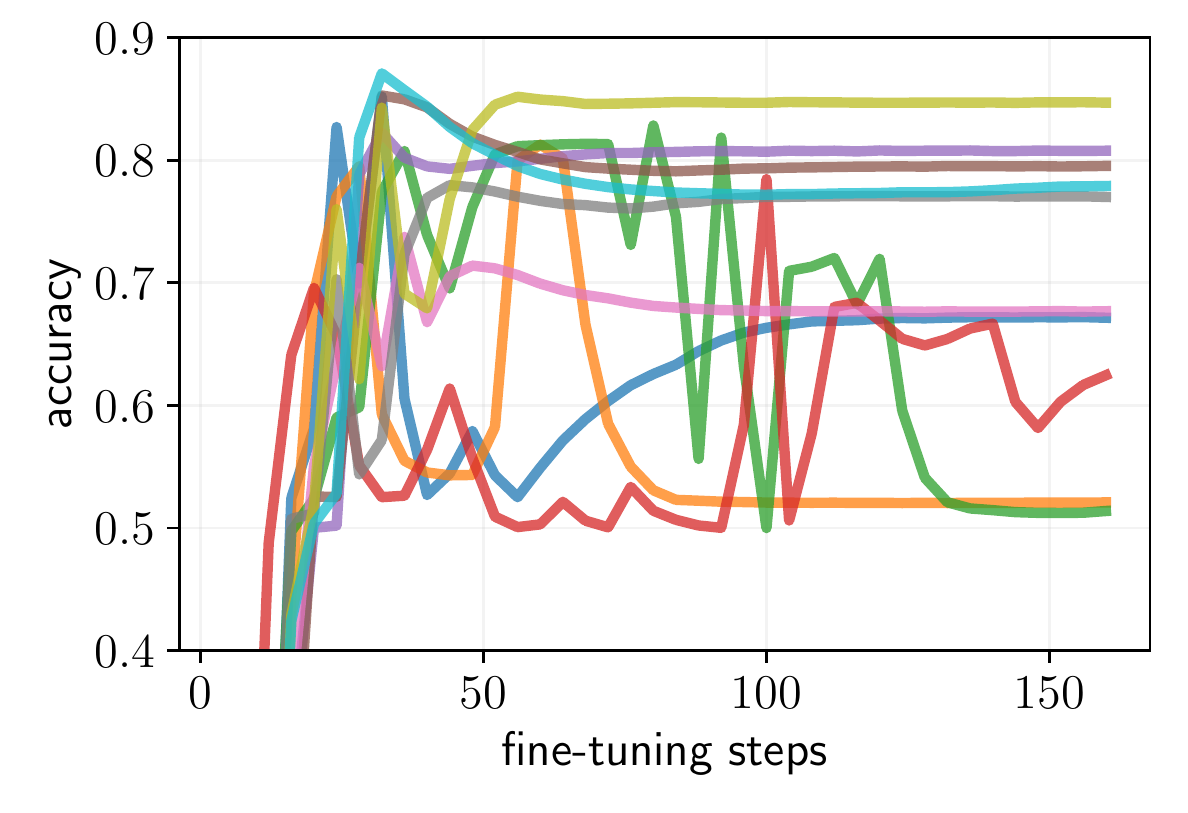}
        \caption{30B -- out-of-domain}
    \end{subfigure}
    
    \caption{\textbf{Generalization throughout PBFT on RTE for OPT models of various sizes.} We train on 128 examples. Colors denote different data seeds. First column shows in-domain, second column out-of-domain performance.
    }
    \label{fig:appendix-individual-runs-rte}
\end{figure*}

Looking at the in-domain and out-of-domain performance for individual checkpoints does not reveal the generalization behavior of individual FT runs during training. In particular, this view does not tell us how stable the generalization of individual runs is during FT. Therefore, in Figures \ref{fig:appendix-individual-runs-mnli} and \ref{fig:appendix-individual-runs-rte} we visualize both in-domain and out-of-domain performance throughout FT on MNLI and RTE when using 128 examples.
We observe that out-of-domain performance varies considerably across seeds and even during fine-tuning.

\end{document}